\documentclass[10pt,twocolumn,letterpaper]{article}

\usepackage[pagenumbers]{cvpr} %

\usepackage{framed,multirow}

\usepackage{amssymb}
\usepackage{latexsym}
\usepackage{subcaption}
\usepackage{enumitem}
\usepackage{float}
\usepackage{times}
\usepackage{epsfig}
\usepackage{graphicx}
\usepackage{amsmath}
\usepackage{amssymb}
\usepackage{pifont}
\usepackage{xcolor}
\usepackage{bbm}
\usepackage{xpatch}
\usepackage[percent]{overpic}
\usepackage{soul}
\usepackage{nicefrac}
\usepackage{bbold}
\usepackage{mathtools}
\usepackage{amssymb,amsmath,amsthm,enumitem}
\xapptocmd\normalsize{%
 \abovedisplayskip=12pt plus 3pt minus 9pt
 \abovedisplayshortskip=0pt plus 3pt
 \belowdisplayskip=12pt plus 3pt minus 9pt
 \belowdisplayshortskip=7pt plus 3pt minus 4pt
}{}{}

\usepackage[pagebackref,breaklinks,colorlinks]{hyperref}
\usepackage[linesnumbered,ruled,vlined]{algorithm2e}

\usepackage[capitalize]{cleveref}
\usepackage[accsupp]{axessibility}

\newcommand{\grayedout}[1]{{\color{gray}  #1}}
\definecolor{britishracinggreen}{rgb}{0.0, 0.26, 0.15}
\definecolor{cadmiumgreen}{rgb}{0.0, 0.42, 0.24}
\definecolor{darkspringgreen}{rgb}{0.09, 0.45, 0.27}
\definecolor{ferngreen}{rgb}{0.31, 0.47, 0.26}
\definecolor{forestgreen}{rgb}{0.0, 0.27, 0.13}
\definecolor{htmlcssgreen}{rgb}{0.0, 0.5, 0.0}
\newcommand{\increase}[1]{\footnotesize \textcolor{htmlcssgreen}{#1}\normalsize}

\newcommand{\supstar}{\textsuperscript{\textasteriskcentered}}

\newcommand{\xmark}{\ding{55}}
\newcommand{\cmark}{\ding{51}}

\definecolor{newcolor}{rgb}{.8,.349,.1}
\definecolor{c0}{rgb}{0.12156862745098039,0.4666666666666667,0.7058823529411765}
\definecolor{c1}{rgb}{1.0,0.4980392156862745,0.054901960784313725}
\definecolor{c2}{rgb}{0.17254901960784313,0.6274509803921569,0.17254901960784313}
\definecolor{c3}{rgb}{0.8392156862745098,0.15294117647058825,0.1568627450980392}
\definecolor{c4}{rgb}{0.5803921568627451,0.403921568627451,0.7411764705882353}
\definecolor{c5}{rgb}{0.5490196078431373,0.33725490196078434,0.29411764705882354}
\definecolor{c6}{rgb}{0.8901960784313725,0.4666666666666667,0.7607843137254902}
\definecolor{c7}{rgb}{0.4980392156862745,0.4980392156862745,0.4980392156862745}
\definecolor{c8}{rgb}{0.7372549019607844,0.7411764705882353,0.13333333333333333}
\definecolor{c9}{rgb}{0.09019607843137255,0.7450980392156863,0.8117647058823529}
\definecolor{FieldHockeyPenalty}{rgb}{0.12156862745098039,0.4666666666666667,0.7058823529411765}
\definecolor{GolfSwing}{rgb}{1.0,0.4980392156862745,0.054901960784313725}
\definecolor{SoccerJuggling}{rgb}{0.17254901960784313,0.6274509803921569,0.17254901960784313}
\definecolor{SoccerPenalty}{rgb}{0.8392156862745098,0.15294117647058825,0.1568627450980392}
\definecolor{PlayingGuitar}{rgb}{0.5803921568627451,0.403921568627451,0.7411764705882353}
\definecolor{PlayingPiano}{rgb}{0.5490196078431373,0.33725490196078434,0.29411764705882354}
\definecolor{PlayingSitar}{rgb}{0.8901960784313725,0.4666666666666667,0.7607843137254902}
\definecolor{ApplyEyeMakeup}{rgb}{0.4980392156862745,0.4980392156862745,0.4980392156862745}
\definecolor{ApplyLipstick}{rgb}{0.7372549019607844,0.7411764705882353,0.13333333333333333}
\definecolor{BlowDryHair}{rgb}{0.09019607843137255,0.7450980392156863,0.8117647058823529}

\begin{document}

\title{\textbf{TCLR}: Temporal Contrastive Learning for Video Representation}

\author{Ishan Dave, Rohit Gupta, Mamshad Nayeem Rizve, Mubarak Shah\\
Center for Research in Computer Vision, University of Central Florida, Orlando, USA\\
\tt\small \{ishandave, rohitg, nayeemrizve\}@knights.ucf.edu, shah@crcv.ucf.edu
}

\maketitle

\begin{abstract}
Contrastive learning has nearly closed the gap between supervised and self-supervised learning of image representations, and has also been explored for videos. However, prior work on contrastive learning for video data has not explored the effect of explicitly encouraging the features to be distinct across the temporal dimension. We develop a new temporal contrastive learning framework consisting of two novel losses to improve upon existing contrastive self-supervised video representation learning methods. The local-local temporal contrastive loss adds the task of discriminating between non-overlapping clips from the same video, whereas the global-local temporal contrastive aims to discriminate between timesteps of the feature map of an input clip in order to increase the temporal diversity of the learned features. Our proposed temporal contrastive learning framework achieves significant improvement over the state-of-the-art results in various downstream video understanding tasks such as action recognition, limited-label action classification, and nearest-neighbor video retrieval on multiple video datasets and backbones. We also demonstrate significant improvement in fine-grained action classification for visually similar classes. With the commonly used 3D ResNet-18 architecture with UCF101 pretraining, we achieve 82.4\% (+5.1\% increase over the previous best) top-1 accuracy on UCF101 and 52.9\% (+5.4\% increase) on HMDB51 action classification, and 56.2\% (+11.7\% increase) Top-1 Recall on UCF101 nearest neighbor video retrieval. Code released at \url{https://github.com/DAVEISHAN/TCLR}.
\end{abstract}

\section{Introduction}
\vspace{-1mm}

Large-scale labeled datasets such as Kinetics~\cite{kinetics}, LSHVU~\cite{diba2020large} etc have been crucial for recent advances in video understanding tasks. Since training a video encoder using existing supervised learning approaches is label-inefficient~\cite{kenshohara_mega_scale}, annotated video data is required at a large scale. This costs enormous human effort and time, much more so than annotating images. At the same time, a tremendous amount of unlabeled video data is easily available on the internet. Research in self-supervised video representation learning can unlock the corpus of readily available unlabeled video data and unshackle progress in video understanding.

Recently, Contrastive Self-supervised Learning (CSL) based methods~\cite{simclr, moco, swav} have demonstrated the ability to learn powerful {\em image representations} in a self-supervised manner, and have narrowed down the performance gap between {\em unsupervised} and {\em supervised} representation learning on various image understanding downstream tasks.

\begin{figure}
    \centering
    \subfloat[Running and Jumping: distinct stages of \texttt{LongJump} action]{{\includegraphics[clip, trim=0cm 0.6cm 0cm 0cm,width=0.99\linewidth]{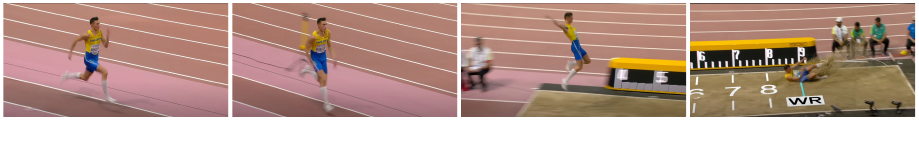}{\label{fig:longjump}} }} \\
    \subfloat[NN-Retrieval]{{\includegraphics[width=0.49\linewidth]{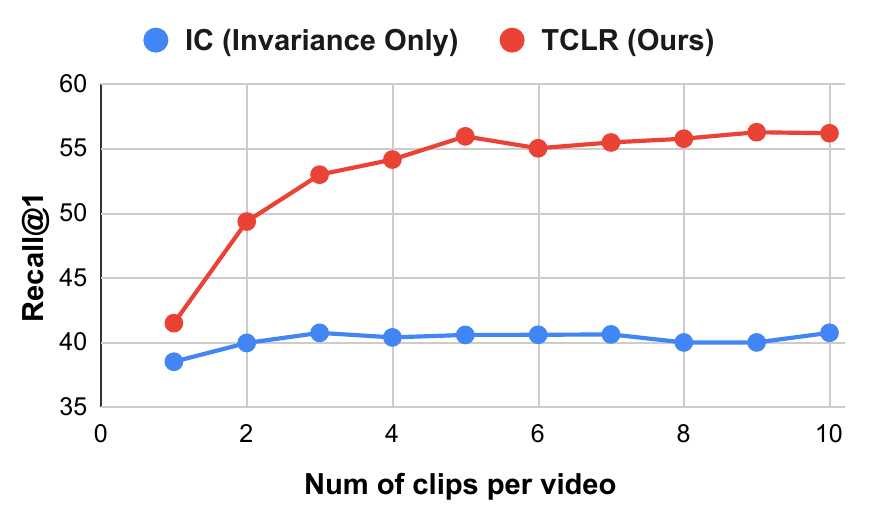}{\label{fig:nnr}} }}\hspace*{-0.4em}
    \subfloat[Linear Classification]{{\includegraphics[width=0.49\linewidth]{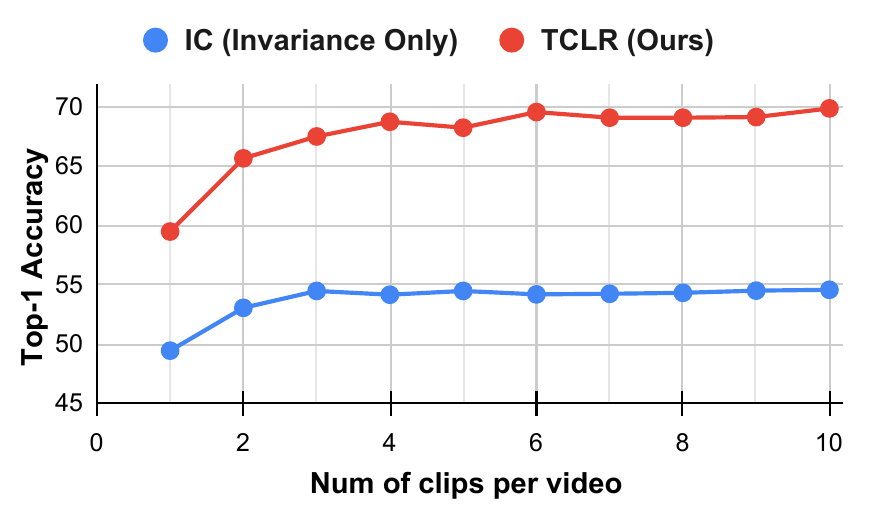}{\label{fig:lineval}} }}
    \caption{Videos from standard action recognition datasets often have distinct temporal stages. For example, in figure \textbf{(a)} we can see the two distinct stages (Running and Jumping) of the \texttt{LongJump} action. Typically predictions across multiple short clips are aggregated, as a single short clip may not capture both stages of the action. We show the comparison of vanilla instance discrimination based contrastive (IC) self-supervision and our proposed TCLR method on \textbf{(b)} Nearest neighbour retrieval and \textbf{(c)} Linear classification tasks. We find that IC trained models do not benefit much from using multiple clips during evaluation. This is a result of IC imposing within instance temporal invariance. This motivates our proposed TCLR pre-training, which explicitly encourages learning distinct features across time.}
    \vspace{-4mm}
    \label{fig:intro}

\end{figure}

\begin{figure*}
\begin{center}
  \includegraphics[width=\textwidth]{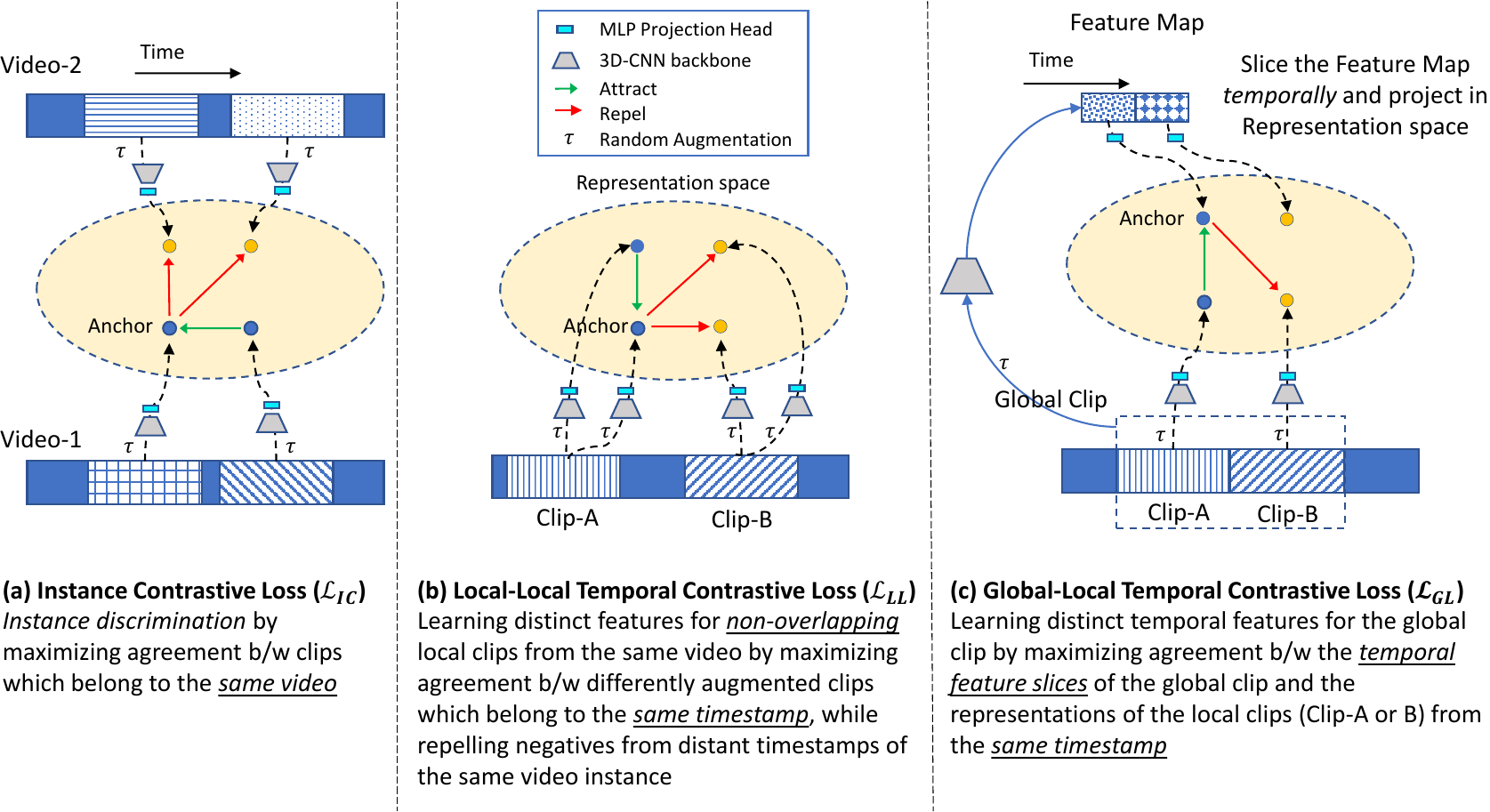}
\end{center}
\vspace{-5mm}
\caption{The proposed temporal contrastive learning framework (TCLR) for learning temporally distinct video representations consists of three different losses.}

  \vspace{-5mm}

\label{fig:main}
\end{figure*}

A simple yet effective extension of CSL to the video domain can be obtained by using the InfoNCE instance discrimination objective, where the model learns to distinguish clips of a given video from the clips of other videos in the dataset (see Figure~\ref{fig:main}a). Unlike images, videos have both time-invariant and the temporally varying properties. For example, in a \texttt{LongJump} video from UCF101 (See Figure~\ref{fig:intro}), running and jumping represent two very different stages of the action. Usually, video understanding models utilize temporally varying features by aggregating along the temporal dimension to obtain a video level prediction. While the  significant success can be achieved on many video understanding tasks by only modelling the temporally invariant properties, it maybe possible that the temporally varying properties can also play an important role in further improvements on these tasks. Whether video representations should be invariant or distinct along the temporal dimension is an open question in the literature.  Instance contrastive pre-training, however, encourages the model to learn similar features to represent temporally distant clips from the video, i.e. it enforces temporal invariance on the features. While instance level contrastive learning lies on one end of the spectrum, some recent works have tried to relax the invariance constraint through various means, such as, using a weighted temporal sampling strategy to avoid enforcing invariance between temporally distant clips~\cite{cvrl}, cross-modal mining of positive samples from across video instances~\cite{cotraining} or adding additional pretext tasks that require learning temporal features~\cite{pace_pred, iclr21submitted_st_invariant, seco, taco}.

We take a different approach by explicitly encouraging the learning of temporally distinct video representations. The challenge with video classification is modelling variable length videos with a fixed number of parameters. 3D CNNs tackle this challenge by temporal aggregation of features across two levels: averaging across distinct fixed length temporal segments of a video (clips) and also temporal pooling across the feature map of each clip. Based on this observation, we propose two different temporal contrastive losses in order to learn temporally distinct features across the video: one which acts across clips of the same video, and another which acts across the timesteps of the feature map of the same clip. Combined with the vanilla instance contrastive loss, these novel losses result in an increase in the temporal diversity of the learned features, and better accuracy on downstream tasks.

Our first proposed loss is the {\em local-local temporal contrastive loss} (Figure~\ref{fig:main}b), which ensures that temporally non-overlapping clips from the same video are mapped to distinct representations. This loss treats randomly augmented versions of the same clip as positive pairs to be brought together, and other non-overlapping clips from the same video as negative matches to be pushed away. While the local-local loss ensures that distinct clips have distinct representations, in order to encourage temporal variation {\em within each clip}, we introduce a second temporal contrastive loss, the {\em global-local temporal contrastive loss} (Figure~\ref{fig:main}c). This loss constrains the \textit{timesteps} of the feature map of a long \textit{``global"} video clip to match the representations of the temporally aligned shorter \textit{``local"} video clips.

Our complete framework is called \textit{Temporal Contrastive Learning of video Representations} (henceforth referred to as \textit{TCLR}). TCLR retains the ability of representations to successfully discriminate between video instances due to its instance contrastive loss. In addition, TCLR attempts to capture the {\em within-instance} temporal variation. Through extensive experiments on various downstream video understanding tasks, we demonstrate that both of our proposed Temporal Contrastive losses contribute to the learning of powerful video representations, and provide significant improvements.

\noindent{The original contributions of this work can be summarized as below:}
\setlist{nolistsep}
\begin{itemize}
    \item TCLR is the first contrastive learning framework to explicitly enforce \textit{within} instance temporal feature variation for video understanding tasks. 
    \item Novel {\em local-local} and {\em global-local temporal contrastive losses}, which when combined with the standard instance contrastive loss significantly outperform the state-of-the-art on various downstream video understanding tasks like action recognition, nearest neighbor video retrieval and action classification with limited labeled data, while using 3 different 3D CNN architectures and 2 datasets (UCF101 \& HMDB51).  
    \item We propose the use of the challenging Diving48 fine-grained action classification task for evaluating the quality of learned video representations. 
    
\end{itemize}

\section{Related Work}
 Recent approaches for self-supervised video representation learning can be categorized into two major groups based on the self-supervised learning objective: (1) Pretext task based methods, and (2) Contrastive Learning based methods.

\noindent{\bf Pretext task based approaches:}  Various pretext tasks have been devised for self-supervised video representation learning based on learning the correct temporal order of the data: verifying correct frame order~\cite{misra2016shuffle}, identifying the correctly ordered tuple from a set of shuffled orderings~\cite{fernando2017self, suzuki2018learning}, sorting frame order~\cite{lee2017unsupervised}, and predicting clip order~\cite{vcop}. Some methods extend existing pretext tasks from the image domain to video domain, for example, solving spatio-temporal jigsaw puzzles~\cite{videojigsaw, aaai19, csj} and identifying the rotation of transformed video clips~\cite{3drotnet}. Many recent works rely on predicting video properties like playback rate of the video~\cite{cho2020self,prp,pace_pred,iclr21submitted_st_invariant}, temporal transformation that has been applied from a given set~\cite{simon,jenni2021time}, speediness of moving objects~\cite{speedNet}, and motion and appearance statistics of the video~\cite{statistics_cvpr19, statistics2}.

\noindent {\bf Contrastive Self-supervised Learning (CSL) based approaches:}  Following the success of contrastive learning approaches of self-supervised image representation learning such as SimCLR~\cite{simclr} and MoCo~\cite{moco}, there have been many extensions of contrastive learning to the video domain. For instance, various video CSL methods~\cite{pace_pred,videomoco, taco, cvrl, iic, seco,idt, iclr21submitted_st_invariant,vtc} leverage \textit{Instance level Discrimination} objectives, and build their method upon them, where clips from the same video are treated as positives and clips from the different videos as negatives. CVRL~\cite{cvrl} studies the importance of temporal augmentation and develops a temporal sampler to avoid enforcing excessive temporal invariance in learning video representation. VideoMoCo~\cite{videomoco} improves image-based MoCo framework for video representation by encouraging temporal robustness of the encoder and modeling temporal decay of the keys. VTHCL~\cite{vtc} employs SlowFast architecture~\cite{slowfast} and uses contrastive loss with the slow and fast pathway representations as the positive pair. VIE~\cite{vie} is proposed as a deep neural embedding-based method to learn video representation in an unsupervised manner, by combining both static image representation from 2D CNN and dynamic motion representation from 3D CNN. Generative contrastive learning-based approaches such as predicting the the dense representation of the next video block~\cite{dpc, memdpc}, or Contrastive Predictive Coding (CPC)~\cite{cpc} for videos~\cite{wacv20} have also been studied in the literature. 

AMDIM~\cite{amdim} is another CSL approach for image representation learning, where a local view (spatial slice of the feature map taken from an intermediate layer) and a global view (full feature map) of differently augmented versions of the same image are considered as a positive pair, and global views of other images form negative pair of the contrastive loss. The method is adapted for the video domain~\cite{vdim,dvim} by generating local views from the spatio-temporal features. Unlike this class of methods, which try to maximize agreement across features from different levels of the encoder, our Global-Local loss tries to learn distinct features across temporal slices of the feature map instead. 

Some recent works combine pretext tasks along with contrastive learning in a multi-task setting to learn temporally varying features in the video representation. For example, using video clips with different playback rates as positive pairs for contrastive loss along with predicting the playback rate~\cite{pace_pred}, or temporal transforms~\cite{taco}. Other works propose frame-based contrastive learning, along with existing pretext tasks of frame rotation prediction~\cite{tce} and frame-tuple order verification~\cite{seco}. Unlike these works, TCLR takes a different approach by adding explicit temporal contrastive losses that encourage temporal diversity in the learned features, instead of utilizing a pretext task for this purpose. 

Some works which try to capture intra-video variance using optical flow, but are nevertheless interesting to compare with. IIC~\cite{iic} uses intra-instance negatives, but it relies on frame repeating and shuffling to generate these ``hard" negatives, and does not focus on learning distinct features across the temporal axis. DSM~\cite{scene-motion} tries to decouple scene and motion features by an intra-instance triplet loss, which uses negatives generated by optical flow scaling and spatial warping. Some recent works use extra supervisory signals in addition to the RGB video data to learn video representation in a self-supervised manner. However, these methods either require additional cross-modal data (e.g. text narration~\cite{miech2020end}, audio~\cite{afouras2020self}) or expensive and time-consuming computation of hand-crafted visual priors (e.g. optical flow~\cite{iic,cbt,aot,local_motion_cues,cotraining} or dense trajectories~\cite{idt}). In this work we focus only on learning from RGB data without using any auxiliary data from any extra modality or additionally computed visual priors. 

\section{Method}
\vspace{-2mm}

The key idea in our proposed framework is to learn two levels of contrastive discrimination: instance discrimination using the {\em instance contrastive loss} and within-instance temporal level discrimination using our novel temporal contrastive losses. The two different temporal contrastive losses which are applied within the same video instance: {\em Local-Local Loss} and {\em Global-Local loss}. Each of these losses is explained in the following sections.

\subsection{Instance Contrastive Loss}

We leverage the idea of \textit{instance discrimination} using InfoNCE~\cite{nce} based contrastive loss for learning video representations. In the video domain, in addition to leveraging image-based spatial augmentations, temporal augmentations can also be applied to generate different transformed versions of a particular instance. For a video instance, we extract various  clips (starting from different timestamps and/or having different frame sampling rates). We consider a randomly sampled mini-batch of size $N_B$ from different video instances, and from each instance we extract a pair of clips from random timesteps resulting in a total of 2N clips. The extracted clips are augmented using standard stochastic appearance and geometric transformations.\footnote{More details about the augmentations are available in Section \ref{para:dataset} and Section C of the supplementary material.} Each of the transformed clips is then passed through a 3D-CNN based video encoder which is followed by a non-linear projection head (multi-layer perceptron) to project the encoded features on the representation space. Hence, for each video-instance $i$ we get two clip representations $(G_{i}, G'_{i})$. The instance contrastive loss is defined as follows:

\begin{equation}\label{eq:IC}
\small
  \mathcal{L}_{IC}^{i}=-\log \frac{h\left(G_{i}, G'_{i}\right)}{\sum_{j=1}^{N_B}[\mathbb{1}_{[j\neq i]} h(G_{i}, G_{j}) + h(G_{i}, G'_{j})]},
\end{equation}

\noindent where, $h(u, v)=\exp \left(u^{T}v/(\|u\| \|v\| \theta) \right)$ is used to compute the similarity between $u$ and $v$ vectors with an adjustable parameter temperature, $\theta$. $\mathbb{1}_{[j\neq i]} \in \{0, 1\}$ is an indicator function which equals 1 iff $j \neq i$.
\subsection{Temporal Contrastive Losses}

For self-supervised training using the instance contrastive loss of Equation~\ref{eq:IC}, the model is presented with multiple clips cropped from random spatio-temporal locations within a single video as positive matches. This encourages the model to become invariant to the inherent variation present within an instance. In order to enable contrastive learning to represent within instance temporal variation, we introduce two novel temporal contrastive losses: {\em local-local temporal contrastive loss} and {\em global-local temporal contrastive loss}.

\subsubsection{Local-Local Temporal Contrastive Loss}
\label{sec:ccl}

\begin{figure}[t]
\centering
    \includegraphics[width=0.8\linewidth]{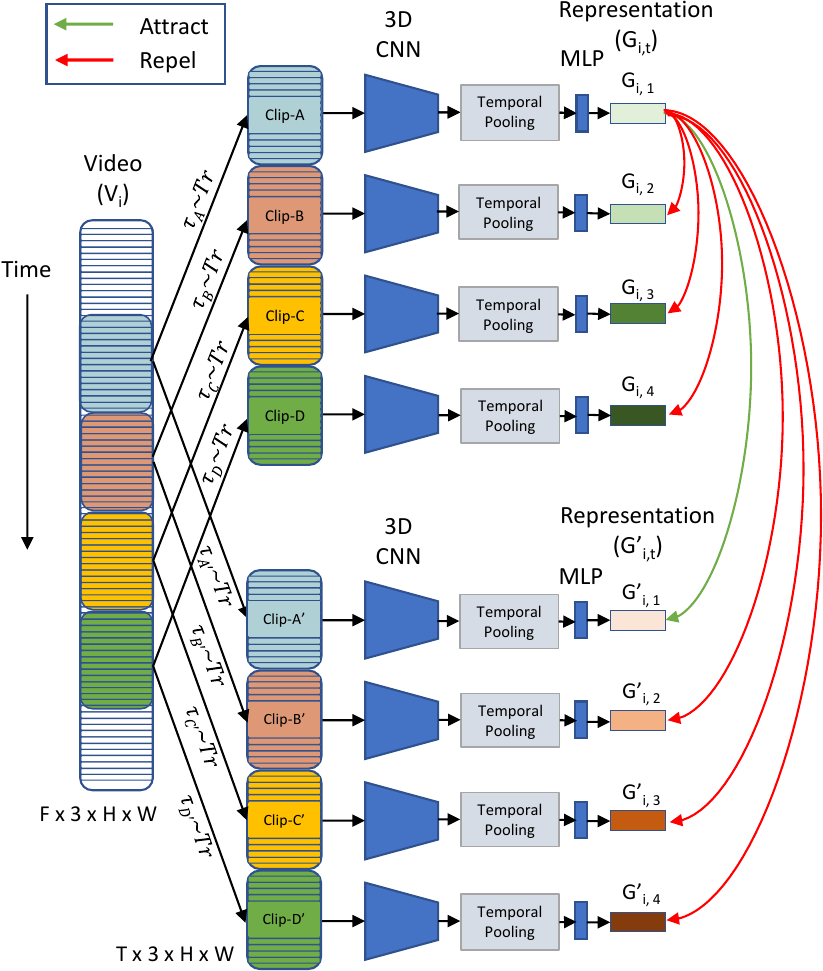}
    \vspace{-2mm}
  \caption{\textbf{Local-Local Temporal Contrastive Loss} is applied to representations of non-overlapping clips extracted from same video instance $i$. For the clip starting at timestep $p$, two randomly transformed versions are generated and their representations $G_{i,p}$ and $G'_{i,p}$ serve as the positive pair for the loss, whereas the other non-overlapping clips along with the anchor, $G_{i,p}$, form the negative pairs. $p=1$ serves as anchor, further details in Section~\ref{sec:ccl}. $\tau_{A}$, $\tau_{A'}$, ..., $\tau_{D'}$ are random set of augmentation sampled from universal set $Tr$.}
\label{fig:tcw}
\end{figure}

For this loss, we treat non-overlapping clips sampled from different temporal segments of the same video instance as negative pairs, and randomly transformed versions of the same clip as a positive pair.   

The local-local loss is defined by Equation~\ref{eq:ccont} and illustrated in Figure~\ref{fig:tcw}. A given video instance $i$ is divided into $N_{clips}$ non-overlapping clips. For the anchor clip starting at timestep $p$, its representation $G_{i,p}$, and the representation of its transformed version form the positive pair ($G_{i,p}$,$G'_{i,p})$ for this loss; whereas the other $N_{clips} - 1$ clips from the same video instance (and their transformed versions) form the negative pairs. Hence, for every positive pair, the local-local contrastive loss has $2 \times N_{clips} - 2$ negative pairs as defined in the following loss:

\begin{equation}
\label{eq:ccont}
\small
    \mathcal{L}_{LL}^{i}=-\sum_{p=1}^{N_{clips}}\log\frac{h\left(G_{i,p} , G'_{i,p}\right)}{\sum_{q=1}^{N_{clips}}[\mathbb{1}_{[q\neq p]} h(G_{i,p} , G_{i,q})+h(G_{i,p} , G'_{i,q})]}.
\end{equation}

The key difference between the Instance contrastive loss (Equation~\ref{eq:IC}) and the proposed Local-Local Temporal contrastive loss (Equation~\ref{eq:ccont}) is that for the local-local loss the negatives come from the same video instance but from a different temporal segment (clips), whereas in Equation~\ref{eq:IC}, the negative pairs come from different video instances.

\subsubsection{Global-Local Temporal Contrastive Loss}
\label{sec:pcl}

\begin{figure}[h]
\centering
  \includegraphics[width=0.95\linewidth]{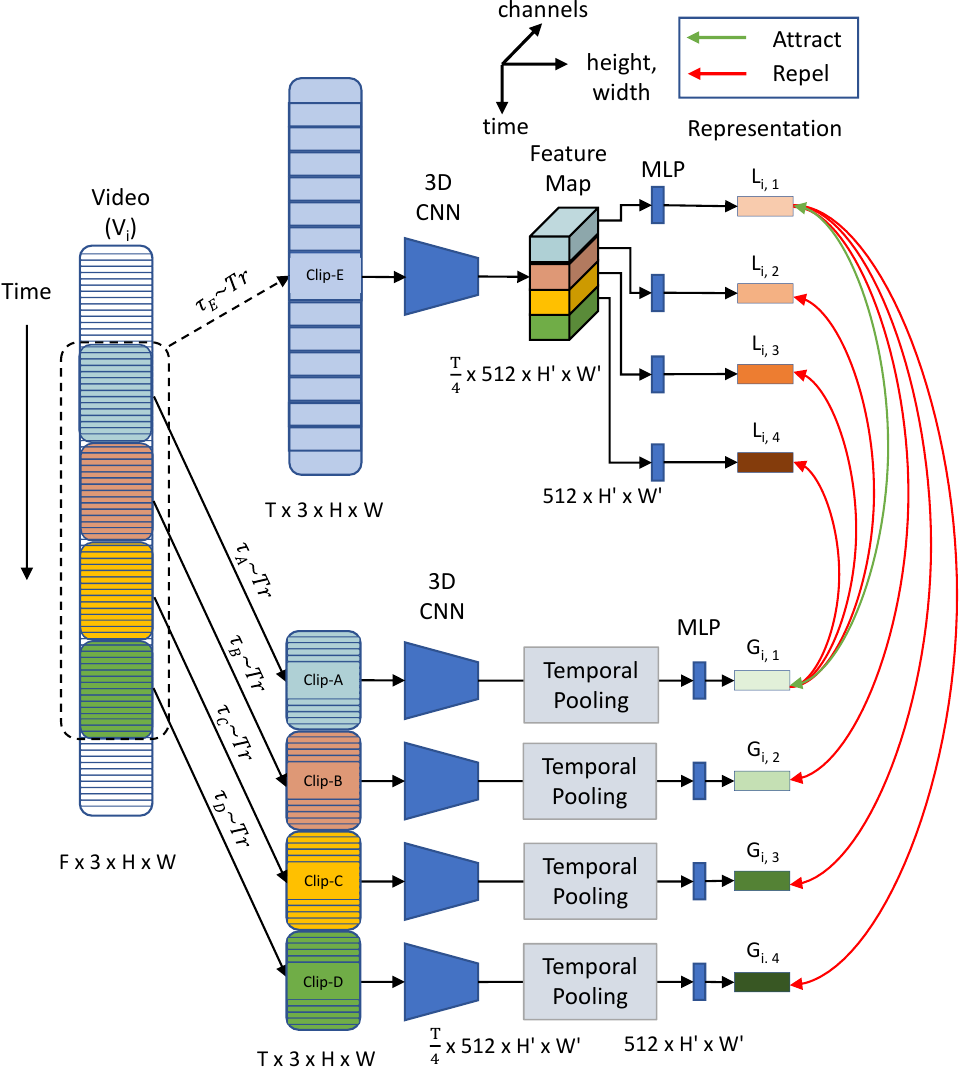}
 \vspace{-3mm}
 \caption{\textbf{Global-Local Temporal Contrastive Loss} A global clip (Clip E) is extracted from a video instance and divided into 4 equal length local clips (Clips A through D). The global clip is temporally downsampled to have the same number of frames as each local clip. The local representations $L_{i,1}$ through $L_{i,4}$ from the global clip are obtained from the penultimate layer of the 3D-CNN (prior to temporal pooling). Global representations of the local clips, $G_{i,1}$ through $G_{i,4}$ are obtained from the CNN (after temporal pooling layer). This loss aims to maximize the similarity between the local representation of the global clip and the global representations of the corresponding local clip. Further details in Section~\ref{sec:pcl}.}
\label{fig:tcp}
\end{figure}

The feature map in the higher layers of 3D CNNs are capable of representing temporal variation in the input clip, which is temporally pooled before being used for classification, or projected in the representation space in the case of contrastive learning. The objective of our proposed global-local temporal contrastive loss is to explicitly encourage the model to learn feature maps that represent the temporal locality of the input clip across temporal dimension of the feature map.

This loss is illustrated in Figure~\ref{fig:tcp}. The notion of \textit{local} and \textit{global} is used at two different levels: at the input clip level and the feature level. Clip-E is a global clip and Clip A-D are local clips contained within Clip-A. Features are referred to as \textit{global} after the final pooling operation and a temporal slice of the feature map before the pooling operation is referred to as a \textit{local} feature. In Figure~\ref{fig:tcp}, $L_{i,1}$ is the local feature of the global Clip-A and $G_{i,1}$ is the global feature of the local Clip-B.

For a video instance $i$, divided into $N_{clips}$ clips, the local clip $k$ can either be represented by a global (pooled) representation $G_{i,k}$ or a local representation $L_{i,k}$ of the corresponding timestep in the feature map of the global clip. This loss has two sets of reciprocal terms, with $G_{i,k}$ and $L_{i,k}$ serving as the anchor for each term. The negative pairs are supplied by matching the anchors with representations corresponding to other non-overlapping local clips. Note that similar to our local-local temporal contrastive loss we do not use negatives from other video instances for calculating this loss. The loss is defined by the following equations: %

\begin{equation}\label{eq:tcp}
 \small
    \mathcal{L}_{GL_{k}}^{i}= \log\frac{h\left(L_{i,k} , G_{i,k}\right)}{\sum_{q=1}^{N_{clips}} h(L_{i,k}, G_{i,q})} + \log\frac{h\left(G_{i,k}, L_{i,k}\right)}{\sum_{q=1}^{N_{clips}} h(G_{i,k}, L_{i,q})}, 
\end{equation}
\begin{equation}\label{eq:tcp1}
 \small
    \mathcal{L}_{GL}^{i}= -\sum_{k=1}^{N_{clips}}\mathcal{L}_{GL_{k}}^{i}.
\end{equation}

\section{Experiments}
\noindent {\bf Datasets and Implementation:}\label{para:dataset} We use three action recognition datasets: UCF101~\cite{ucf101}, Kinetics400~\cite{kinetics}, and HMDB51~\cite{hmdb} for our experiments. We use the three most commonly used networks from the literature: 3D-ResNet-18 (R3D-18)~\cite{kenshohara}, R(2+1)D-18~\cite{r2plus1d}, and C3D~\cite{c3d} %
for our experiments. For non-linear projection head, we use a multi-layer perceptron with 1-hidden layer following experimental setting of \cite{simclr}. We utilize 4 local clips per global clip for the global-local temporal contrastive loss.  For all reported results, we utilize commonly used random augmentations including appearance-based transforms such as grayscale, channel dropping, and color jittering and geometry-based transforms like random scaling, random cropping, random cut-out and random horizontal-flip. Our results can be further improved by using more complex augmentations like Gaussian blurring, shearing and rotation, however these are not used in the results reported in this paper. We provide results with more complex augmentation in Section D of the supplementary material. For self-supervised pretraining we use UCF101 training set (split-1) or Kinetics400 training set, without using any class labels. For all self-supervised pretraining, supervised finetuning and other downstream tasks, we use clips of 16 frames with a resolution of $112 \times 112$. More implementation details can be found in Section C of the supplementary material.

\subsection{Evaluating Self-supervised Representations}

 We evaluate the learned video representation using different downstream video understanding tasks: i) action recognition and ii) nearest neighbor video retrieval on UCF101 and HMDB51 datasets, and iii) limited label training on UCF101, following protocols from prior works \cite{memdpc}. We also evaluate our learned representations on the challenging Diving-48 fine-grained action recognition task~\cite{diving}; to the best of our knowledge TCLR is the first work that reports result on this challenging task. Our method is also employed in Knights~\cite{dave2021knights} to get first place in ICCV-21 Action recognition challenge~\cite{lengyel2022vipriors}.

\noindent {\bf Action Recognition on UCF101 and HMDB51: }

\begin{table*}
\begin{center}
\small
\vspace{-5em}
\begin{tabular}{ccccccc}
\hline

\hline

\hline\\[-3mm]
 \multicolumn{1}{l}{\multirow{2}{*}{\textbf{Method}}} & 
 \multicolumn{1}{l}{\multirow{2}{*}{\textbf{Venue}}} &
 \multicolumn{1}{c}{\multirow{2}{*}{\textbf{Input Size}}} & \multicolumn{2}{c}{\textbf{UCF101 Pre-Training}}  & \multicolumn{2}{c}{\textbf{Kinetics400 Pre-Training}} \\
 & & & UCF101 & HMDB51 & UCF101 & HMDB51
 \\[-3mm]
\\
 \hline

\hline

\hline
 \multicolumn{7}{c}{\textbf{Backbone: R3D-18}}\\
 \hline
ST-Puzzle~\cite{aaai19} & AAAI-19 & $16\times112$ & $-$ & $-$ & $65.8$ & $33.7$\\
STS~\cite{statistics2} & TPAMI-21 & $16\times112$ & $67.2$ & $32.7$ & $68.1$ & $34.4$\\
DPC~\cite{dpc} & ICCVw-19 & $40\times128$ & $60.6$&$-$ & $68.2$&$34.5$ \\
VCOP~\cite{vcop} & CVPR-20 & $16\times112$ & $64.9$&$29.5$  & $-$&$-$ \\
Pace Pred~\cite{pace_pred} & ECCV-20 & $16\times112$ & $65.0$&$-$ & $-$&$-$ \\
VCP~\cite{vcp} & AAAI-20 & $16\times112$ & $66.0$&$31.5$  & $-$&$-$ \\
PRP~\cite{prp} & CVPR-20 & $16\times112$ & $66.5$&$29.7$  & $-$&$-$     \\
Var. PSP~\cite{cho2020self} & Access-21 & $16\times112$ & $69.0$&$33.7$ & $-$&$-$ \\
MemDPC~\cite{memdpc} & ECCV-20 & $40\times224$ & $69.2$&$-$ & $-$&$-$ \\
TCP~\cite{wacv20} & WACV-21 & $-\times224$ & $64.8$&$34.7$  & $70.5$&$41.1$ \\
VIE~\cite{vie} & CVPR-20 & $16\times112$ & $-$ & $-$ & $72.3$ & $44.8$  \\
UnsupIDT~\cite{idt} & ECCVw-20 & $16\times112$ & $-$ & $-$ & $73.0$   & $41.6$  \\
CSJ~\cite{csj} & - & $16\times224$ & $70.4$  & $36.0$ & $76.2$ & $46.7$  \\

BFP~\cite{bfp} & WACV-21 & $40\times128$ & $63.6$ & $-$ & $66.4$ & $45.3$\\
IIC (RGB)~\cite{iic} & ACMMM-20 & $16\times112$ & $61.6$ & $-$ & $-$ & $-$\\
CVRL (Reproduced)~\cite{cvrl} & CVPR-21 & $16\times112$ & $75.77$   & $44.6$  & $-$& $-$   \\
SSTL~\cite{iclr21submitted_st_invariant} & - & $16\times112$ & $-$ & $-$ & $79.1$ & \textcolor{blue}{$49.7$} \\
VTHCL~\cite{vtc} & - & $8\times224$ & $-$ & $-$ & $80.6$ & $48.6$ \\
VideoMoCo~\cite{videomoco} & CVPR-21 & $16\times112$ & $-$ & $-$ & $74.1$ & $43.6$\\
RSPNet~\cite{rspnet} & AAAI-21 & $16\times112$ & $-$ & $-$ & $74.3$ & $41.8$\\
Temp Trans~\cite{simon}& ECCV-20 & $16\times112$ & \textcolor{blue}{$77.3$}  & \textcolor{blue}{$47.5$}  & $79.3*$ & $49.8*$ \\
TaCo~\cite{taco} & - & $16\times224$ & $-$ & $-$ & $\textcolor{blue}{81.4}$ & $45.4$\\
MFO~\cite{iccv21qian} & ICCV-21 & $16\times112$ & $-$ & $-$ & $79.1$ & $47.6$\\

\hline
\textbf{TCLR} & & $16\times112$ & \textcolor{red}{$82.4$}  & \textcolor{red}{$52.9$}  & \textcolor{red}{$84.1$}  & \textcolor{red}{$53.6$}  \\\hline
\textbf{TCLR (Best Ablation)} & & $16\times112$ & \textcolor{red}{$83.9$}  & \textcolor{red}{$53.5$}  & \textcolor{red}{$85.4$}  & \textcolor{red}{$55.4$}  \\
\hline
 \multicolumn{7}{c}{\textbf{Backbone: R(2+1)D-18}}\\
 \hline
VCP~\cite{vcp} & AAAI-20 & $16\times112$ & $66.3$  & $32.2$  & $-$ & $-$  \\
PRP~\cite{prp} & CVPR-20 & $16\times112$ & $72.1$  & $35.0$  & $-$ & $-$  \\
VCOP~\cite{vcop} & CVPR-20 & $16\times112$ & $72.4$  & $30.9$  & $-$     & $-$      \\
Pace Pred~\cite{pace_pred} & ECCV-20 & $16\times112$ & $75.9$  & $35.9$  & $77.1$  & $36.6$   \\
STS~\cite{statistics2} & TPAMI-21 & $16\times112$ & $73.6$  & $34.1$  & $77.8$     & $40.5$  \\
VideoMoCo~\cite{videomoco} & CVPR-21 & $16\times112$ & $-$ & $-$ & $78.7$ & $\textcolor{blue}{49.2}$\\
VideoDIM~\cite{vdim} & - & $32\times128$ & $-$ & $-$ & $79.7*$ & $49.2*$  \\
RSPNet~\cite{rspnet} & AAAI-21 & $16\times112$ & $-$ & $-$ & $81.1$ & $44.6$\\
Temp Trans~\cite{simon} & ECCV-20 & $16\times112$ & \textcolor{blue}{$81.6$}  & \textcolor{blue}{$46.4$}  & $-$ & $-$  \\
TaCo~\cite{taco} & - & $16\times224$ & $-$ & $-$ & $\textcolor{blue}{81.8}$ & $46.0$\\
\hline
{\bf TCLR }&  & $16\times112$ & \textcolor{red}{$82.8$}     & \textcolor{red}{$53.6$} & \textcolor{red}{$88.2$} & \textcolor{red}{$60.0$}      \\\hline

\hline
 \multicolumn{7}{c}{\textbf{Backbone: C3D}}\\
 \hline
MA Stats-1~\cite{statistics_cvpr19} & CVPR-19 & $16\times112$ & $58.8$  & $32.6$  & $61.2$ & $33.4$  \\
Temp Trans~\cite{simon} & ECCV-20 & $16\times112$ & $68.3$  & $38.4$  & $69.9*$ & $39.6*$  \\
PRP~\cite{prp} & CVPR-20 & $16\times112$& $69.1$  & $34.5$  & $-$    & $-$     \\
VCP~\cite{vcp} & AAAI-20 & $16\times112$& $68.5$  & $32.5$  & $-$    & $-$     \\
VCOP~\cite{vcop} & CVPR-20 & $16\times112$ & $65.6$  & $28.4$  & $-$    & $-$     \\
Pace Pred~\cite{pace_pred} & ECCV-20 & $16\times112$& $68.0$  & $-$  & $-$  & $-$   \\
STS~\cite{statistics2} & TPAMI-21 & $16\times112$& $69.3$  & $34.2$  & $71.8$  & $37.8$ \\
Var. PSP~\cite{cho2020self} & Access-21 & $16\times112$ & \textcolor{blue}{$70.4$} & $34.3$ & $-$  & $-$     \\
DSM~\cite{scene-motion} & AAAI-21 & $16\times112$ & $70.3$ & $\textcolor{blue}{40.5}$ & $-$ & $-$\\
\hline
{\bf TCLR} & & $16\times112$ & \textcolor{red}{$76.1$}     & \textcolor{red}{$48.6$}  & $-$& $-$   \\\hline

\hline
\multicolumn{6}{c}{\textbf{Other Configurations}}\\
\hline
\grayedout{CVRL (R3D-50)~\cite{cvrl}} & \grayedout{CVPR-21} & \grayedout{$32\times224$} & $-$     & $-$  & \grayedout{$92.2$} & \grayedout{$66.7$}   \\
\grayedout{RSPNet (S3D-G)~\cite{rspnet}} & \grayedout{AAAI-21} & \grayedout{$64\times224$} & $-$     & $-$  & \grayedout{$93.7$} & \grayedout{$64.7$}   \\
\grayedout{CoCLR\textsuperscript{\textdagger} (S3D-23)~\cite{cotraining}} & \grayedout{NeurIPS-20} & \grayedout{$16\times112$} & \grayedout{$87.3$} & \grayedout{$58.7$} & \grayedout{$90.6$} & \grayedout{$62.9$}\\
\grayedout{SpeedNet (S3D-G)~\cite{speedNet}} & \grayedout{CVPR-20} & \grayedout{$16\times224$} & $-$ & $-$ & \grayedout{$81.1$} & \grayedout{$48.8$}\\
\grayedout{$\rho$SimCLR (R50)~\cite{Feichtenhofer_2021_CVPR}} & \grayedout{CVPR-21} & \grayedout{$8\times224$} & \grayedout{$-$} & \grayedout{$-$} & \grayedout{$85.6$} & \grayedout{$-$}\\
\grayedout{SeCO (R50+TSN)~\cite{seco}} & \grayedout{AAAI-21} & \grayedout{$50\times224$} & $-$ & $-$ & \grayedout{$88.3\ddagger$} & \grayedout{$55.6\ddagger$}\\
\hline
\end{tabular}
\end{center}
\caption{Finetuning Results (average of 3 splits) for action classification on UCF101 and HMDB51. Self supervised pretraining  was done on UCF101 (left) and Kinetics (right). \textsuperscript{\textdagger} indicates models that utilize optical flow. $*$ indicates Kinetics-600 self-supervised pretraining. $\ddagger$ indicates ImageNet+Kinetics pre-training. \textcolor{red}{Best} and \textcolor{blue}{second best} results are highlighted.}
\label{tab:finetuning-results}
\end{table*}

\begin{table*}
\footnotesize

\centering
 \begin{tabular}{cccccccccc}
\hline

\hline

\hline\\[-3mm]
\multirow{2}{*}{Method}   & Venue &  \multicolumn{4}{c}{\textbf{UCF101}}  & \multicolumn{4}{c}{\textbf{HMDB51}}  \\
                &  R@1   & R@5   & R@10  & R@20 &  R@1   & R@5   & R@10  & R@20    \\ 
\hline

\hline

\hline\\[-3mm]
\multicolumn{10}{c}{\textbf{Backbone: R3D-18}} \\
\hline
VCOP~\cite{vcop} & CVPR-20 & $14.1$  & $30.3$  & $40.4$  & $51.1$     & $~7.6$  & $22.9$  & $34.4$  & $48.8$     \\
VCP~\cite{vcp}   & AAAI-20 & $18.6$  & $33.6$  & $42.5$  & $53.5$     & $~7.6$  & $24.4$  & $36.6$  & $53.6$     \\
Pace Pred~\cite{pace_pred} & ECCV-20 & $23.8$  & $38.1$  & $46.4$  & $56.6$     & $~9.6$  & $26.9$  & $41.1$  & $56.1$     \\
Var. PSP~\cite{cho2020self}               & Access-21 & $24.6$  & $41.9$  & $51.3$  & $62.7$     & $10.3$  & $26.6$  & $38.8$  & $51.6$     \\
Temp Trans~\cite{simon} & ECCV-20 & $26.1$  & $48.5$  & $59.1$  & $69.6$     & $-$  & $-$  & $-$  & $-$     \\
STS\supstar~\cite{statistics2}              & TPAMI-21 & $38.3$  & $\textcolor{blue}{59.9}$  & $68.9$  & $77.2$     & $18.0$  & $37.2$  & $50.7$  & $\textcolor{blue}{64.8}$     \\
SSTL\supstar~\cite{iclr21submitted_st_invariant}  & - & \textcolor{blue}{$44.5$}  & $57.4$  & $63.5$  & $70.0$     & \textcolor{blue}{$21.8$}  & $35.7$  & $44.2$  & $57.7$     \\
CSJ\supstar~\cite{csj}                                        & - & $21.5$     & $40.5$     & $53.2$     & $64.9$        & $-$     & $-$     & $-$     & $-$        \\
MemDPC~\cite{memdpc} & ECCV-20 & $20.2$  & $40.4$  & $52.4$  & $64.7$     & $~7.7$  & $25.7$  & $40.6$  & $57.7$     \\
RSPNet~\cite{rspnet}                                     & AAAI-21 & $41.1$  & $59.4$  & $68.4$  & $77.8$   & $-$  & $-$  & $-$  & $-$   \\
MFO~\cite{iccv21qian}                                     & ICCV-21 & $39.6$  & $57.6$  & $\textcolor{blue}{69.2}$  & $\textcolor{blue}{78.0}$   & $18.8$  & $\textcolor{blue}{39.2}$  & $\textcolor{blue}{51.0}$  & $63.7$   \\
\hline
\textbf{TCLR}                                        & - & \textcolor{red}{$56.2$} & \textcolor{red}{$72.2$} & \textcolor{red}{$79.0$} & \textcolor{red}{$85.3$}  & \textcolor{red}{$22.8$} & \textcolor{red}{$45.4$} & \textcolor{red}{$57.8$} & \textcolor{red}{$73.1$}  \\
\hline 

\hline
\multicolumn{10}{c}{\textbf{Backbone: C3D}} \\
\hline 
VCOP~\cite{vcop}                                     & CVPR-20 & $12.5$  & $29.0$  & $39.0$  & $50.6$   & $5.7$  & $19.5$  & $30.7$  & $45.8$   \\
VCP~\cite{vcp}                                       & AAAI-20 & $17.3$  & $31.5$  & $42.0$  & $52.6$   & $7.8$  & $23.8$  & $35.3$  & $49.3$   \\
Pace Pred~\cite{pace_pred}                           & ECCV-20 & $31.9$  & $49.7$  & $59.2$  & $68.9$   & $12.5$  & $32.2$  & $45.4$  & $61.0$   \\
DSM~\cite{pace_pred}                                   & AAAI-21 & $16.8$  & $33.4$  & $43.4$  & $54.6$     & $8.2$  & $25.9$  & $38.1$  & $52.0$     \\
STS\supstar~\cite{statistics2}      & TPAMI-21 & $\textcolor{blue}{39.1}$  & $\textcolor{blue}{59.2}$  & $\textcolor{blue}{68.8}$  & $\textcolor{blue}{77.6}$   & \textcolor{blue}{$16.4$} & \textcolor{blue}{$\textcolor{blue}{36.9}$}  & $\textcolor{blue}{49.9}$  & $\textcolor{blue}{64.9}$   \\
RSPNet~\cite{rspnet}                                     & AAAI-21 & $36.0$  & $56.7$  & $66.5$  & $76.3$   & $-$  & $-$  & $-$  & $-$   \\
\hline
\textbf{TCLR}                                        & - & \textcolor{red}{$48.6$} & \textcolor{red}{$67.6$} & \textcolor{red}{$75.5$} & \textcolor{red}{$82.5$}  & \textcolor{red}{$19.3$} & \textcolor{red}{$43.3$} & \textcolor{red}{$57.6$} &  \textcolor{red}{$70.1$}  \\

\hline

\hline
\multicolumn{10}{c}{\textbf{Backbone: R(2+1)D-18}} \\
\hline 
VCOP~\cite{vcop}                                     & CVPR-20 & $10.7$  & $25.9$  & $35.4$  & $47.3$   & $ 7.4$  & $22.6$  & $34.4$  & $48.5$   \\
VCP~\cite{vcp}                                       & AAAI-20 & $19.9$  & $33.7$  & $42.0$  & $50.5$   & $6.7$  & $21.3$  & $32.7$  & $49.2$   \\
Pace Pred~\cite{pace_pred}                           & ECCV-20 & $25.6$  & $42.7$  & $51.3$  & $61.3$   & $12.9$  & $31.6$  & $43.2$  & $58.0$   \\
STS\supstar~\cite{statistics2}      & TPAMI-21 & \textcolor{blue}{$38.1$}  & \textcolor{blue}{$58.9$}  & \textcolor{blue}{$68.1$}  & \textcolor{blue}{$77.0$}   & \textcolor{blue}{$16.4$}  & \textcolor{blue}{$36.9$}  & \textcolor{blue}{$50.5$}  & \textcolor{blue}{$65.4$}   \\
\hline
\textbf{TCLR}               & - & \textcolor{red}{$56.9$} & \textcolor{red}{$72.2$}  & \textcolor{red}{$79.0$}  & \textcolor{red}{$84.6$}   & \textcolor{red}{$24.1$} & \textcolor{red}{$45.8$}  & \textcolor{red}{$58.3$}  & \textcolor{red}{$75.3$}   \\
\hline

\hline

\hline\\[-3mm]
\end{tabular}
\caption{\textbf{Nearest  neighbor video retrieval} results on UCF101 and HMDB51, after self-supervised pretraining on UCF101. \supstar marks models pretrained on Kinetics-400. \textcolor{red}{Best} and \textcolor{blue}{second best} results highlighted. Methods based on optical flow and audio modalities are excluded.} 
\label{tab:nn-results} 
\end{table*}

For the action recognition task on UCF101 and HMDB51, we first pretrain different video encoders in self-supervised manner on UCF101 or Kinetics400, and then perform supervised fine-tuning. In order to ensure fair comparison, we evaluate the method on the three most commonly used 3D CNN backbones in the prior works, while also listing details about the \textit{input clip resolution} and \textit{number of frames} used, as it is known to affect the results significantly~\cite{r2plus1d, gdt}. Comparison results are shown in Table~\ref{tab:finetuning-results}. Previous results based on multi-modal approaches that utilize text, audio etc are excluded~\cite{gdt, xdc, milnce}. Results from prior works which do not utilize the three common architectures or use optical flow as input are presented in grey. We reproduce the results for CVRL~\cite{cvrl} using the R3D-18 model and 112 resolution, and carefully implement their temporal sampling and augmentation strategy. TCLR consistently outperforms the state-of-art by wide margins for all comparable combinations of backbone, pre-training dataset and fine-tuning dataset. The best prior results are reported by TaCo~\cite{taco}, which relies on learning temporal features using pretext tasks on top of instance discrimination. Our consistent improvement over TaCo suggests that using temporal contrastive losses results in better features than using existing temporal pre-text tasks in a multi-task setting.

\noindent {\bf Nearest Neighbor Video Retrieval:}

We evaluate the learned representation by performing nearest neighbor retrieval after self-supervised pretraining on UCF101 videos and without any supervised finetuning. Videos from the test set are used as the query and the training set as the search gallery, following the protocol used in prior work~\cite{memdpc}. Results for retrieval are presented for both UCF101 and HMDB51 in Table~\ref{tab:nn-results}. {\em TCLR outperforms previous state-of-the-art in UCF101 Top-1 Retrieval by 12\% to 30\% depending on the architecture}

\begin{figure}[h]
\begin{center}
 \includegraphics[width=1.0\linewidth, trim = {0 1cm 0 1cm}]{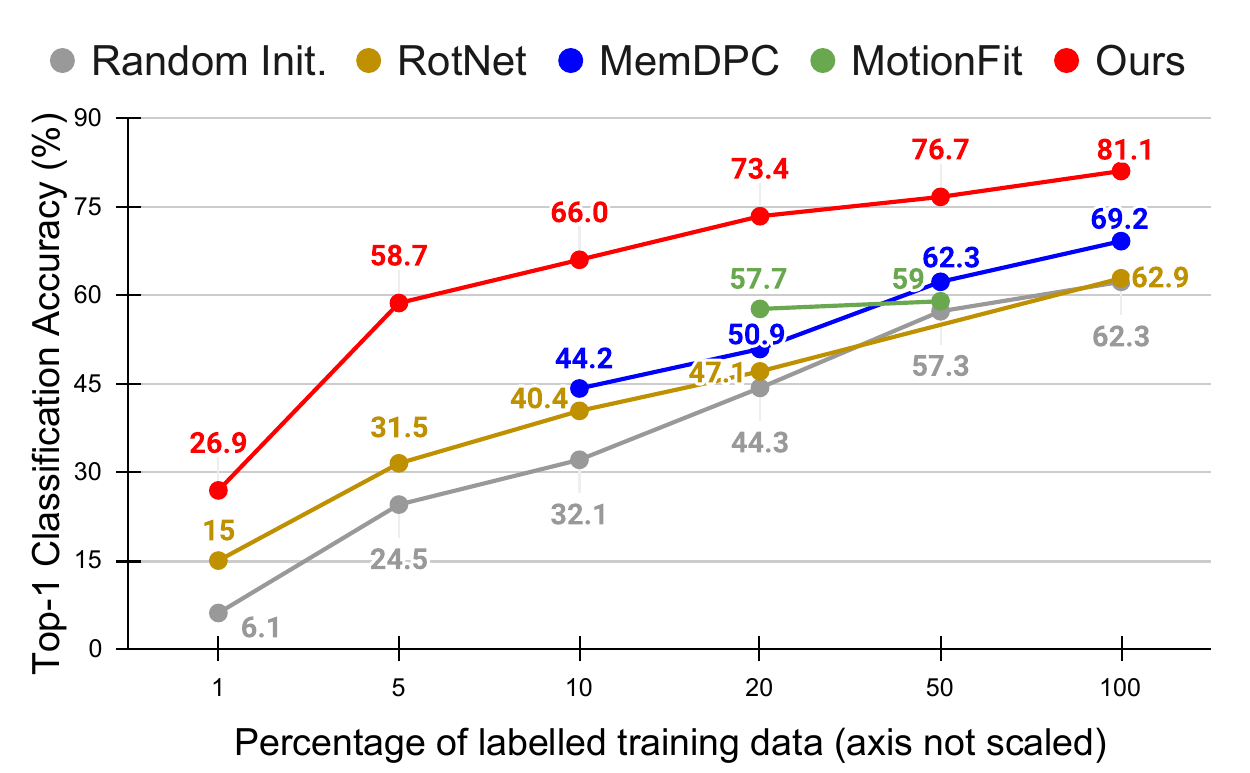}
\end{center}
\vspace{-2mm}
   \caption{\textbf{Evaluating Label Efficiency} using Limited Label Learning on UCF101 (split-1) action classification task.}
\label{fig:few-shot}
\end{figure}

\noindent{\bf Label Efficiency/ Finetuning with limited data:} We evaluate our pretrained model for action recognition task on UCF101 (split-1) with limited labeled training data following the protocols from prior work~\cite{memdpc, 3drotnet, motionfit}. Our method outperforms MotionFit~\cite{motionfit}, MemDPC~\cite{memdpc} and RotNet3D~\cite{3drotnet} in all settings of limited percentage of training data as shown in Figure~\ref{fig:few-shot}. This result in addition to NN results demonstrate that the learned representations from TCLR are significantly better than other recent works, \textit{TCLR can achieve competitive performance to MemDPC with only 10\% of the labeled data}.

\vspace{1mm}
\noindent{\bf Experiments on Diving-48 Dataset:} This task presents some additional challenges over and above the standard action recognition task: action categories in Diving48 are defined by a combination of takeoff (dive groups), movements in flight (somersaults and/or twists), and entry (dive positions) stages. Two otherwise identical categories may only have fine grained differences limited to only one of the three stages. This makes Diving48 useful for evaluating the fine-grained representation capabilities of the model, which are not well tested by action recognition tasks on common benchmark datasets like UCF101 and HMDB51. Our proposed evaluation protocol consists of self-supervised pretraining followed by supervised finetuning on the Diving48-Train set. We adopt the 3D ResNet-18 architecture, with input resolution and clip length fixed at $112\times112$ and 16 frames, respectively.

\begin{table}[h]
\vspace{-2mm}
 
\small
\centering
\begin{tabular}{lc}
\hline

\hline

\hline\\[-3mm]
\textbf{Pre-Training}  & \textbf{Accuracy}  \\
\hline

\hline

\hline\\[-3mm]
None  (Random Initialization)  & $13.4$ \\
MiniKinetics Supervised~\cite{choi} & $18.0$ \\
Instance Contrastive & $15.8$     \\
VCOP~\cite{vcop} & $14.7$ \\
CVRL~\cite{cvrl} & $17.6$ \\ \hline
TCLR & $22.9$     \\\hline

\hline

\hline\\[-3mm]

\end{tabular}
\vspace{-2mm}
\caption{\label{tab:diving-results} Diving48 fine-grained action classification results.}
\vspace{-4mm}
\end{table}

 Results are summarized in Table~\ref{tab:diving-results}. TCLR pretraining on Diving48 without extra data outperforms random initialization and MiniKinetics~\cite{minikinetics} supervised pretraining. The within-instance temporal discrimination losses in TCLR help it outperform the instance contrastive loss. This is due to TCLR learning features to represent fine-grained differences between parts of diving video instances.

\vspace{-2mm}
\subsection{Ablation Study}
\vspace{-4mm}
\begin{table}[H]
\small
\centering
\begingroup
\setlength{\tabcolsep}{3pt} %
\renewcommand{\arraystretch}{1.0} %
\resizebox{\columnwidth}{!}{
\begin{tabular}{@{\hskip0pt}ccccccc@{\hskip0pt}}
\hline

\hline 

\hline\\[-3mm]

 \multicolumn{3}{c}{\multirow{2}{*}{\textbf{Contrastive Losses}}} &
 \multicolumn{3}{c}{\textbf{Classification Top1 Acc.}} &
\textbf{Retrieval} \\

 & & & \textbf{Linear Eval} & \textbf{Finetune} & \textbf{Transfer} & R@1 \\
 $\mathcal{L}_{IC}$ & $\mathcal{L}_{LL}$ & $\mathcal{L}_{GL}$ & UCF101 & UCF101 & HMDB51 & UCF101 \\
 \hline

\hline 

\hline\\[-3mm]
\multicolumn{3}{c}{Random Init.} & $17.15$ & $62.39$ & $26.95$ & $8.21$ \\
\xmark & \cmark & \xmark & $21.58$ & $68.42$ & $-$ & $13.66$ \\
\xmark & \xmark & \cmark & $20.61$ & $70.19$ & $-$ & $12.83$ \\
\xmark & \cmark & \cmark & $23.39$ & $74.29$ & $47.35$ & $14.17$ \\
\hline
\cmark & \xmark & \xmark & $54.58$ & $71.31$ & $38.32$ & $40.76$ \\
\cmark & \cmark & \xmark & $62.70$ \increase{~~+8\%} & $77.70$\increase{~~+6\%} & $49.77$\increase{+11\%} & $51.10$\increase{+10\%} \\
\cmark & \xmark & \cmark & $64.55$ \increase{+10\%}  & $76.30$\increase{~~+5\%} & $47.87$\increase{+10\%} & $47.32$\increase{~~+7\%} \\
\cmark & \cmark & \cmark & $\mathbf{69.91}$ \increase{+15\%}  & $\mathbf{82.40}$\increase{+11\%}  & $\mathbf{52.80}$\increase{+14\%} & $\mathbf{56.17}$\increase{+15\%} \\
 \hline

\hline 

\hline\\[-3mm]
\end{tabular}}
\endgroup
\vspace{-2mm}
\caption{Ablation study of the \textbf{impact of temporal contrastive losses} on downstream tasks. \textcolor{htmlcssgreen}{Green} indicates improvements over instance contrastive baseline.}
\label{tab:lossabl}
\vspace{-3mm}
\end{table}

In order to study the impact of each contrastive loss used in TCLR, we test R3D-18 models pre-trained on UCF101 videos with a subset of the losses on each downstream task. The results for linear evaluation, full fine-tuning, transfer learning to HMDB51 and nearest neighbour retrieval are shown in Table~\ref{tab:lossabl}. Addition of each temporal contrastive loss ($\mathcal{L}_{LL}$ \& $\mathcal{L}_{GL}$) leads to significant gains over instance contrastive and random initialization baselines, with the best results coming from combined use of all losses. We verify the correctness of our baselines by comparing them with similar results reported in prior work. Details can be found in Section E of the supplementary material. One interesting observation is that purely temporal contrastive learning, without instance discrimination, does not learn strong features directly (as can be seen from results on linear evaluation and NN-Retrieval), but it provides an useful initialization prior for supervised finetuning experiments.

\begin{figure*}
    \centering
    \subfloat[Scratch]{{\includegraphics[width=0.32\linewidth]{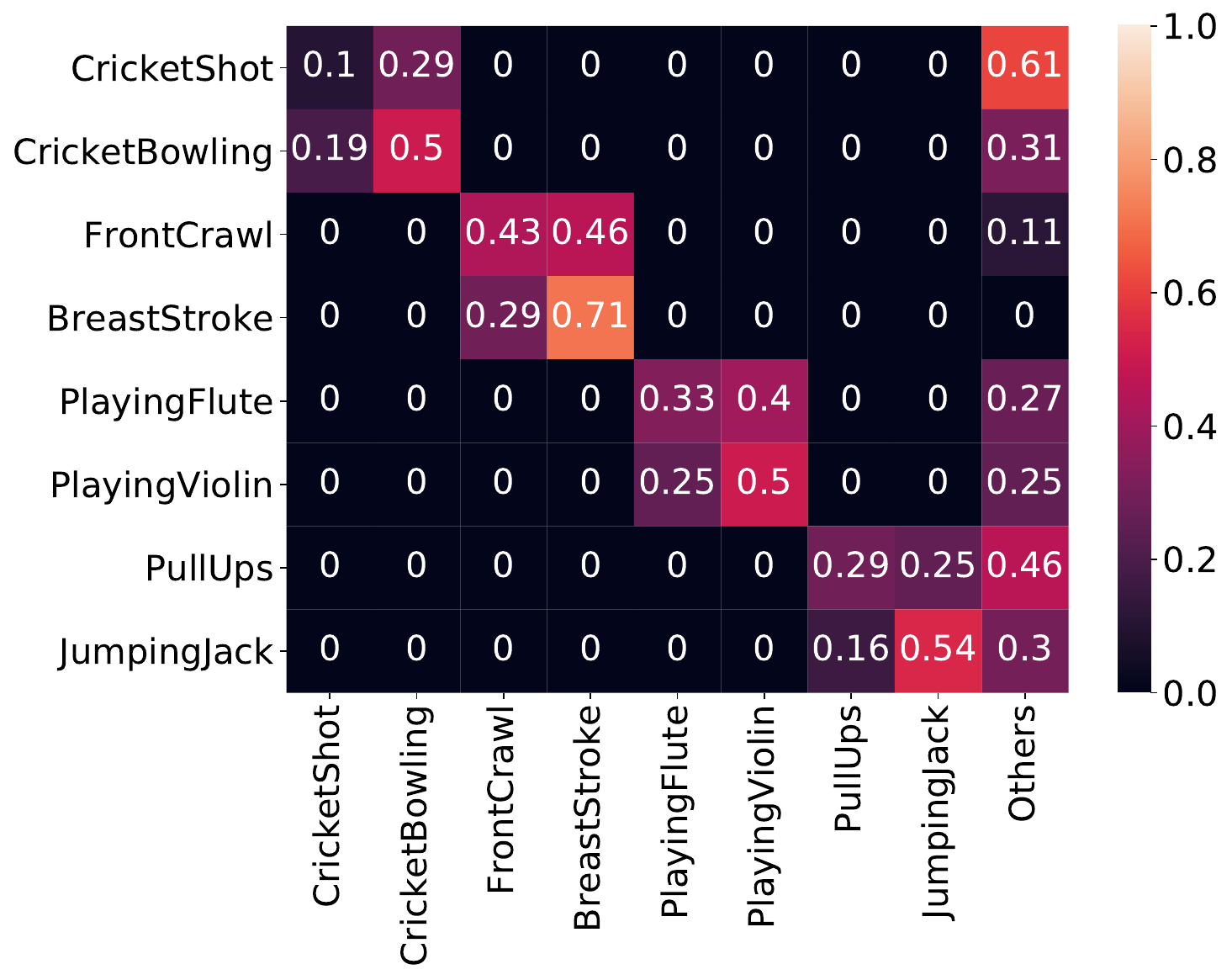}{\label{fig:scratch_conf}} }}\hspace*{-0.4em}
    \subfloat[IC pretraining]{{\includegraphics[width=0.32\linewidth]{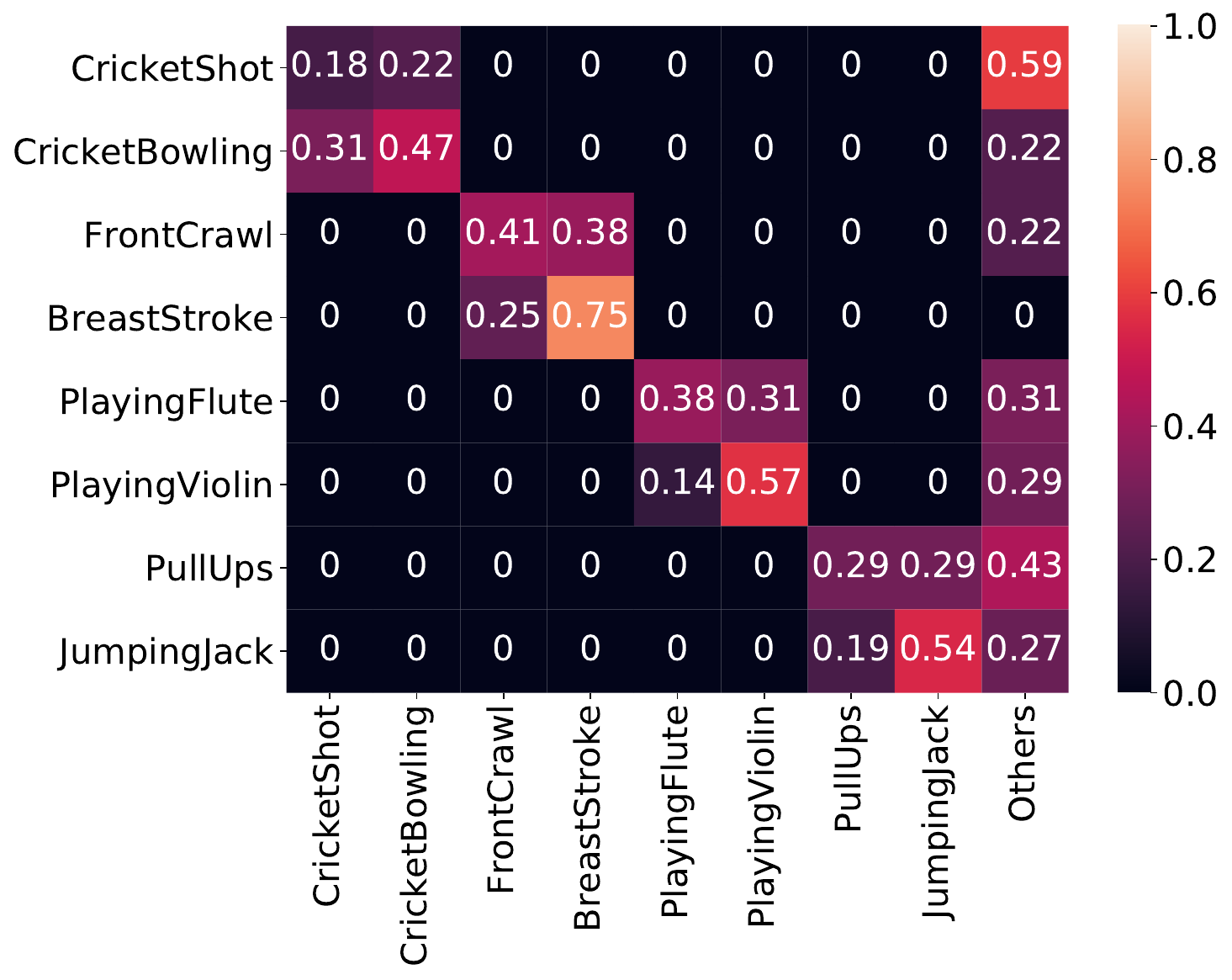}{\label{fig:ic_conf}} }}\hspace*{-0.4em}
    \subfloat[TCLR pretraining]{{\includegraphics[width=0.32\linewidth]{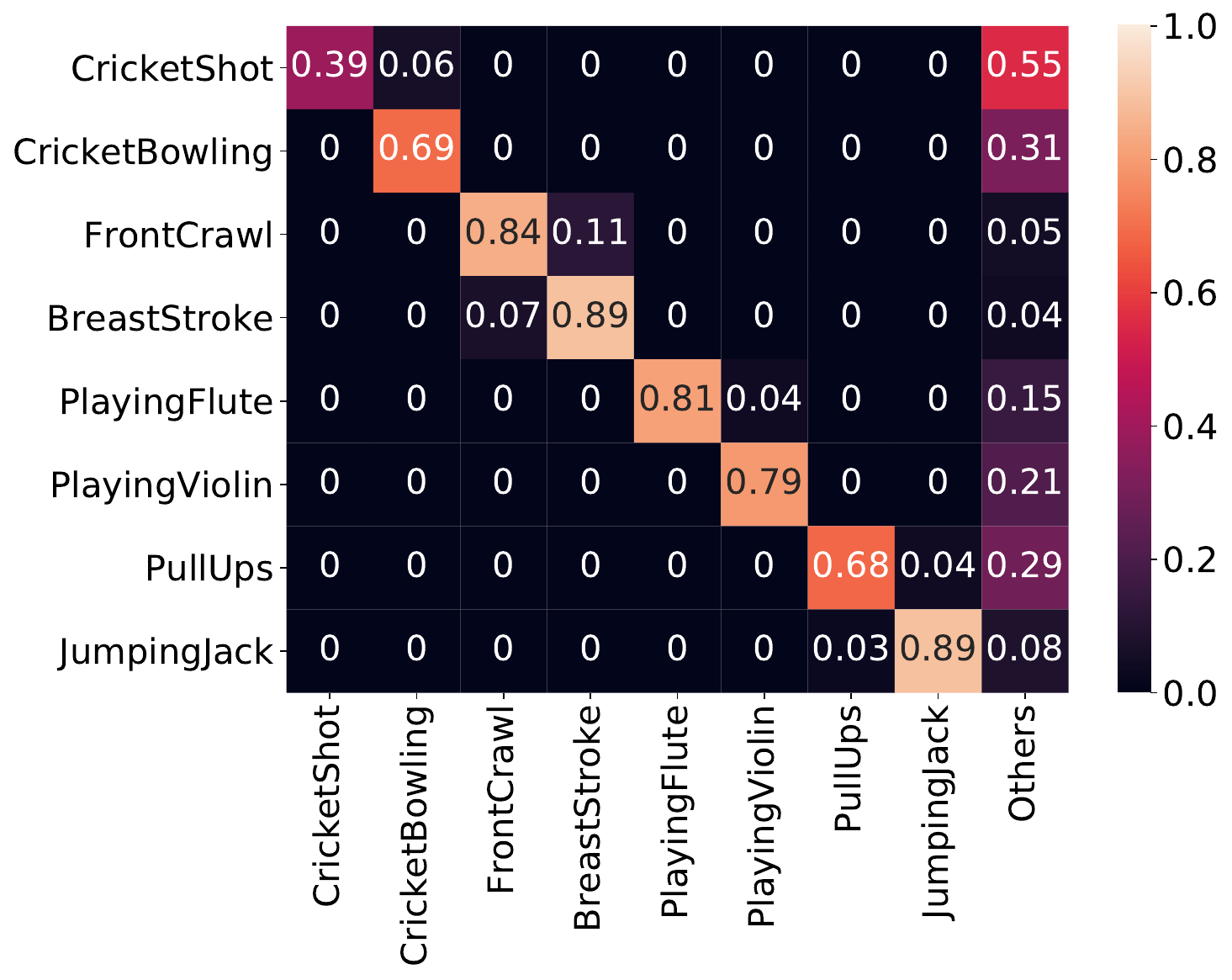}{\label{fig:tclr_conf}} }}\\ 
    \subfloat[Strongly Confused class pairs]{{\includegraphics[width=0.99\linewidth]{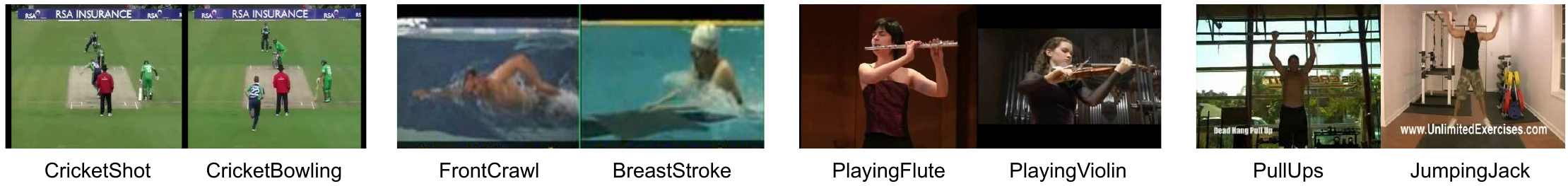}{\label{fig:class_pairs}} }}
    \vspace{-2mm}
    \caption{Confusion matrices for 4 highly confused class-pairs from UCF101 classification models with \textbf{(a)} no pretraining,  \textbf{(b)} IC pretraining, and  \textbf{(c)} TCLR pretraining. \textbf{(d)} Classes illustrated with a sample frame. TCLR significantly improves over IC in distinguishing visually similar classes. }
    \vspace{-6mm}
    \label{fig:conf_matrices}
    
\end{figure*}

\subsection{Temporal diversity helps video understanding}

\begin{figure}[H]
\begin{center}
 \includegraphics[width=1.0\linewidth]{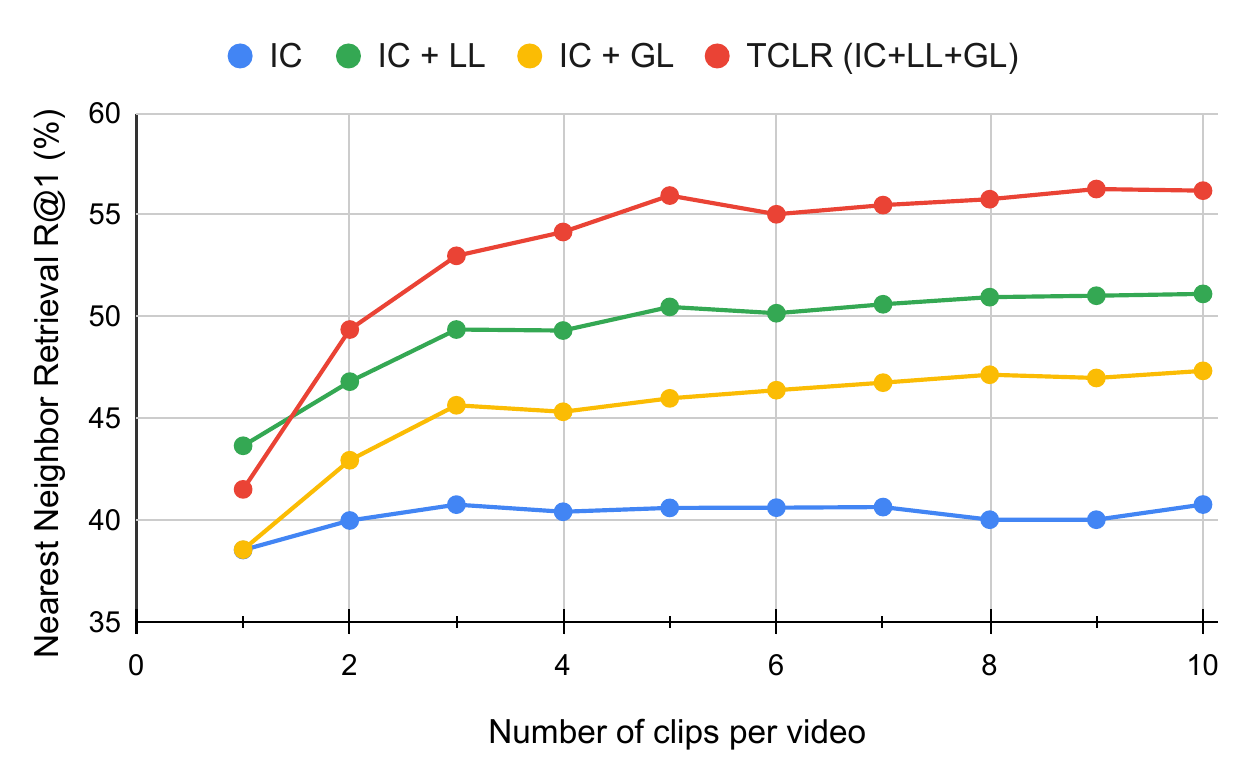}
\end{center}
\vspace{-7mm}
   \caption{Temporally distinct features learned by TCLR result in a significant improvement in NN-Retrieval on UCF101 (split-1) with increasing number of clips per video.}
\label{fig:nnr_clips}
\vspace{-2mm}
\end{figure}

To study the impact of temporal feature diversity directly, we utilize self-supervised pre-trained models only to avoid influence of supervised fine-tuning. As shown in Fig \ref{fig:nnr_clips}, increasing from 1 clip to 10 clips per video, we observe that the pretraining strategies using temporal contrastive losses get significant performance gains (about \textit{7-8\% for each individual loss and 14.67\% for TCLR}) with increasing number of clips. Instance Contrastive pretraining which enforces temporal invariance in learned features of video, does not see a similar improvement. It is also worth noting that each of the LL and GL losses help in learning different types of \textit{temporal diversity} which results in TCLR having bigger improvements relative to either of the temporal contrastive losses. Performance gains of a similar nature can also be observed in other downstream tasks as well, which are reported in Section F of the supplementary material.

\subsection{Distinguishing confusing class pairs}
To examine the ability of TCLR to distinguish confusing classes, we looked at the most confused action class pairs for UCF101 action recognition models trained from scratch. We observe that the these pairs mostly consist of fine-grained variants of action classes, for example the swimming actions \texttt{BreastStroke} and \texttt{FrontCrawl}.  Some such pairs of classes are visualized in Figure \ref{fig:conf_matrices}d. We can see that these classes are confusing because the corresponding frames are visually similar. In this study we considered a model without pretraining as a baseline, and tried to see the impact of instance contrastive and TCLR pretraining on it. We observe that despite a significant overall improvement in accuracy, instance contrastive pre-training does not provide any significant gain in distinguishing these confused class pairs over the scratch baseline. On the other hand, TCLR pre-training helps remarkably with the confused classes. Average recall for these 8 classes is 42.5\% for the scratch model, 44.9\% for the IC model and 74.8\% for the TCLR model. Since the classes are visually similar, distinguishing them requires learning the temporal variation in videos.

\vspace{-2mm}
\section{Conclusion}
\vspace{-2mm}

In this work, we propose two novel Temporal Contrastive losses to improve the quality of learned self-supervised video representations over standard instance discrimination contrastive learning. We provide extensive experimental evidence on three diverse datasets and obtain state-of-the-art results across various downstream video understanding tasks. The success of our approach underscores the benefits of contrastive learning beyond instance discrimination.

\section*{Acknowledgments}
Ishan Dave would like to acknowledge support from the Office of the Director of National Intelligence (ODNI), Intelligence Advanced Research Projects Activity (IARPA), via IARPA R\&D Contract No. D17PC00345. The views and conclusions contained herein are those of the authors and should not be interpreted as necessarily representing the official policies or endorsements, either expressed or implied, of the ODNI, IARPA, or the U.S. Government. The U.S. Government is authorized to reproduce and distribute reprints for Governmental purposes notwithstanding any copyright annotation thereon.

\bibliographystyle{ieee_fullname}
\bibliography{main}

\begin{thebibliography}{10}\itemsep=-1pt

\bibitem{afouras2020self}
Triantafyllos Afouras, Andrew Owens, Joon~Son Chung, and Andrew Zisserman.
\newblock Self-supervised learning of audio-visual objects from video.
\newblock In {\em The European Conference on Computer Vision}, 2020.

\bibitem{videojigsaw}
Unaiza Ahsan, Rishi Madhok, and Irfan Essa.
\newblock {\em Video Jigsaw: Unsupervised learning of spatiotemporal context
  for video action recognition}, pages 179--189.
\newblock 2019.

\bibitem{xdc}
Humam Alwassel, Dhruv Mahajan, Bruno Korbar, Lorenzo Torresani, Bernard Ghanem,
  and Du Tran.
\newblock {\em Self-Supervised Learning by Cross-Modal Audio-Video Clustering},
  pages 9758--9770.
\newblock 2020.

\bibitem{amdim}
Philip Bachman, R.~Devon Hjelm, and William Buchwalter.
\newblock Learning representations by maximizing mutual information across
  views.
\newblock In {\em Advances in Neural Information Processing Systems}, pages
  15535--15545, 2019.

\bibitem{taco}
Yutong Bai, Haoqi Fan, Ishan Misra, Ganesh Venkatesh, Yongyi Lu, Yuyin Zhou,
  Qihang Yu, Vikas Chandra, and Alan Yuille.
\newblock Can temporal information help with contrastive self-supervised
  learning?
\newblock {\em arXiv preprint arXiv:2011.13046}, 2020.

\bibitem{bfp}
Nadine Behrmann, Jurgen Gall, and Mehdi Noroozi.
\newblock Unsupervised video representation learning by bidirectional feature
  prediction.
\newblock In {\em Proceedings of the IEEE/CVF Winter Conference on Applications
  of Computer Vision}, pages 1670--1679, 2021.

\bibitem{speedNet}
Sagie Benaim, Ariel Ephrat, Oran Lang, Inbar Mosseri, William~T. Freeman,
  Michael Rubinstein, Michal Irani, and Tali Dekel.
\newblock Speednet: Learning the speediness in videos.
\newblock In {\em Proceedings of the IEEE/CVF Conference on Computer Vision and
  Pattern Recognition}, pages 9922--9931, 2020.

\bibitem{buchler2018improving}
Uta Buchler, Biagio Brattoli, and Bjorn Ommer.
\newblock Improving spatiotemporal self-supervision by deep reinforcement
  learning.
\newblock In {\em Proceedings of the European conference on computer vision
  (ECCV)}, pages 770--786, 2018.

\bibitem{swav}
Mathilde Caron, Ishan Misra, Julien Mairal, Priya Goyal, Piotr Bojanowski, and
  Armand Joulin.
\newblock {\em Unsupervised Learning of Visual Features by Contrasting Cluster
  Assignments}, pages 9912--9924.
\newblock 2020.

\bibitem{kinetics}
Joao Carreira and Andrew Zisserman.
\newblock Quo vadis, action recognition? a new model and the kinetics dataset.
\newblock In {\em Proceedings of the IEEE Conference on Computer Vision and
  Pattern Recognition}, pages 6299--6308, 2017.

\bibitem{rspnet}
Peihao Chen, Deng Huang, Dongliang He, Xiang Long, Runhao Zeng, Shilei Wen,
  Mingkui Tan, and Chuang Gan.
\newblock Rspnet: Relative speed perception for unsupervised video
  representation learning.
\newblock In {\em The AAAI Conference on Artificial Intelligence}, 2021.

\bibitem{simclr}
Ting Chen, Simon Kornblith, Mohammad Norouzi, and Geoffrey Hinton.
\newblock A simple framework for contrastive learning of visual
  representations.
\newblock In {\em ICML}, 2020.

\bibitem{cho2020self}
Hyeon Cho, Taehoon Kim, Hyung~Jin Chang, and Wonjun Hwang.
\newblock Self-supervised visual learning by variable playback speeds
  prediction of a video.
\newblock {\em IEEE Access}, 9:79562--79571, 2021.

\bibitem{choi}
Jinwoo Choi, Chen Gao, Joseph~C.E. Messou, and Jia-Bin Huang.
\newblock Why can't i dance in the mall? learning to mitigate scene bias in
  action recognition.
\newblock In {\em Advances in Neural Information Processing Systems}, pages
  853--865, 2019.

\bibitem{dave2021knights}
Ishan Dave, Naman Biyani, Brandon Clark, Rohit Gupta, Yogesh Rawat, and Mubarak
  Shah.
\newblock " knights": First place submission for vipriors21 action recognition
  challenge at iccv 2021.
\newblock {\em arXiv preprint arXiv:2110.07758}, 2021.

\bibitem{vdim}
R Devon~Hjelm and Philip Bachman.
\newblock Representation learning with video deep infomax.
\newblock {\em arXiv preprint arXiv:2007.13278}, 2020.

\bibitem{diba2020large}
Ali Diba, Mohsen Fayyaz, Vivek Sharma, Manohar Paluri, J\"urgen Gall, Rainer
  Stiefelhagen, and Luc Van~Gool.
\newblock {\em Large scale holistic video understanding}, pages 593--610.
\newblock 2020.

\bibitem{slowfast}
Christoph Feichtenhofer, Haoqi Fan, Jitendra Malik, and Kaiming He.
\newblock Slowfast networks for video recognition.
\newblock In {\em Proceedings of the IEEE International Conference on Computer
  Vision}, pages 6202--6211, 2019.

\bibitem{Feichtenhofer_2021_CVPR}
Christoph Feichtenhofer, Haoqi Fan, Bo Xiong, Ross Girshick, and Kaiming He.
\newblock A large-scale study on unsupervised spatiotemporal representation
  learning.
\newblock In {\em Proceedings of the IEEE/CVF Conference on Computer Vision and
  Pattern Recognition}, pages 3299--3309, 2021.

\bibitem{fernando2017self}
Basura Fernando, Hakan Bilen, Efstratios Gavves, and Stephen Gould.
\newblock Self-supervised video representation learning with odd-one-out
  networks.
\newblock In {\em Proceedings of the IEEE Conference on Computer Vision and
  Pattern Recognition}, pages 3636--3645, 2017.

\bibitem{motionfit}
Kirill Gavrilyuk, Mihir Jain, Ilia Karmanov, and Cees~GM Snoek.
\newblock Motion-augmented self-training for video recognition at smaller
  scale.
\newblock In {\em Proceedings of the IEEE/CVF International Conference on
  Computer Vision}, pages 10429--10438, 2021.

\bibitem{nce}
Michael Gutmann and Aapo Hyv\"arinen.
\newblock Noise-contrastive estimation: A new estimation principle for
  unnormalized statistical models.
\newblock In {\em Proceedings of the Thirteenth International Conference on
  Artificial Intelligence and Statistics}, pages 297--304, 2010.

\bibitem{dpc}
Tengda Han, Weidi Xie, and Andrew Zisserman.
\newblock Video representation learning by dense predictive coding.
\newblock In {\em Proceedings of the IEEE/CVF International Conference on
  Computer Vision (ICCV) Workshops}, 2019.

\bibitem{memdpc}
Tengda Han, Weidi Xie, and Andrew Zisserman.
\newblock {\em Memory-augmented dense predictive coding for video
  representation learning}, pages 312--329.
\newblock 2020a.

\bibitem{cotraining}
Tengda Han, Weidi Xie, and Andrew Zisserman.
\newblock {\em Self-supervised Co-Training for Video Representation Learning},
  pages 5679--5690.
\newblock 2020b.

\bibitem{kenshohara}
K. {Hara}, H. {Kataoka}, and Y. {Satoh}.
\newblock Towards good practice for action recognition with spatiotemporal 3d
  convolutions.
\newblock In {\em 2018 24th International Conference on Pattern Recognition},
  pages 2516--2521, 2018.

\bibitem{moco}
Kaiming He, Haoqi Fan, Yuxin Wu, Saining Xie, and Ross Girshick.
\newblock Momentum contrast for unsupervised visual representation learning.
\newblock In {\em Proceedings of the IEEE/CVF Conference on Computer Vision and
  Pattern Recognition}, pages 9729--9738, 2020.

\bibitem{csj}
Yuqi Huo, Mingyu Ding, Haoyu Lu, Zhiwu Lu, Tao Xiang, Ji-Rong Wen, Ziyuan
  Huang, Jianwen Jiang, Shiwei Zhang, Mingqian Tang, Songfang Huang, and Ping
  Luo.
\newblock Self-supervised video representation learning with constrained
  spatiotemporal jigsaw, 2021.

\bibitem{jenni2021time}
Simon Jenni and Hailin Jin.
\newblock Time-equivariant contrastive video representation learning.
\newblock In {\em Proceedings of the IEEE/CVF International Conference on
  Computer Vision}, pages 9970--9980, 2021.

\bibitem{simon}
Simon Jenni, Givi Meishvili, and Paolo Favaro.
\newblock Video representation learning by recognizing temporal
  transformations.
\newblock In {\em The European Conference on Computer Vision}, 2020.

\bibitem{3drotnet}
Longlong Jing, Xiaodong Yang, Jingen Liu, and Yingli Tian.
\newblock Self-supervised spatiotemporal feature learning via video rotation
  prediction.
\newblock {\em arXiv preprint arXiv:1811.11387}, 2018.

\bibitem{kenshohara_mega_scale}
Hirokatsu Kataoka, Tenga Wakamiya, Kensho Hara, and Yutaka Satoh.
\newblock Would mega-scale datasets further enhance spatiotemporal 3d cnns?
\newblock {\em arXiv preprint arXiv:2004.04968}, 2020.

\bibitem{aaai19}
Dahun Kim, Donghyeon Cho, and In~So Kweon.
\newblock Self-supervised video representation learning with space-time cubic
  puzzles.
\newblock In {\em Proceedings of the AAAI Conference on Artificial
  Intelligence, vol. 33}, pages 8545--8552, 2019.

\bibitem{adam}
Diederik~P. Kingma and Jimmy Ba.
\newblock Adam: {A} method for stochastic optimization.
\newblock In Yoshua Bengio and Yann LeCun, editors, {\em 3rd International
  Conference on Learning Representations, {ICLR} 2015, San Diego, CA, USA, May
  7-9, 2015, Conference Track Proceedings}, 2015.

\bibitem{tce}
Joshua Knights, Ben Harwood, Daniel Ward, Anthony Vanderkop, Olivia
  Mackenzie-Ross, and Peyman Moghadam.
\newblock Temporally coherent embeddings for self-supervised video
  representation learning.
\newblock In {\em 2020 25th International Conference on Pattern Recognition
  (ICPR)}, pages 8914--8921. IEEE, 2021.

\bibitem{hmdb}
H. Kuehne, H. Jhuang, E. Garrote, T. Poggio, and T. Serre.
\newblock Hmdb: a large video database for human motion recognition.
\newblock In {\em Proceedings of the International Conference on Computer
  Vision}, 2011.

\bibitem{lee2017unsupervised}
Hsin-Ying Lee, Jia-Bin Huang, Maneesh Singh, and Ming-Hsuan Yang.
\newblock Unsupervised representation learning by sorting sequences.
\newblock In {\em Proceedings of the IEEE International Conference on Computer
  Vision}, pages 667--676, 2017.

\bibitem{lengyel2022vipriors}
Attila Lengyel, Robert-Jan Bruintjes, Marcos~Baptista Rios, Osman~Semih Kayhan,
  Davide Zambrano, Nergis Tomen, and Jan van Gemert.
\newblock Vipriors 2: Visual inductive priors for data-efficient deep learning
  challenges.
\newblock {\em arXiv preprint arXiv:2201.08625}, 2022.

\bibitem{diving}
Yingwei Li, Yi Li, and Nuno Vasconcelos.
\newblock Resound: Towards action recognition without representation bias.
\newblock In {\em Proceedings of the European Conference on Computer Vision},
  pages 513--528, 2018.

\bibitem{wacv20}
Guillaume Lorre, Jaonary Rabarisoa, Astrid Orcesi, Samia Ainouz, and Stephane
  Canu.
\newblock Temporal contrastive pretraining for video action recognition.
\newblock In {\em The IEEE Winter Conference on Applications of Computer
  Vision}, pages 662--670, 2020.

\bibitem{vcp}
Dezhao Luo, Chang Liu, Yu Zhou, Dongbao Yang, Can Ma, Qixiang Ye, and Weiping
  Wang.
\newblock Video cloze procedure for self-supervised spatio-temporal learning.
\newblock In {\em Proceedings of the AAAI Conference on Artificial
  Intelligence, vol. 33}, pages 11701--11708, 2020.

\bibitem{miech2020end}
Antoine Miech, Jean-Baptiste Alayrac, Lucas Smaira, Ivan Laptev, Josef Sivic,
  and Andrew Zisserman.
\newblock End-to-end learning of visual representations from uncurated
  instructional videos.
\newblock In {\em Proceedings of the IEEE/CVF Conference on Computer Vision and
  Pattern Recognition}, pages 9879--9889, 2020a.

\bibitem{milnce}
Antoine Miech, Jean-Baptiste Alayrac, Lucas Smaira, Ivan Laptev, Josef Sivic,
  and Andrew Zisserman.
\newblock End-to-end learning of visual representations from uncurated
  instructional videos.
\newblock In {\em Proceedings of the IEEE/CVF Conference on Computer Vision and
  Pattern Recognition}, pages 9879--9889, 2020b.

\bibitem{misra2016shuffle}
Ishan Misra, C.~Lawrence Zitnick, and Martial Hebert.
\newblock {\em Shuffle and learn: unsupervised learning using temporal order
  verification}, pages 527--544.
\newblock 2016.

\bibitem{cpc}
Aaron van~den Oord, Yazhe Li, and Oriol Vinyals.
\newblock Representation learning with contrastive predictive coding.
\newblock {\em arXiv preprint arXiv:1807.03748}, 2018.

\bibitem{videomoco}
Tian Pan, Yibing Song, Tianyu Yang, Wenhao Jiang, and Wei Liu.
\newblock Videomoco: Contrastive video representation learning with temporally
  adversarial examples.
\newblock In {\em Proceedings of the IEEE/CVF Conference on Computer Vision and
  Pattern Recognition}, pages 11205--11214, 2021.

\bibitem{gdt}
Mandela Patrick, Yuki Asano, Polina Kuznetsova, Ruth Fong, Joao~F. Henriques,
  Geoffrey Zweig, and Andrea Vedaldi.
\newblock Multi-modal self-supervision from generalized data transformations,
  2021.

\bibitem{iccv21qian}
Rui Qian, Yuxi Li, Huabin Liu, John See, Shuangrui Ding, Xian Liu, Dian Li, and
  Weiyao Lin.
\newblock Enhancing self-supervised video representation learning via
  multi-level feature optimization.
\newblock In {\em Proceedings of the International Conference on Computer
  Vision}, 2021.

\bibitem{cvrl}
Rui Qian, Tianjian Meng, Boqing Gong, Ming-Hsuan Yang, Huisheng Wang, Serge
  Belongie, and Yin Cui.
\newblock Spatiotemporal contrastive video representation learning.
\newblock In {\em Proceedings of the IEEE/CVF Conference on Computer Vision and
  Pattern Recognition}, pages 6964--6974, 2021.

\bibitem{iclr21submitted_st_invariant}
Hao Shao, Yu Liu, and Hongsheng Li.
\newblock Self-supervised temporal learning, 2021.

\bibitem{ucf101}
Khurram Soomro, Amir~Roshan Zamir, and Mubarak Shah.
\newblock Ucf101: A dataset of 101 human actions classes from videos in the
  wild.
\newblock {\em arXiv preprint arXiv:1212.0402}, 2012.

\bibitem{cbt}
Chen Sun, Fabien Baradel, Kevin Murphy, and Cordelia Schmid.
\newblock Learning video representations using contrastive bidirectional
  transformer.
\newblock {\em arXiv preprint arXiv:1906.05743}, 2019.

\bibitem{suzuki2018learning}
Tomoyuki Suzuki, Takahiro Itazuri, Kensho Hara, and Hirokatsu Kataoka.
\newblock Learning spatiotemporal 3d convolution with video order
  self-supervision.
\newblock In {\em Proceedings of the European Conference on Computer Vision},
  2018.

\bibitem{iic}
Li Tao, Xueting Wang, and Toshihiko Yamasaki.
\newblock Self-supervised video representation learning using inter-intra
  contrastive framework.
\newblock In {\em Proceedings of the 28th ACM International Conference on
  Multimedia}, pages 2193--2201, 2020.

\bibitem{local_motion_cues}
Yuan Tian, Zhaohui Che, Wenbo Bao, Guangtao Zhai, and Zhiyong Gao.
\newblock {\em Self-supervised Motion Representation via Scattering Local
  Motion Cues}, pages 71--89.
\newblock 2020.

\bibitem{idt}
Pavel Tokmakov, Martial Hebert, and Cordelia Schmid.
\newblock Unsupervised learning of video representations via dense trajectory
  clustering.
\newblock In {\em European Conference on Computer Vision}, pages 404--421.
  Springer, 2020.

\bibitem{c3d}
Du Tran, Lubomir Bourdev, Rob Fergus, Lorenzo Torresani, and Manohar Paluri.
\newblock Learning spatiotemporal features with 3d convolutional networks.
\newblock In {\em Proceedings of the IEEE International Conference on Computer
  Vision}, pages 4489--4497, 2015.

\bibitem{r2plus1d}
Du Tran, Heng Wang, Lorenzo Torresani, Jamie Ray, Yann LeCun, and Manohar
  Paluri.
\newblock A closer look at spatiotemporal convolutions for action recognition.
\newblock In {\em Proceedings of the IEEE Conference on Computer Vision and
  Pattern Recognition}, pages 6450--6459, 2018.

\bibitem{tsne}
Laurens Van Der~Maaten.
\newblock Accelerating t-sne using tree-based algorithms.
\newblock {\em The Journal of Machine Learning Research}, 15(1):3221--3245,
  2014.

\bibitem{scene-motion}
Jinpeng Wang, Yuting Gao, Ke Li, Xinyang Jiang, Xiaowei Guo, Rongrong Ji, and
  Xing Sun.
\newblock Enhancing unsupervised video representation learning by decoupling
  the scene and the motion.
\newblock In {\em The AAAI Conference on Artificial Intelligence}, 2021.

\bibitem{statistics2}
Jiangliu Wang, Jianbo Jiao, Linchao Bao, Shengfeng He, Wei Liu, and Yun-Hui
  Liu.
\newblock Self-supervised video representation learning by uncovering
  spatio-temporal statistics.
\newblock {\em IEEE Transactions on Pattern Analysis and Machine Intelligence},
  2021.

\bibitem{statistics_cvpr19}
Jiangliu Wang, Jianbo Jiao, Linchao Bao, Shengfeng He, Yunhui Liu, and Wei Liu.
\newblock Self-supervised spatio-temporal representation learning for videos by
  predicting motion and appearance statistics.
\newblock In {\em Proceedings of the IEEE Conference on Computer Vision and
  Pattern Recognition}, pages 4006--4015, 2019.

\bibitem{pace_pred}
Jiangliu Wang, Jianbo Jiao, and Yun-Hui Liu.
\newblock Self-supervised video representation learning by pace prediction.
\newblock In {\em The European Conference on Computer Vision}, 2020.

\bibitem{aot}
Donglai Wei, Joseph~J. Lim, Andrew Zisserman, and William~T. Freeman.
\newblock Learning and using the arrow of time.
\newblock In {\em Proceedings of the IEEE Conference on Computer Vision and
  Pattern Recognition}, pages 8052--8060, 2018.

\bibitem{minikinetics}
Saining Xie, Chen Sun, Jonathan Huang, Zhuowen Tu, and Kevin Murphy.
\newblock Rethinking spatiotemporal feature learning: Speed-accuracy trade-offs
  in video classification.
\newblock In {\em Proceedings of the European Conference on Computer Vision},
  pages 305--321, 2018.

\bibitem{vcop}
Dejing Xu, Jun Xiao, Zhou Zhao, Jian Shao, Di Xie, and Yueting Zhuang.
\newblock Self-supervised spatiotemporal learning via video clip order
  prediction.
\newblock In {\em Proceedings of the IEEE Conference on Computer Vision and
  Pattern Recognition}, pages 10334--10343, 2019.

\bibitem{dvim}
Fei Xue, Hongbing Ji, Wenbo Zhang, and Yi Cao.
\newblock Self-supervised video representation learning by maximizing mutual
  information.
\newblock {\em Signal Processing: Image Communication}, 88:115967, 2020.

\bibitem{vtc}
Ceyuan Yang, Yinghao Xu, Bo Dai, and Bolei Zhou.
\newblock Video representation learning with visual tempo consistency.
\newblock {\em arXiv preprint arXiv:2006.15489}, 2020.

\bibitem{seco}
Ting Yao, Yiheng Zhang, Zhaofan Qiu, Yingwei Pan, and Tao Mei.
\newblock Seco: Exploring sequence supervision for unsupervised representation
  learning.
\newblock In {\em AAAI}, volume~2, page~7, 2021.

\bibitem{prp}
Yuan Yao, Chang Liu, Dezhao Luo, Yu Zhou, and Qixiang Ye.
\newblock Video playback rate perception for self-supervised spatio-temporal
  representation learning.
\newblock In {\em Proceedings of the IEEE/CVF Conference on Computer Vision and
  Pattern Recognition}, pages 6548--6557, 2020a.

\bibitem{attentiontransfer}
Sergey Zagoruyko and Nikos Komodakis.
\newblock Paying more attention to attention: Improving the performance of
  convolutional neural networks via attention transfer.
\newblock In {\em 5th International Conference on Learning Representations,
  {ICLR} 2017, Toulon, France, April 24-26, 2017, Conference Track
  Proceedings}. OpenReview.net, 2017.

\bibitem{vie}
Chengxu Zhuang, Tianwei She, Alex Andonian, Max~Sobol Mark, and Daniel Yamins.
\newblock Unsupervised learning from video with deep neural embeddings.
\newblock In {\em Proceedings of the IEEE/CVF Conference on Computer Vision and
  Pattern Recognition}, pages 9563--9572, 2020.

\end{thebibliography}

\clearpage
\appendix
\section{Supplementary Overview}

The supplementary material is organized into the following sections:

\begin{itemize}
    \item Section~\ref{sec:dataset}: Dataset details
    \item Section~\ref{sec:impl_details}: Implementation details such as network architectures, data augmentations, self supervised training details (losses, optimizers, etc) and and downstream task evaluation protocols. 
    \item Section~\ref{sec:ablation}: Results for additional ablation experiments, to assess the impact of augmentation types, projection head design and loss hyperparameters on performance.
    \item Section~\ref{sec:bl_verification}: Comparison of our instance contrastive and random initialization baseline with results reported in prior work
    \item Section~\ref{sec:temp_diversity}: Effect of temporal diversity on downstream tasks
    \item Section~\ref{sec:additional_comparison}: Additional results from prior work which were excluded from the main paper
    \item Section~\ref{sec:feature_slice_similarity}: Feature slice similarity matrix of the learned representation 
    \item Section~\ref{sec:qualitative}: Qualitative results of NN-Retrieval
    \item Section~\ref{sec:tsne}: tSNE visualization of learned features and comparison with baselines
    \item Section~\ref{sec:attention}: Visualization of model attention
    \item Section~\ref{sec:detailed_comparison}: Detailed comparison with the most relevant recent prior works
\end{itemize}

\section{Datasets}
\label{sec:dataset}
\noindent\textbf{UCF101}~\cite{ucf101} is the most commonly used dataset for the action recognition task. UCF101 has around 13,320 videos representing 101 different human activities.

\vspace{1mm}

\noindent\textbf{Kinetics400}~\cite{kinetics} is a large-scale human action recognition dataset having having 400 action classes summing up to 222,454 training videos at the time of our download. 

\vspace{1mm}

\noindent\textbf{HMDB51}~\cite{hmdb} is a relatively smaller action recognition having around 7,000 total videos collected from 51 different human actions.

\section{Implementation Details}
\label{sec:impl_details}
\subsection{Network Architecture}
We use the implementation of 3D-ResNet-18, R(2+1)D-18 and C3D provided by authors of prior works~\cite{pace_pred, prp, vcop}. For 3D-ResNet and R(2+1)D, to get a feature map with temporal dimension 4 at the end of the penultimate layer, we change the temporal stride of the conv5 layer to 1, and the dilation factor to 2 in order to keep the temporal receptive field the same, while increasing the temporal resolution of the features. We spatially pool the feature map only after the penultimate layer which results in an encoded feature of dimension $4 \times 512$ where 4 is temporal dimension and 512 is number of channels.

The MLP projection head consists of 2 layers in the default setting: Linear($512$, $512$)  with ReLU activation and Linear($512$, $128$) followed by L2-Normalization. We use the 128-D output as representation vector to compute losses. We use shared weights for the 3D-CNN backbone and MLP across computation of different losses. 
\subsection{Augmentation}

The augmentations we use for reporting benchmark scores are standard and prior methods also utilize similar augmentation strategies. We apply augmentation transformations from  two broad categories: (1) Appearance based transforms and (2) Geometry based transforms. 

\vspace{1mm}

\noindent\textbf{Geometry based transforms} Basic geometry-based augmentations include random scaling, random cropping, random cut-out and random horizontal-flip (details in Algorithm~\ref{algo:augmentation}). Apart from basic augmentations, we also try the use of shear and rotation transformation with a small range (details in Section~\ref{sec:ablation}), however for our benchmark results we only use basic augmentations.

\vspace{1mm}

\noindent\textbf{Appearance based transforms} Basic appearance-based transforms include random grayscale, random color drop, and color jittering with HSV color space (details in Algorithm~\ref{algo:augmentation}). We also experiment with using Gaussian blurring filter (details in Section~\ref{sec:ablation}), but we don't use it for our reported results. 

\noindent Each augmentation is applied with a certain probability and random strength selected from a specified interval. The exact same augmentation is applied on each frames of a video clip to ensure temporal consistency. Our augmentation pipeline is explained in Algorithm~\ref{algo:augmentation}.

\begin{algorithm}
\DontPrintSemicolon
    \small
  \textbf{Input:} Frames $F$, \textbf{Output:} Transformed Frames $F'$
   \tcc{Random Resized Crop}
   $w$, $h$ = dimensions($F$)\;
   Generate $scale\sim \mathcal{U}(0.6,\,1.0)$\;
   Generate $s_x\sim \mathcal{U}(0,\,1),\,s_y\sim \mathcal{U}(0,\,1)$\;
   $x_0,\, y_0 = s_x\times(w - w \times scale),\, s_y\times(h - h \times scale)$\;
   $x_1,\,y_1 = x_0 + scale\times w,\,y_0 + scale\times h$\;
   $F'=F$.spatial\_crop$(x_0, y_0, x_1, y_1)$.resize$(112, 112)$\;
   \tcc{Random contrast}
   Generate $p_0\sim \mathcal{U}(0,\,1)$\;
  \If{$p_0 < 0.25$}
    {
        Generate $m_0\sim \mathcal{U}(0.75,\,1.25)$\;
        $F'$ = $F'$.adjust\_contrast($m_0$)\;
    }
    \tcc{Random Hue}
   Generate $p_1\sim \mathcal{U}(0,\,1)$\;
  \If{$p_1 < 0.3$}
    {
        Generate $m_1\sim \mathcal{U}(-0.1,\,0.1)$\;
        $F'$ = $F'$.adjust\_hue($m_1$)\;
    }
    \tcc{Random Saturation}
   Generate $p_2\sim \mathcal{U}(0,\,1)$\;
  \If{$p_2 < 0.3$}
    {
        Generate $m_2\sim \mathcal{U}(0.75,\,1.25)$\;
        $F'$ = $F'$.adjust\_saturation($m_2$)\;
    }
    \tcc{Random Brightness}
   Generate $p_3\sim \mathcal{U}(0,\,1)$\;
  \If{$p_3 < 0.3$}
    {
        Generate $m_3\sim \mathcal{U}(0.75,\,1.25)$\;
        $F'$ = $F'$.adjust\_brightness($m_3$)\;
    }
    \tcc{Color Dropping}
    Generate $p_4\sim \mathcal{U}(0,\,1)$\;
    \If{$p_4 > 0.7$}
    {
        \If{$p_4 < 0.875$}
        {
        	$F'$ = $F'$.to\_grayscale()\;
        	Generate $p_5\sim \mathcal{U}(0,\,1)$\;
        	\If{$p_5 < 0.75$}
        	{
        	    Generate $m_4\sim \mathcal{U}(0.75,\,1.25)$\;
        	    $F'$ = $F'$.adjust\_gamma($m_4$)\;
        	}
        }
        \Else
        {
        	$F'$ = $F'$.drop\_random\_channel()\;
        }
    }
    \tcc{Random Horizontal Flip}
    Generate $p_6\sim \mathcal{U}(0,\,1)$\;
  \If{$p_6 < 0.5$}
    {
        $F'$ = $F'$.horizontal\_flip()\;
    }
    \tcc{Random Erase}
    Generate $p_7\sim \mathcal{U}(0,\,1)$\;
    \If{$p_7 < 0.5$}
    {
        Generate $e_x\sim 112\times\mathcal{U}(0,\,1),\,e_y\sim 112\times\mathcal{U}(0,\,1)$\;
        Generate $escale_x,\,escale_y\sim 20\times \mathcal{U}(0.5,\,1)$\;
        $F'$ = $F'$.erase($e_x,\,e_y,\,escale_x,\,escale_y$)\;
    }
 \caption{TCLR self-supervised training data augmentation.}
 \label{algo:augmentation}
\end{algorithm}

\subsection{Self-supervised training}
The model for self-supervised pretraining consists of a  backbone architecture combined with an MLP projection head. The loss is computed using the embedding from the final layer of the MLP. We use an input clip resolution of $112\times112$ and 16 frames. From a set of 64 consecutive frames, the global clip is composed of 16 frames sampled with a skip rate of 4, whereas the 4 local clips also have 16 frames but are sampled with a skip rate of 1. We train our model by combining all loss terms with batch size of $40$ as shown in Equation \ref{eq:all_losses}, where, $N_B$ is the size of mini-batch, $\mathcal{L}_{IC}$ is the instance contrastive loss, $\mathcal{L}_{LL}$ is the local-local temporal contrastive loss and $\mathcal{L}_{GL}$ is the global-local temporal contrastive loss.

\begin{equation}\label{eq:all_losses}
  \mathcal{L} = \sum_{i=1}^{N_B} \mathcal{L}_{IC}^{i} + \mathcal{L}_{LL}^{i} + \mathcal{L}_{GL}^{i},
\end{equation}

\begin{figure}[h]
\vspace{-2mm}
\centering
\begin{subfigure}[t]{0.49\columnwidth}
    \includegraphics[width=\textwidth]{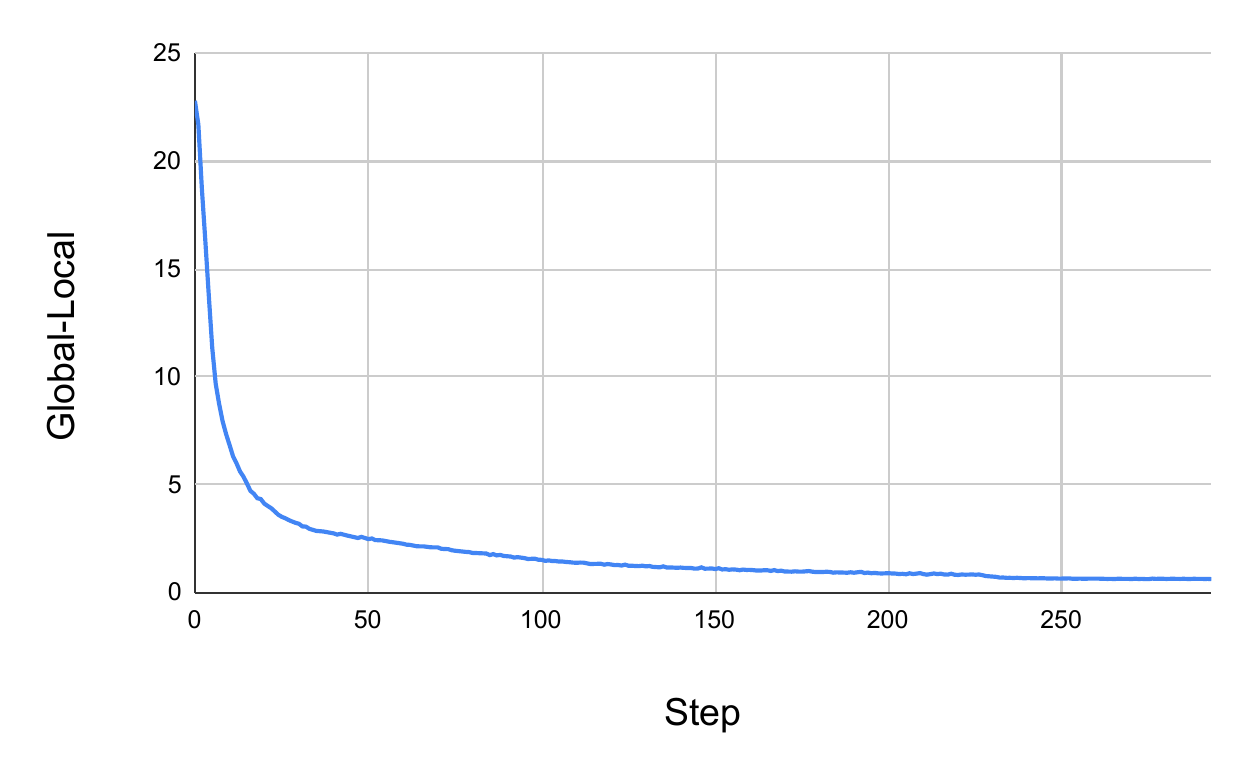}
    \caption{Global-Local}
\end{subfigure}
\hfill%
\begin{subfigure}[t]{0.49\columnwidth}
    \includegraphics[width=\textwidth]{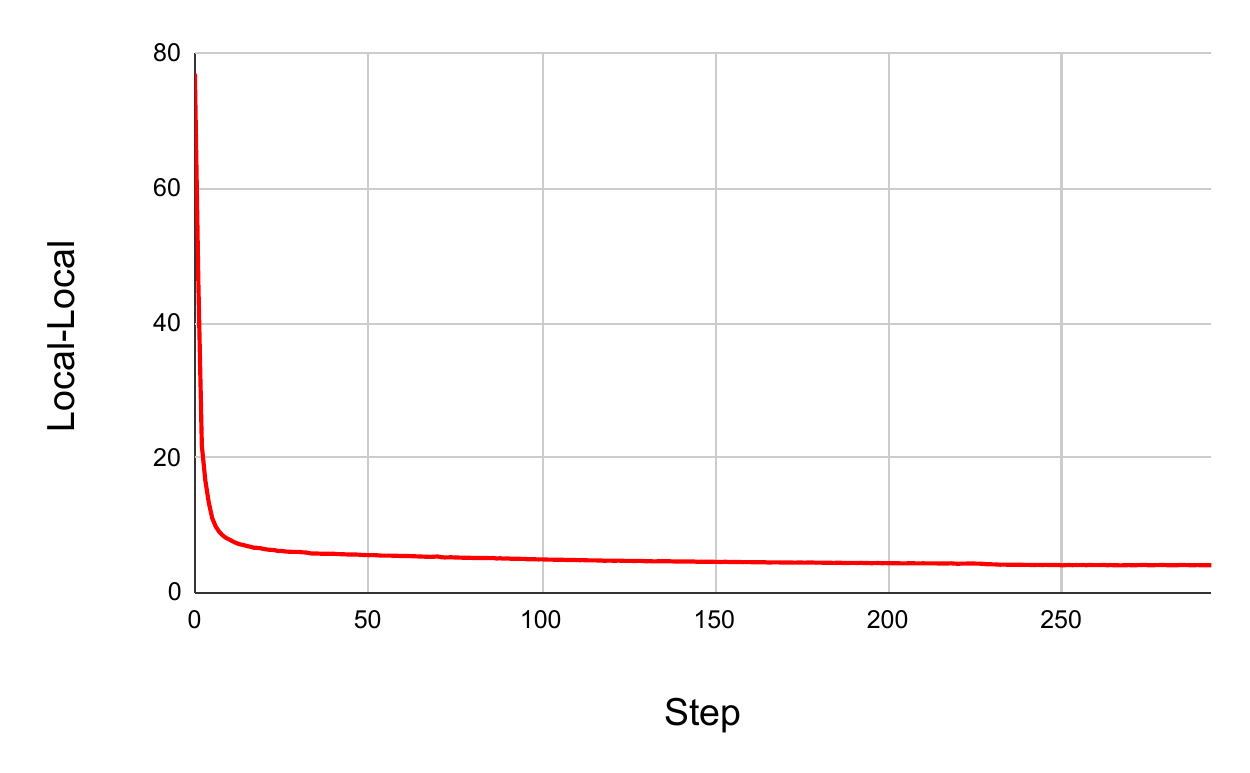}
    \caption{Local-Local}
\end{subfigure}
\begin{subfigure}[t]{\columnwidth}
\centering
    \includegraphics[width=0.5\columnwidth]{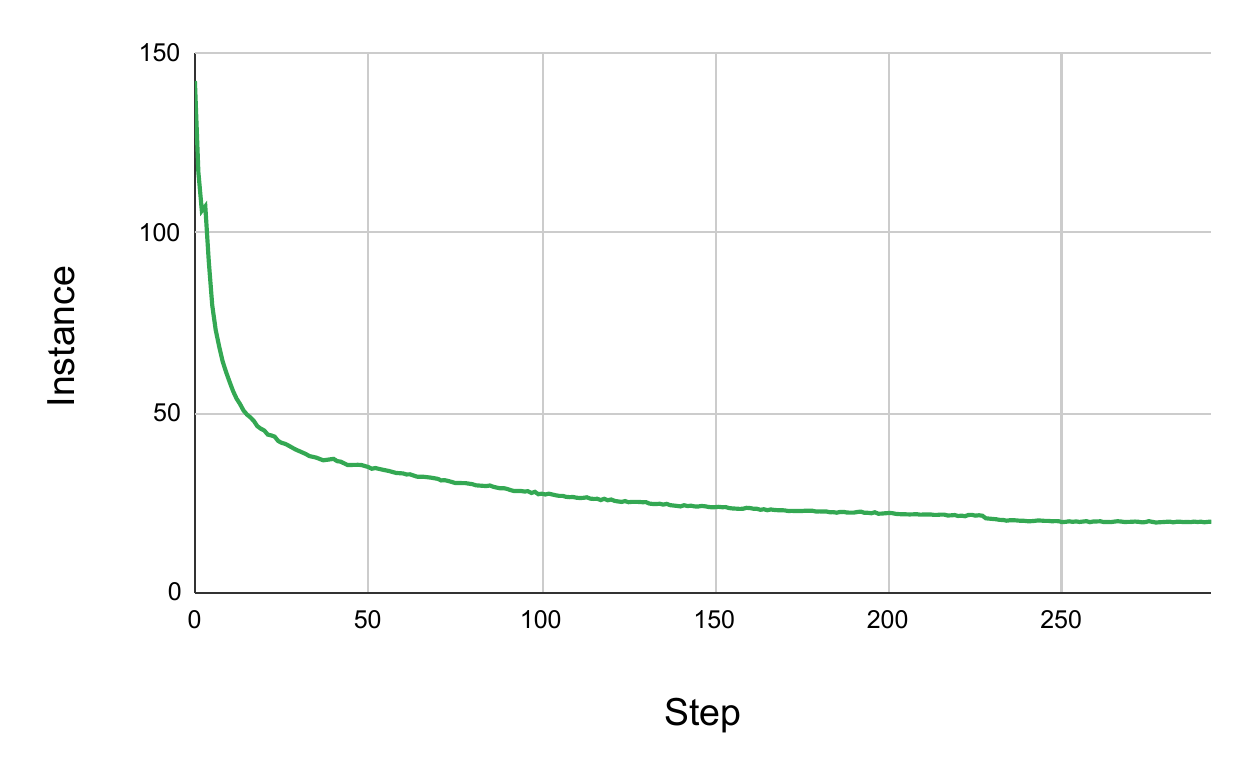}
    \caption{Instance}
\end{subfigure}
\vspace{-3mm}
\caption{Values of contrastive losses during progression of self-supervised pre-training.}
\label{fig:log_losses}
\vspace{-3mm}
\end{figure}

Figure~\ref{fig:log_losses} shows the values of different losses as the training progresses. We run the self-supervised pretraining for 400 epochs for UCF101 and 100 epochs for pretraining of Kinetics400. We use Adam~\cite{adam} optimizer with $\beta_1=0.9, \beta_2=0.999, \epsilon=10^{-8}$, the default parameters for PyTorch v1.6.0. We use an initial learning of $0.001$ and decay the learning rate by a factor of 10 when the loss plateaus. We also use linear warm-up for the first 10 epochs.

As mentioned in Table-1 of the main paper, we reproduce results of CVRL for the commonly used configuration (model depth = 18, input frames = 16, resolution = $112\times112$). In our replication, we use the exact same temporally consistent augmentations and temporal interval sampler $P(t) \propto - t + c$. Pretraining is done on UCF101 split-1 training set with batch size of 512 for 800 epochs. 

\subsection{Downstream Task Protocols}
We evaluate self-supervised pretrained model for three downstream tasks:   Action Recognition, Nearest Neighbor Video Retrieval, and Label Efficiency.

\vspace{1mm}

\noindent\textbf{Action Recognition:} For the action recognition task we  follow the protocol used in prior works~\cite{simon,memdpc,pace_pred, dpc,idt}. We attach a fully-connected layer (randomly initialized) to the pretrained video encoder and train all layers of the model using the training data with a cross-entropy loss. Following the prior works, we also utilize basic augmentations such as random crop, random scale and horizontal flip while training. These augmentations are standard for action recognition training~\cite{kenshohara, r2plus1d}. For the testing, we sample 10 uniformly spaced clips from the video average their predictions to get a video-level prediction. We utilize input clips of $112\times112$ resolution and 16 frames with a skip rate of 2. A base learning rate of $0.001$ is used, which decreases by a factor of 10 on loss plateau. Linear warm-up starting from minimum learning rate of $0.00001$ and rising to base Learning Rate of $0.001$ is used for the first 10 epochs of the training. 

\vspace{1mm}

\noindent\textbf{Label Efficiency/ Finetuning with limited data:} We use the same experimental setting as used for action recognition task and train with limited training data. For each limited fraction of training data, we repeat the experiment 3 times with different randomly sampled training-subsets and report average result. 

\vspace{1mm}

\noindent\textbf{Nearest Neighbor Video Retrieval:} 
In this task, we directly use the self-supervised pretrained video encoder without further supervised finetuning. In the paper we report results using features from the final layer of the encoder with spatial pooling and averaged over 10 uniformly spaced clips from each video as followed by prior work~\cite{memdpc}.

\section{Additional Ablations}
\label{sec:ablation}
We perform additional ablations using 3D-ResNet-18 with self-supervised pretraining on UCF101 training set (split-1) and linear evaluation on UCF101 test set (split-1).

\subsection{Impact of losses}

In the main paper, we carry out ablation studies to measure the impact of each of our novel loss functions on the downstream tasks. A more complete set of results for these experiments are presented in Tables~\ref{table:nnr_ucf_abl}, \ref{table:nnr_hmdb_abl} and \ref{table:limited_label_abl}.

\begin{table}[h]
\centering
\begingroup
\setlength{\tabcolsep}{3pt} %
\renewcommand{\arraystretch}{1.0} %
\begin{tabular}{lcccc} 
\hline

\hline

\hline\\[-3mm]
\textbf{Method}         & \textbf{Recall@1}   & \textbf{Recall@5}   & \textbf{Recall@10}  & \textbf{Recall@20}   \\ 
\hline

\hline

\hline\\[-3mm]

\textbf{IC}          & 40.76 & 56.41 & 65.33 & 74.44  \\
\textbf{IC+LL}           & 51.1 & 67.83 & 74.57 & 80.89  \\
\textbf{IC+GL}           & 47.32 & 63.10 & 71.42 & 78.72  \\
\textbf{TCLR }            & 56.17 & 72.16 & 79.01 & 85.30  \\ 
\hline

\hline

\hline\\[-3mm]
\end{tabular}
\endgroup
\caption{Nearest Neighbor Video Retrieval on UCF101 using self-supervised R3D-18 models pretrained on UCF101 videos.}

\label{table:nnr_ucf_abl}
\end{table}

\begin{table}[h]
\centering
\begingroup
\setlength{\tabcolsep}{3pt} %
\renewcommand{\arraystretch}{1.0} %
\begin{tabular}{lcccc} 
\hline

\hline

\hline\\[-3mm]
\textbf{Method}         & \textbf{Recall@1}   & \textbf{Recall@5}   & \textbf{Recall@10}  & \textbf{Recall@20}   \\ 
\hline

\hline

\hline\\[-3mm]

\textbf{IC}         &14.38& 35.62& 48.37& 61.57  \\
\textbf{IC+LL}      & 19.07& 42.42& 54.97& 69.35 \\
\textbf{IC+GL}      & 18.43& 41.70& 53.59& 67.19  \\
\textbf{TCLR }      & 22.75& 45.36& 57.84& 73.07  \\ 
\hline

\hline

\hline\\[-3mm]
\end{tabular}
\endgroup
\caption{Nearest Neighbor Video Retrieval on HMDB51 using self-supervised R3D-18 models pretrained on UCF101 videos.}
\label{table:nnr_hmdb_abl}
\end{table}

\begin{table}[h]
\centering
\begin{tabular}{lcccc} 
\hline

\hline

\hline\\[-3mm]
       \textbf{Method} & \textbf{1\%}  & \textbf{10\%}   & \textbf{20\%}  & \textbf{50\%}   \\ 
\hline

\hline

\hline\\[-3mm]
\textbf{Scratch} & 6.13   & 32.10   & 44.39  & 57.35     \\ 
\textbf{IC}      & 22.86 & 53.91 & 60.26 & 65.74   \\ 
\textbf{LL + GL} & 9.78  & 48.66  & 59.45 & 68.64   \\ 
\textbf{IC + LL} & 26.43 & 63.09  & 69.68 & 73.53   \\ 
\textbf{IC + GL} & 23.93 & 60.16  & 67.40  & 72.34   \\ 
\textbf{TCLR}    & 26.90 & 66.08 & 73.41  & 76.68     \\
\hline

\hline

\hline\\[-3mm]
\end{tabular}
\caption{Top-1 Accuracy after limited label finetuning of self-supervised R3D-18 models pretrained on UCF101 (split-1) videos.}
\label{table:limited_label_abl}
\end{table}

\subsection{Augmentation Types}

    We perform an ablation study to measure the impact of computationally expensive transformations (specifically random Gaussian Blurring, Shearing and Rotation). We observe (Results in Table~\ref{table:augablation}) that adding these transforms only has a small effect on the performance of the model on downstream tasks, and removing them can reduce training time by as much as 30\%. In these augmentation ablations, we use Gaussian blur with 30\% probability, kernel size $(15,15)$ and standard deviation selected randomly from $(0,0.5)$ interval. We apply shear and rotation with 30\% probability, with rotation angles selected randomly from $(-20^{\circ}, 20^{\circ})$. Even though it's possible to obtain slightly better results by using blur and other complex augmentations we do not use them for our main reported results.
    
\begin{table}[h]
\centering
\begin{tabular}{llc} 
\hline

\hline

\hline\\[-3mm]
\multicolumn{2}{c}{\textbf{Transforms}} &   \\ 
\textbf{Appearance}   & \textbf{Geometric}         & \begin{tabular}[c]{@{}l@{}}\textbf{Linear Evaluation}\\~(\textbf{Top-1 Accuracy})\end{tabular}                  \\ 
\hline

\hline

\hline\\[-3mm]
\textbf{Basic}  & \textbf{Basic} & \textbf{69.91}                                        \\
+ Blur & Basic & 70.7 \textcolor{c2}{(+0.8)}                                 \\ 
Basic  & + Shear \& Rotate & 70.2 \textcolor{c2}{(+0.3)}  \\
\hline

\hline

\hline\\[-3mm]
\end{tabular}

\caption{Effect of using additional transformations.}
\label{table:augablation}
\end{table}

\subsection{Embedding Size}

It has previously been reported that the size of the embedding doesn't have a significant effect~\cite{simclr} on image self supervised learning. In our experiments we observe that using a large embedding size typically results in slightly better performance, as reported in Table~\ref{table:embedabl}. The main paper utilizes embedding of 128.

\begin{table}[h]
\centering
\begin{tabular}{cc} 
\hline

\hline

\hline\\[-3mm]
\textbf{Embedding size} & \begin{tabular}[c]{@{}l@{}}\textbf{Linear Evaluation}\\~(\textbf{Top-1 Accuracy})\end{tabular}  \\ 
\hline

\hline

\hline\\[-3mm]
\textbf{128} & \textbf{69.91} \\

256 & 70.12 \textcolor{c2}{(+0.21)} \\ 

512 & 70.63 \textcolor{c2}{(+0.72)}  \\ 
\hline

\hline

\hline\\[-3mm]
\end{tabular}
\caption{Effect of embedding size.}
\label{table:embedabl}
\end{table}

\subsection{Number of Timesteps}
A key hyperparameter for the Temporal Constrastive losses is the number of Timesteps $(N_T)$ that a given video instance is sliced into. Apart from the default setting of $N_T = 4$ we also study $N_T = 2$ and find that it degrades performance. We do not use higher values of $N_T$ since that causes a significant increase in amount of GPU memory and computation required. Results are available in Table~\ref{table:timestepabl}.

\begin{table}[h]
\centering
\begin{tabular}{cc} 
\hline

\hline

\hline\\[-3mm]
\textbf{Timesteps $(N_T)$} & \begin{tabular}[c]{@{}l@{}}\textbf{Linear Evaluation}\\~(\textbf{Top-1 Accuracy})\end{tabular}  \\ 
\hline

\hline

\hline\\[-3mm]
\textbf{4} & \textbf{69.91} \\
2 & 66.60 \textcolor{c3}{(-3.31)} \\ 
\hline

\hline

\hline\\[-3mm]
\end{tabular}
\caption{Effect of timesteps used in temporal contrastive losses.}
\label{table:timestepabl}
\end{table}

\subsection{Skip Rate}

For the instance contrastive loss we generally use the same skip rate for both the anchor and pairing clip. We try using different skip rates (selected from $\{4,8,12,16\}$), but this results in a performance degradation of about 2.3\% over the baseline. 

For the Global-Local Temporal contrastive loss, we use a fixed skip rate of 4 for the global clips and a skip rate of 1 for the local clips. We perform an additional experiment, where we randomly select a skip rate for the Global clip from this set: $\{4,8,12,16\}$. We set the skip rate for the local clip to $1/4^{th}$ of the selected skip rate for global clip. This leads to a 2.6\% degradation from the fixed skip rate baseline.

\subsection{Contrastive Loss Temperature}

\begin{figure}[h]
    \centering
    \includegraphics[width=\columnwidth]{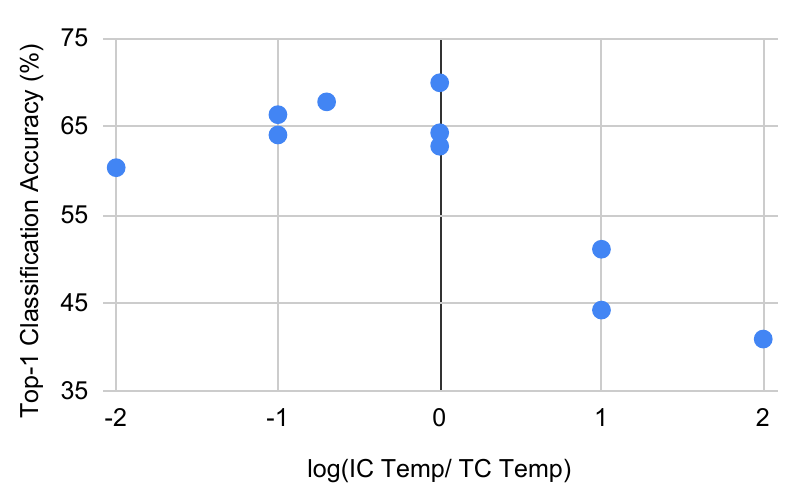}
    \caption{Effect of Contrastive Loss Temperature on downstream linear evaluation task. Using a lower temperature for temporal contrastive loss relative to the instance contrastive loss leads to poor results, as can seen from the right side of the origin in the chart.}
    \label{fig:tempabl}
\end{figure}

Prior works \cite{simclr, cvrl, pace_pred} demonstrate that a temperature of about $0.1$ is optimal for instance contrastive losses. We try different settings (Results in Table~\ref{table:tempabl}) to discover the optimal temperature for the temporal contrastive losses. We discover that $0.1$ is a good choice, and also that having it set to be higher than the temperature for the instance contrastive loss is very important. Setting the temperature for temporal contrastive loss lower than the instance contrastive loss leads to significantly poor performance (right of the origin in the chart), as can be seen in Figure~\ref{fig:tempabl}.

\begin{table}
\centering
\begin{tabular}{ccc} 
\hline

\hline

\hline\\[-3mm]
\begin{tabular}[c]{@{}c@{}}\textbf{\textbf{$\mathcal{L}_{IC}$}}\\~\textbf{Temperature}\end{tabular} &\begin{tabular}[c]{@{}c@{}}\textbf{\textbf{$\mathcal{L}_{LL}$, $\mathcal{L}_{GL}$}}\\~\textbf{Temperature}\end{tabular}  & \begin{tabular}[c]{@{}c@{}}\textbf{Linear Evaluation}\\~(\textbf{Top-1 Accuracy})\end{tabular}  \\ 
\hline

\hline

\hline\\[-3mm]
0.01                          & 0.01                         & 62.75                          \\ 
0.01                          & 0.1                          & 64.03                          \\ 
0.01                          & 1                            & 60.32                          \\ 
0.05                          & 0.05                         & 64.28                           \\ 
0.1                           & 0.01                         & 44.22                          \\ 
\textbf{0.1}                           & \textbf{0.1}                          & \textbf{69.93}                           \\ 
0.1                           & 0.5                          & 67.78                          \\ 
0.1                           & 1                            & 66.33                          \\ 
1                             & 0.01                         & 40.93                          \\ 
1                             & 0.1                          & 51.10    \\
\hline

\hline

\hline\\[-3mm]
\end{tabular}
\caption{Effect of temperatures used in contrastive losses.}
\label{table:tempabl}
\end{table}
\section{Baseline Verification}
\label{sec:bl_verification}
\noindent {\bf Instance Contrastive Baseline:} 
Our instance contrastive baseline using similar model architectures and input clip resolution ($112\times112$, 16 frames) achieves 71.3\% .  We use the same input size, architecture and data augmentations for our instance contrastive baseline and TCLR model. 

To ensure our instance contrastive baseline is correctly trained we also provide results from prior works in Table~\ref{tab:ic-baselines}

\begin{figure*}[h]
\begin{center}
 \includegraphics[width=0.9\linewidth]{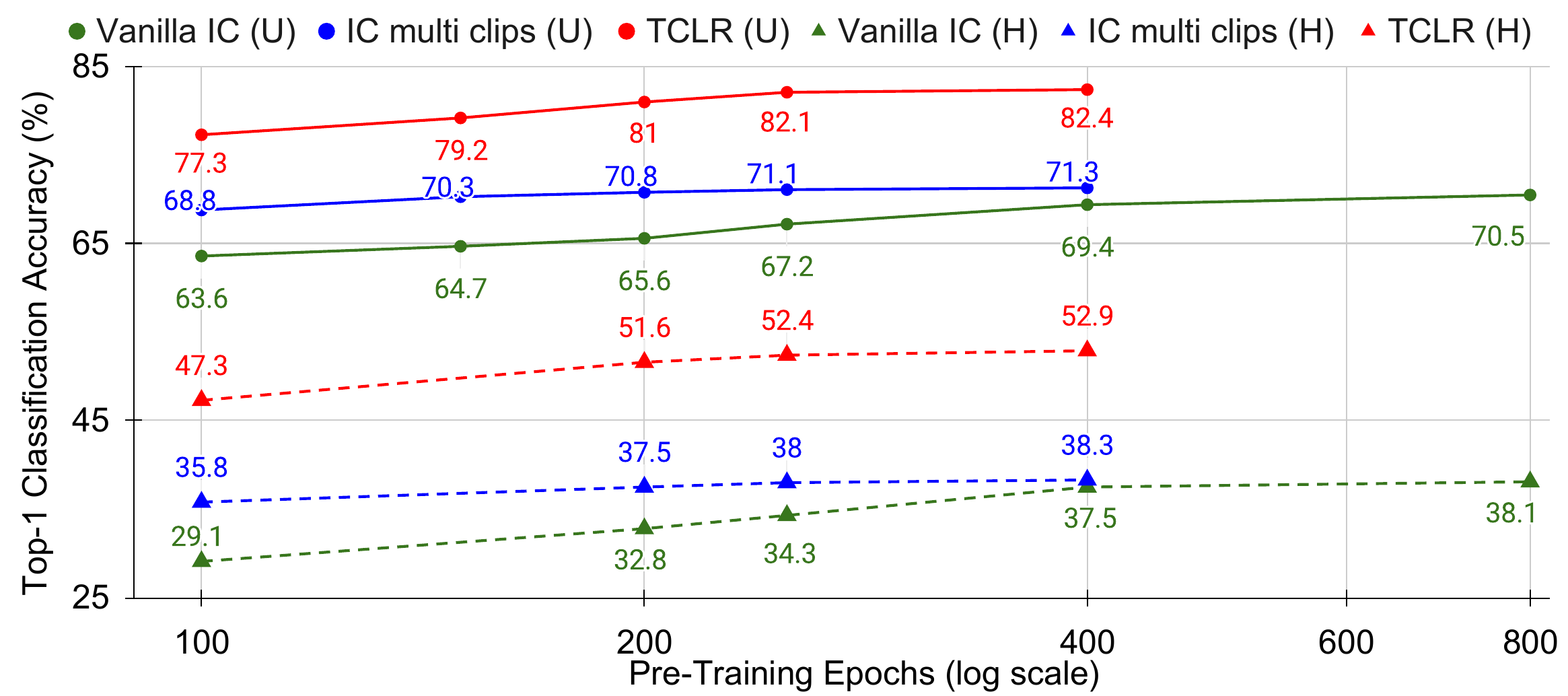}
\end{center}
\vspace{-4mm}
   \caption{\textbf{Effect of UCF101 pre-training duration} on downstream UCF101 and HMDB51 action recognition. \textit{TCLR outperforms the Vanilla IC baseline with 4-8$\times$ times longer pre-training}. (U) $\rightarrow$ UCF101 finetuning  and (H) $\rightarrow$ HMDB51 finetuning}
\label{fig:combined}
\end{figure*}

\begin{table}[h]
\centering
\begin{tabular}{lccc} 
 \hline

\hline

\hline\\[-3mm]
\textbf{Method}  & \textbf{Epochs} & \textbf{UCF} & \textbf{HMDB}  \\[1pt]
 \hline

\hline

\hline\\[-3mm]
IC~[50] & 200   & 70.0     & 39.9 \\ 

IC~[44] & 300   & 69.5   & - \\ 
IC~[7]  & 200$\ast$   & 73.0     & 40.6  \\ 
\textbf{TCLR}   & \textbf{50}   & \textbf{82.2}  & \textbf{51.9}  \\ 
TCLR            & 100           & 84.1   & 53.6 \\
 \hline

\hline

\hline\\[-3mm]
\end{tabular}
\caption{\textbf{Comparison of TCLR vs Vanilla IC with longer pre-training on Kinetics400} as reported in prior work. \\ $\ast$ email from authors}
\label{table:kinetics}
\end{table}

\begin{table}[h]
\small
\centering
\begin{tabular}{llcc}
\hline

\hline

\hline\\[-3mm]
\textbf{Source}	& \textbf{Architecture} & \textbf{Pretrain}	& \textbf{UCF101/HMDB51}  \\ \hline

\hline

\hline\\[-3mm]

\textbf{Our $\mathcal{L}_{IC}$}        & R3D18       & UCF101           & 71.3/ 38.3                         \\ 
\cite{pace_pred}                    & R(2+1)D18     & UCF101           & 67.3/ 28.6                         \\ 
\cite{idt}                          & R3D18       & K400      & 70.0/ 39.9                         \\ 
\cite{iclr21submitted_st_invariant} & R3D18       & K400      & 69.5/ $-$       \\ 
\cite{taco}                         & R3D18 & K400      & 73.0/ 40.6                         \\\hline

\hline

\hline\\[-3mm]

\end{tabular}
\vspace{-2mm}
\caption{\label{tab:ic-baselines} Instance Contrastive baselines reported in prior work match our baseline results.}
\vspace{-2mm}
\end{table}
\noindent {\bf Random Initialization Baseline:} Our baseline results with random initialization UCF101 finetuning (Top-1 classification accuracy \textbf{62.3\%}) which matches the prior results reported in \cite{memdpc} (\textbf{61.8\%}). We report this random initialization baseline in Table 4 of the main paper along with other downstream tasks.

\section{Benefit of Temporal Diversity}
\label{sec:temp_diversity}

\begin{figure}[h]
\vspace{2mm}
\begin{center}
 \includegraphics[width=1.0\linewidth]{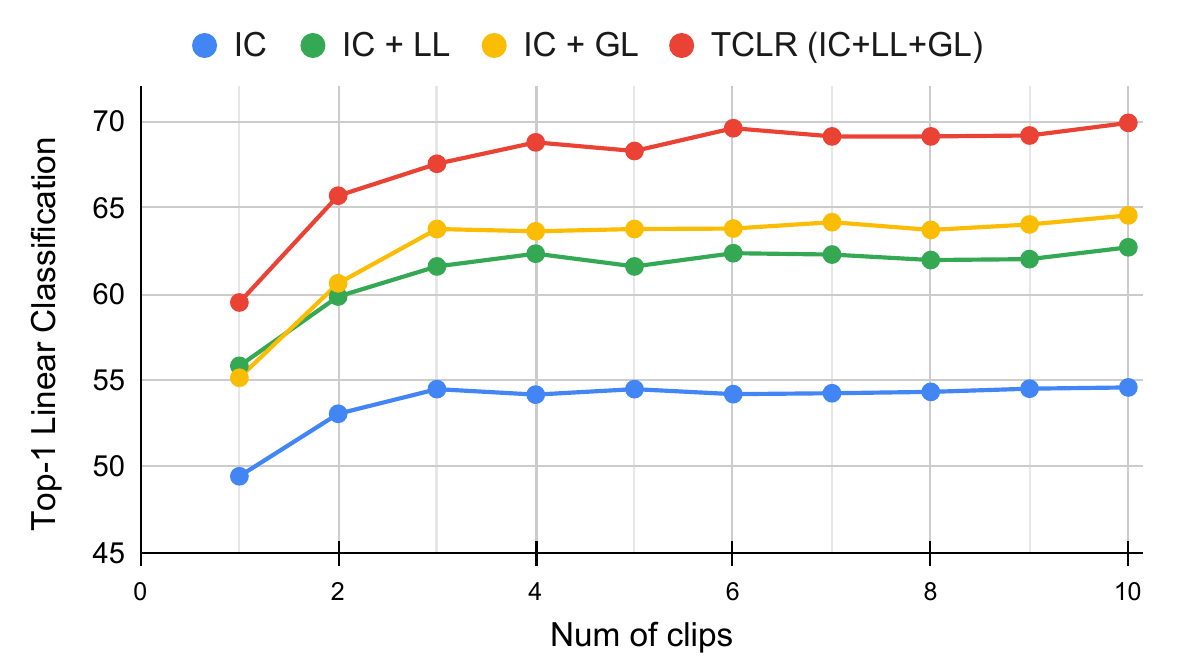}
\end{center}
\vspace{-5mm}
   \caption{Temporally distinct features learned by TCLR result in a significant improvement in Linear Evaluation on UCF101 (split-1) with increasing number of clips per video.}
\label{fig:lin_clips}
\end{figure}

\begin{figure}[h]
\vspace{2mm}
\begin{center}
 \includegraphics[width=1.0\linewidth]{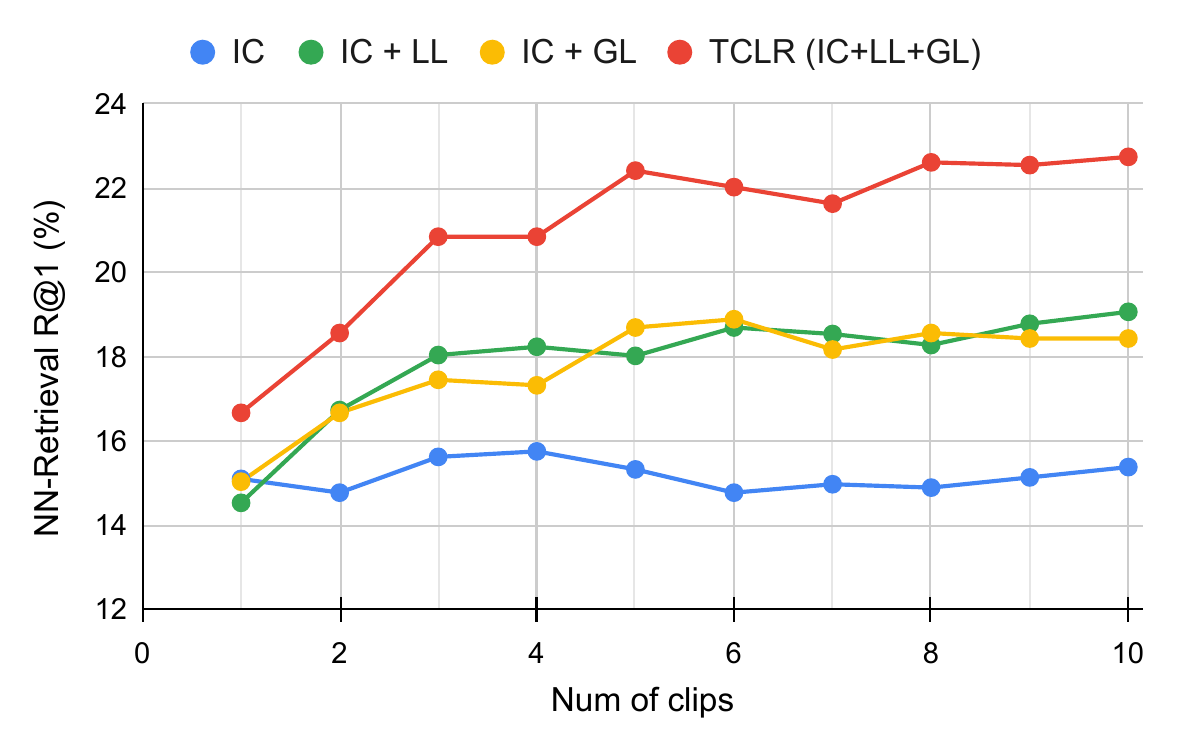}
\end{center}
\vspace{-5mm}
   \caption{Temporally distinct features learned by TCLR result in a significant improvement in NN Retrieval on HMDB51 with increasing number of clips per video.}
\label{fig:nnr_hmdb_clipwise}
\end{figure}

We discussed the effects of temporal diversity on NN-Retrieval task for UCF101 in the main paper and found that increased temporal diversity leads to an increased gap with respect to the IC baseline in the multi-clip setting. Here we provide similar results for other downstream tasks and datasets. Detailed results across 10 different clip counts for UCF101 linear evaluation can be seen in Figure~\ref{fig:lin_clips} and HMDB51 nearest neighbour retrieval can be seen in Figure~\ref{fig:nnr_hmdb_clipwise}. Summary results for 1 and 10 clip evaluation for 5 different tasks can be seen in Table~\ref{tab:summtask_clips}.

\begin{table*}
\vspace{0.3in}
\small
\centering
\begin{tabular}{llll|ll} 
\hline

\hline

\hline\\[-3mm]

& \multicolumn{3}{c}{\textbf{UCF101}} & \multicolumn{2}{|c}{\textbf{HMDB51}} \\
\cline{2-6}
        & \textbf{Linear Eval}          & \textbf{Finetuning}  & \textbf{NN-Retrieval}              & \textbf{Finetuning} & \textbf{NN-Retrieval}  \\ 
\hline

\hline

\hline\\[-3mm]

\textbf{Scratch} & 14.80/ 17.15 (+2.35)  & 56.85/ 62.30 (+5.45) &  8.21/ 9.77  (+1.56)  & 21.91/ 26.95 (+5.04)  & 4.77/ 4.05 (-0.72) \\ 
\textbf{IC}      & 49.43/ 54.58 (+5.15)  & 63.41/ 68.99 (+5.58) & 38.52/ 40.76 (+2.24)  & 33.39/ 38.32 (+4.93)  & 15.1/ 15.38 (+0.28)\\ 
\textbf{IC + LL} & 55.83/ 62.70 (+6.87)  & 66.85/ 74.38 (+7.53) & 43.64/ 51.10 (+7.46)  & 42.80/ 49.77 (+6.96)  & 14.53/ 19.07 (+4.54)\\ 
\textbf{IC + GL} & 55.14/ 64.55 (+9.41)  & 68.96/ 75.44 (+6.48) & 38.54/ 47.32 (+8.78)  & 39.67/ 47.87 (+8.20)  & 15.03/ 18.43 (+3.4)\\ 
\textbf{TCLR}    & 59.50/ 69.91 (+10.41) & 71.73/ 81.10 (+9.37) & 41.50/ 56.17 (+14.67) & 41.37/ 52.80 (+11.48) & 16.67/ 22.75 (+6.08)\\
\hline

\hline

\hline\\[-3mm]
\end{tabular}
\vspace{-3mm}
\caption{Effect of using multiple clips on different evaluation results. \textbf{1-clip/ 10-clip} performance on different downstream tasks.}
\vspace{0.3in}
\label{tab:summtask_clips}
\end{table*}

\section{Additional Comparison} 
\label{sec:additional_comparison}

Results from prior work which were excluded from the main paper are presented in Table~\ref{table:results}. The Table is divided into 3 sections. The first section includes results from papers which utilize visual modality but only provide results on specialized architectures or larger input sizes which are not widely used and hence cannot be compared fairly with other methods.  The second section includes works which utilize multi-modal data, beyond the visual domain, such as text and audio. The third section shows our results for temporal contrastive learning from the main paper to allow for comparison with these works, such a comparison however is not fair for reasons explained earlier. We also perform linear classification of Kinetics-400 using our self-supervised pre-trained R2+1D model, which gives 21.8\% top-1 accuracy.

\begin{table*}[h!]
\centering
\begin{tabular}{lclccccc} 
\hline

\hline

\hline\\[-3mm]
\textbf{Method} & $\mathbf{H/W \times T}$ & \textbf{Backbone}         & \textbf{Modality} & \begin{tabular}[c]{@{}l@{}}\textbf{Params}\\~($\times 10^6$)\end{tabular} & \textbf{Pretraining} & \multicolumn{1}{l}{\begin{tabular}[c]{@{}c@{}}\textbf{UCF101}\\\textbf{finetune}\end{tabular}} & \multicolumn{1}{l}{\begin{tabular}[c]{@{}c@{}}\textbf{HMDB51}\\\textbf{finetune}\end{tabular}}  \\ 
\hline

\hline

\hline\\[-3mm]%
\multicolumn{8}{l}{\textbf{Specialized Architectures}} \\
\hline
DVIM~\cite{dvim}          & $- \times -$        & R2D-18$^\ast$     & V & -            & UCF101 & 62.1  & 28.2     \\
DVIM~\cite{dvim}          & $- \times -$        & R2D-18$^\ast$     & V & -            & K400 & 64.0  & 29.7   \\
TCE~\cite{tce}                 & $224 \times -$        & R2D50& V          & 23.0         & K400 & 71.2  & 36.6   \\ 
O3N~\cite{fernando2017self}      & $227 \times -$        & AlexNet& V           & 61.0         & UCF101 & 60.3  & 32.5      \\ 
Shuffle-Learn~\cite{misra2016shuffle}              & $227 \times -$        & AlexNet& V           & 61.0         & UCF101 & 50.2  & 18.1       \\ 
B\"uchler et al.~\cite{buchler2018improving} & $227 \times -$        & AlexNet$^\ast$& V  & -            & UCF101 & 58.6  & 25.0        \\ 
Video Jigsaw~\cite{videojigsaw}             & $224 \times 25$       & CaffeNet& V          & 61.0         & UCF101 & 46.4  & -     \\
Video Jigsaw~\cite{videojigsaw}             & $224 \times 25$       & CaffeNet& V          & 61.0         & K400  & 55.4  & 27.0   \\
CBT~\cite{xdc}                  & $112 \times 16$       & S3D & V               & 20.0         & K600  & 79.5  & 44.6   \\

\hline
\multicolumn{8}{l}{\textbf{Multi-Modal Methods}} \\
\hline
AVTS~\cite{gdt} & $224 \times 25$  & I3D & V+A & - & K400 & 83.7 & 53.0 \\
AVTS~\cite{gdt} & $224 \times 25$  & MC3 & V+A & - & AudioSet & 89.0 & 61.6 \\
MIL-NCE~\cite{milnce} & $224 \times -$ & S3D & V+T & - & HowTo100M & 91.3 & 61.0 \\
GDT~\cite{gdt} & $112 \times 32$  & R(2+1)D & V+A & - & K400 & 89.3 & 60.0 \\
GDT~\cite{gdt} & $112 \times 32$  & R(2+1)D & V+A & - & IG65M & 95.2 & 72.8 \\
XDC~\cite{gdt} & $224 \times 32$  & R(2+1)D & V+A & - & K400 & 84.2 & 47.1 \\
\hline
\multicolumn{8}{l}{\textbf{TCLR}} \\
\hline
TCLR (Ours)   & $112 \times 16$       & R3D-18& V      & 13.5         & UCF101 & 82.4  & 52.9       \\ 
TCLR (Ours)   & $112 \times 16$       & R3D-18& V      & 13.5         & K400 & 84.1     & 53.6      \\ 
TCLR (Ours)   & $112 \times 16$       & R(2+1)D& V           & 14.4         & UCF101 & 82.8     & 53.6         \\ 
TCLR (Ours)   & $112 \times 16$       & R(2+1)D& V           & 14.4         & K400 &  84.3     & 54.2      \\ 
TCLR (Ours)   & $112 \times 16$       & C3D& V               & 27.7         & UCF101 & 76.1     & 48.6     \\
\hline

\hline

\hline\\[-3mm]
\end{tabular}

\caption{Additional Finetuning Results (average of 3 splits) for action classification on UCF101 and HMDB51 from prior work that were excluded from the main paper, along with our results for comparison. Please note that these results are not strictly comparable since the prior work uses specialized architectures and larger input sizes, which has significant effect on performance and require excessive computational resources. \textbf{V} denotes Video(RGB) modality, \textbf{A} denotes audio, and \textbf{T} denotes text modality. $^\ast -$ modified architecture}
\label{table:results}
\end{table*}

\section{Feature Slice Representation Similarity}
\label{sec:feature_slice_similarity}

\begin{figure}[h]
\vspace{-2mm}
\centering
    \begin{subfigure}{0.25\textwidth}
    \centering

        \includegraphics[width=\textwidth]{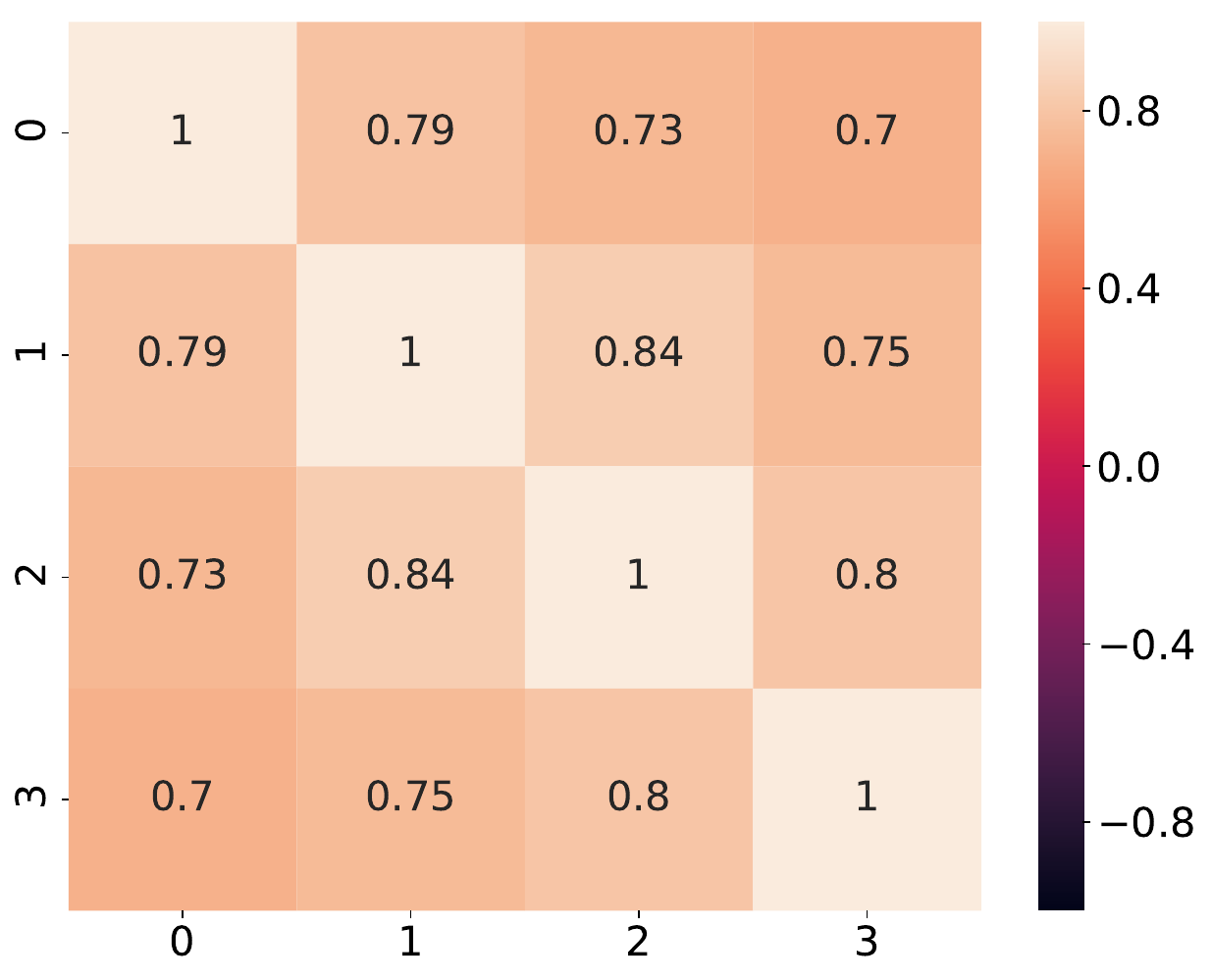}
        \caption{Random Initialization}
    \end{subfigure}
    \hfill
    \begin{subfigure}{0.25\textwidth}
        \includegraphics[width=\textwidth]{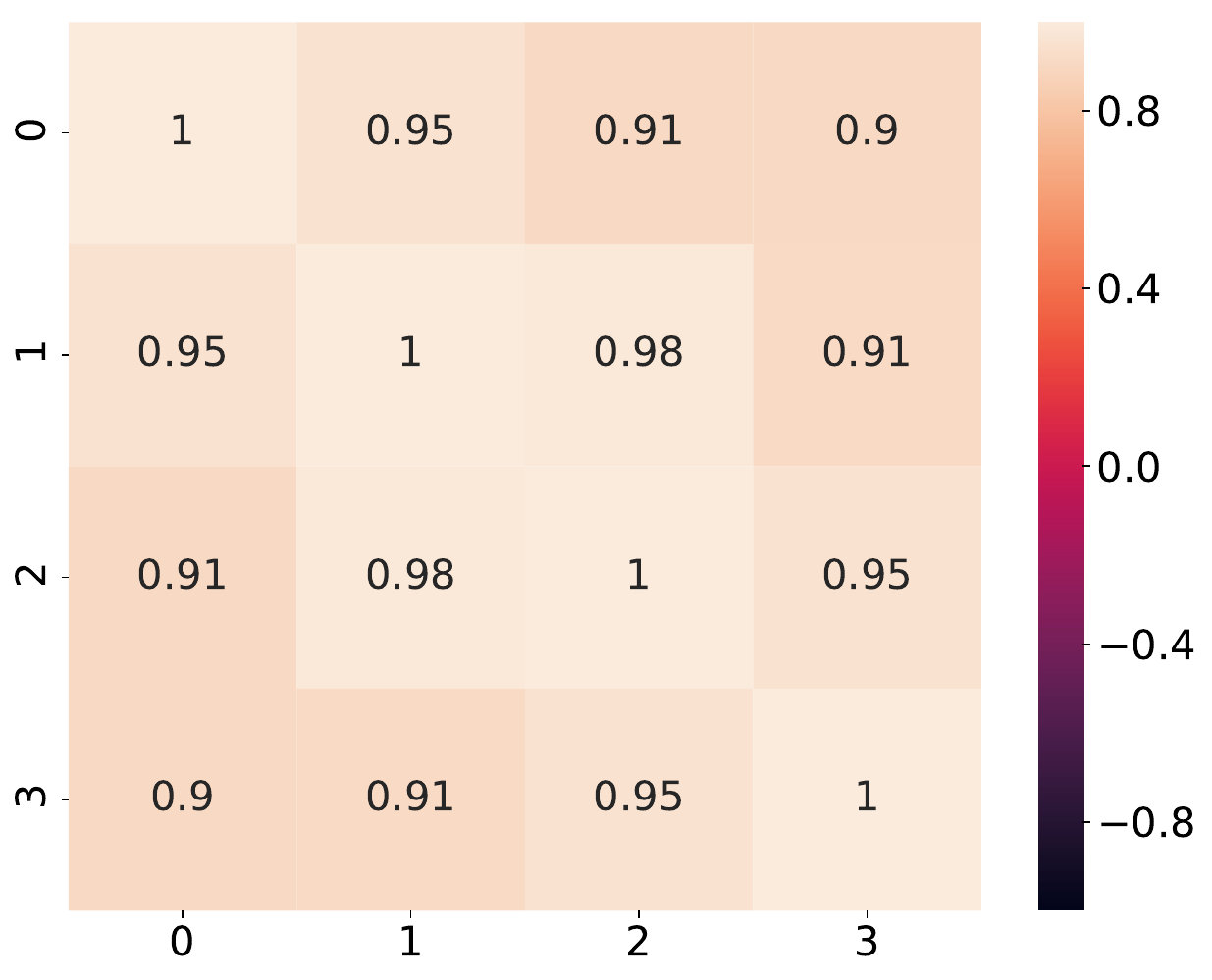}
        \caption{$\mathcal{L}_{IC}$}
    \end{subfigure}
    \hfill
    \begin{subfigure}{0.25\textwidth}
        \centering
        \includegraphics[width=\textwidth]{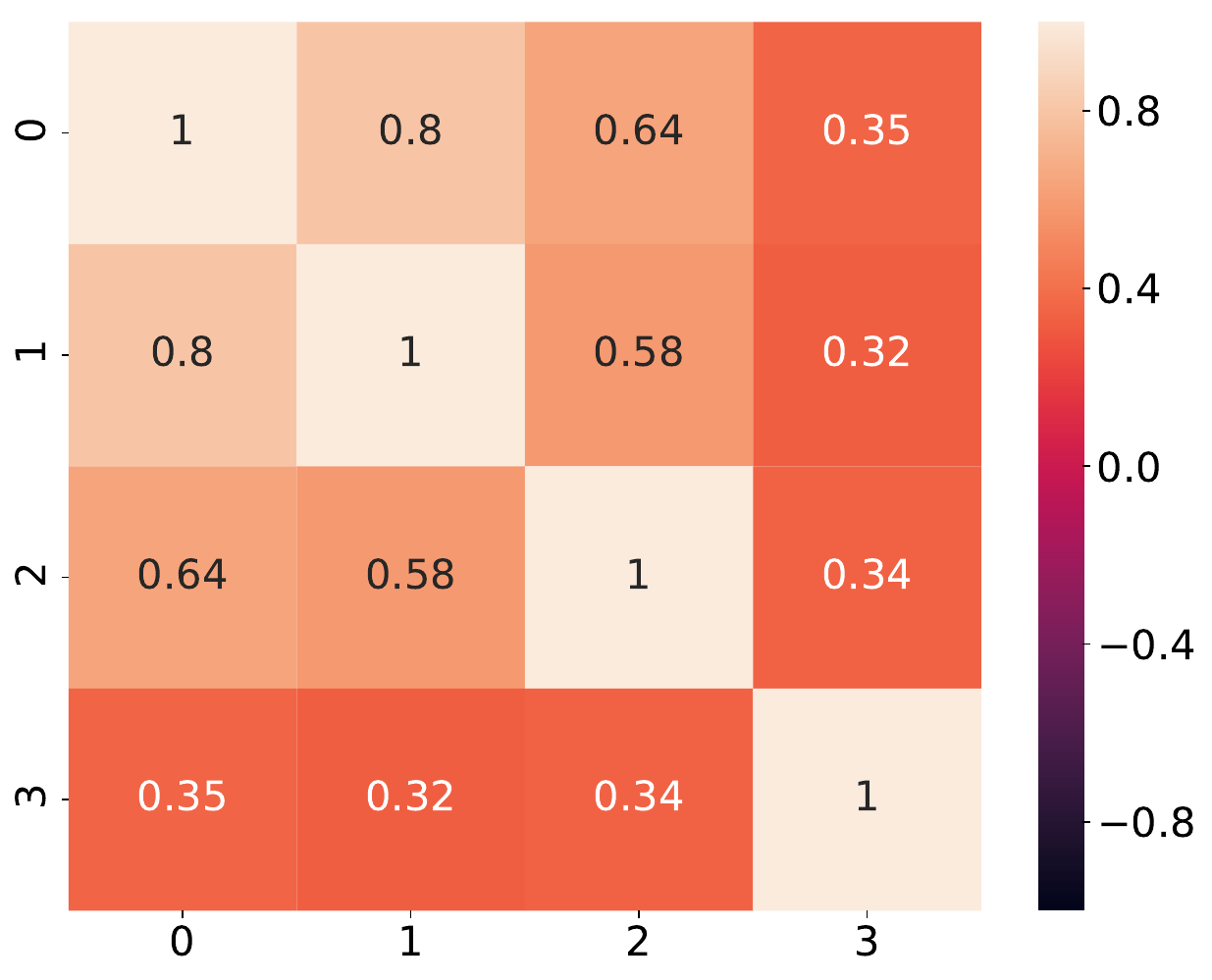}
        \caption{$\mathcal{L}_{IC}$ + $\mathcal{L}_{LL}$}
    \end{subfigure}
    \hfill
    \begin{subfigure}{0.25\textwidth}
       \centering
       \includegraphics[width=\textwidth]{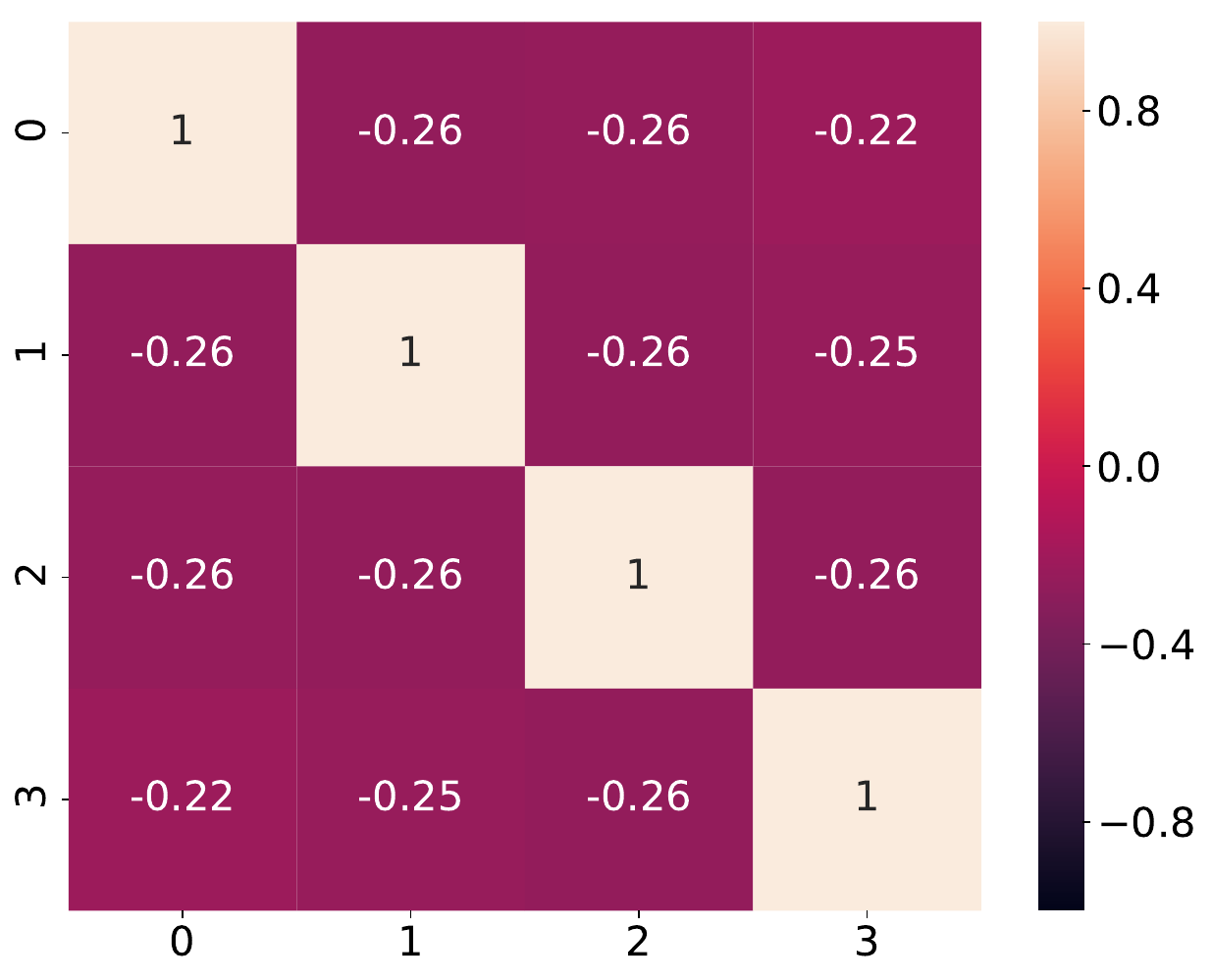}
        \caption{$\mathcal{L}_{IC}$ + $\mathcal{L}_{GL}$}
    \end{subfigure}
    \hfill
    \begin{subfigure}{0.25\textwidth}
        \centering
        \includegraphics[width=\textwidth]{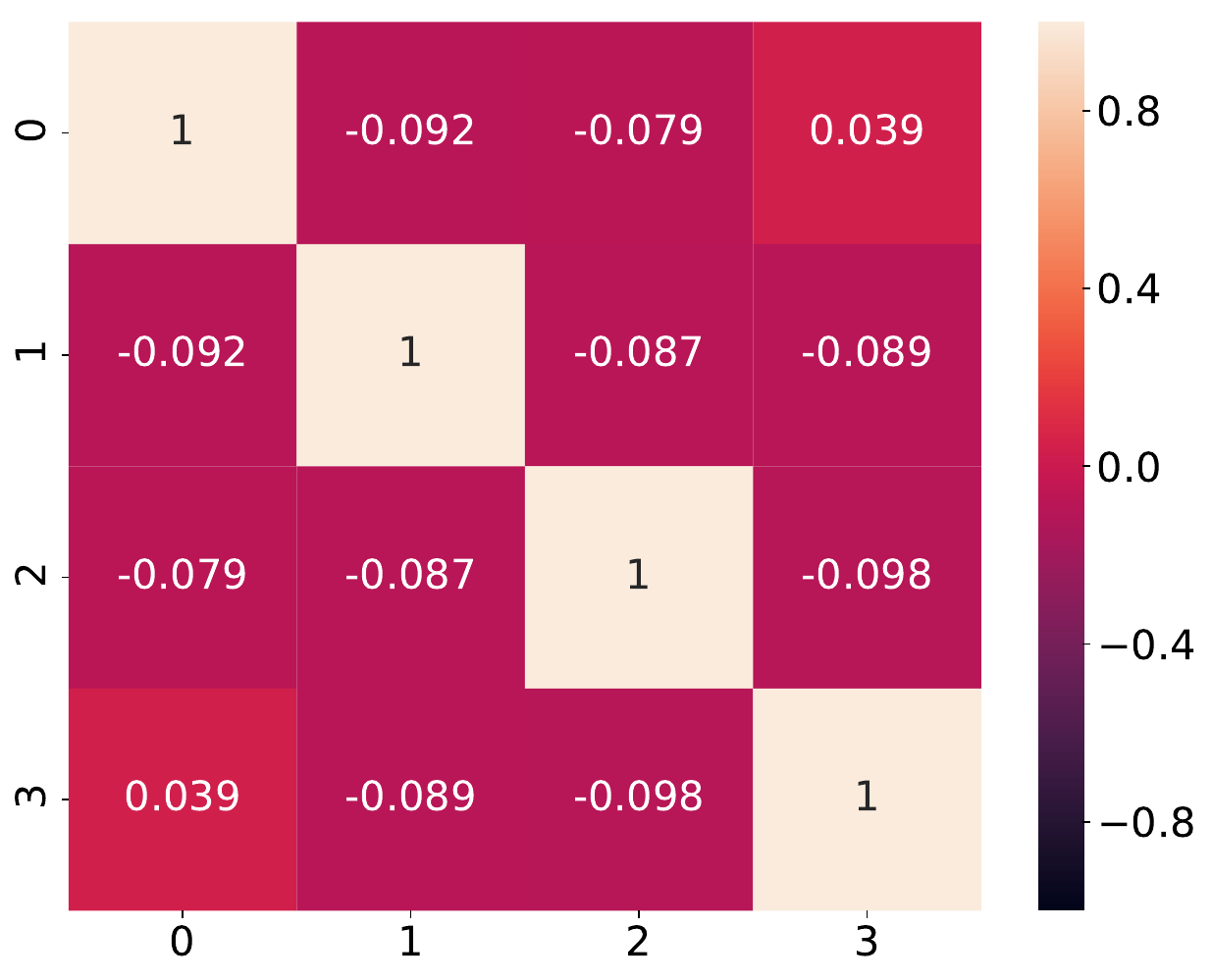}
        \caption{TCLR}
    \end{subfigure}
    
    \vspace{-3mm}
    \caption{Cosine similarity between timesteps of the feature map learned using different losses. It can be observed that temporal distinctiveness learned features increases with the addition of temporal contrastive losses.}    \label{fig:timecontrast}
    \vspace{0.5in}
\end{figure}

The goal of this visualization (see Figure~\ref{fig:timecontrast} is to verify if the model learns temporal diversity in the feature map as intended by the TCLR framework. With a long global clip (16 frames with a skip rate of 4, hence covering 64 frames), representations for 4 different timesteps of the video feature map were obtained using the video encoder and projection head. Cosine similarity was computed between timesteps for each video and then averaged across the dataset. As can be observed, $\mathcal{L}_{GL}$ has the maximum impact on increasing within clip feature diversity, while $\mathcal{L}_{LL}$ also induces more diversity than the IC baseline. 

\section{Qualitative: Nearest Neighbour Retrieval}
\label{sec:qualitative}
\begin{figure*}[t]
\begin{center}
  \includegraphics[width=\linewidth]{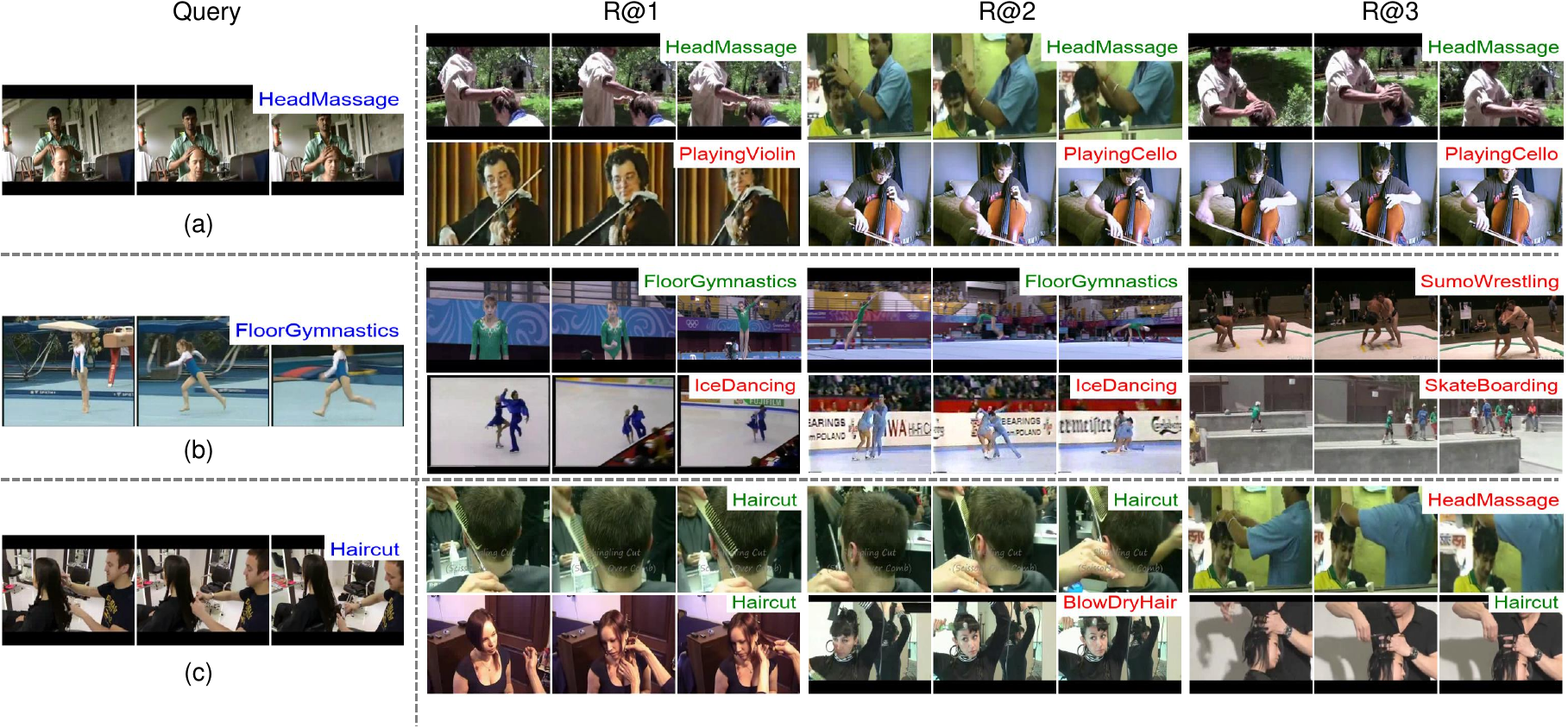}
\end{center}
\vspace{-6mm}
  \caption{\textbf{Qualitative Results:} Nearest Neighbour Video Retrieval results on UCF101. For each query, the upper row show videos retrieved by our method and the lower row shows results from standard instance contrastive loss.}
\label{fig:retrieval_qual}
\vspace{0.2in}
\end{figure*}

Qualitative results for nearest neighbour retrieval task on UCF101 are presented in Figure~\ref{fig:retrieval_qual}. These results are obtained using self-supervised pretrained models, without any supervised training with labels.

\begin{figure*}[t]
\vspace{1.5in}
\centering
\begin{center}
    \includegraphics[clip, trim=2cm 1.2cm 0.5cm 1cm,width=0.33\textwidth]{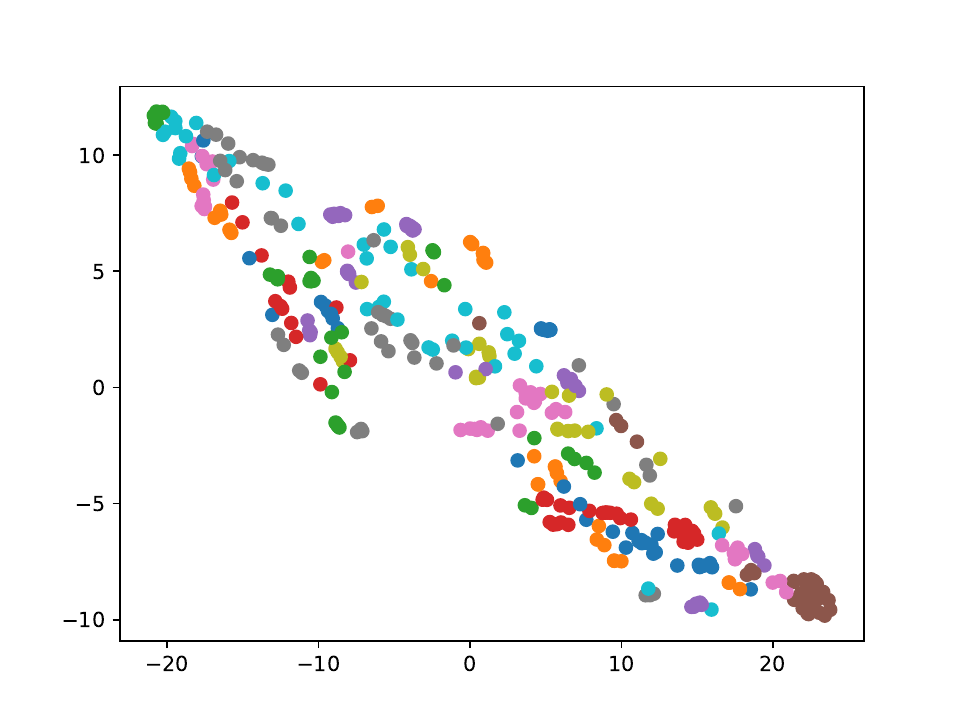}
    \includegraphics[clip, trim=2cm 1.2cm 0.5cm 1cm,width=0.33\textwidth]{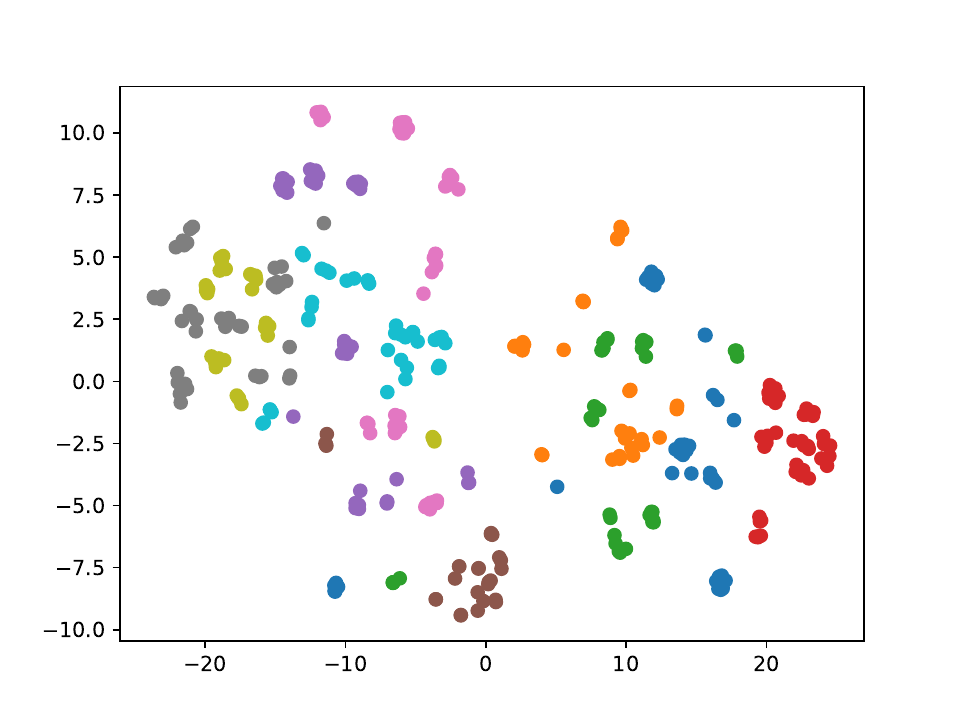}
    \includegraphics[clip, trim=2cm 1.2cm 0.5cm 1cm,width=0.33\textwidth]{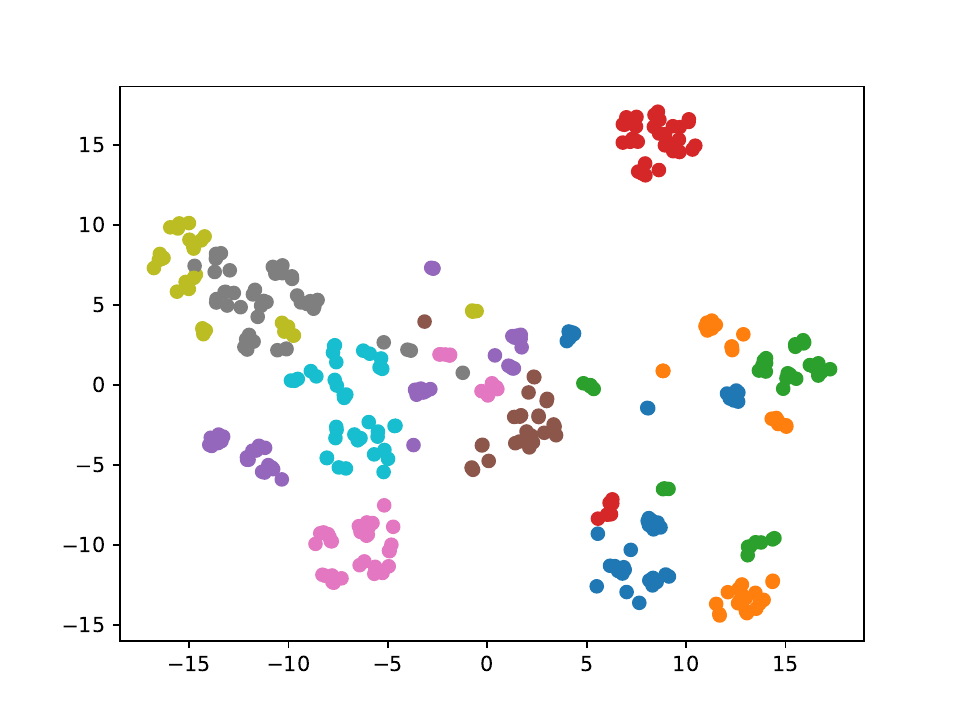}
\end{center}
    \caption{t-SNE visualization of Randomly Initialized (left) features; features learned using Instance Contrastive loss (center);  and TCLR (right),  for 10  randomly chosen action classes from UCF101 Test set:  \textcolor{FieldHockeyPenalty}{FieldHockeyPenalty}, \textcolor{GolfSwing}{GolfSwing}, \textcolor{SoccerJuggling}{SoccerJuggling}, \textcolor{SoccerPenalty}{SoccerPenalty}, \textcolor{PlayingGuitar}{PlayingGuitar}, \textcolor{PlayingPiano}{PlayingPiano}, \textcolor{PlayingSitar}{PlayingSitar}, \textcolor{ApplyEyeMakeup}{ApplyEyeMakeup}, \textcolor{ApplyLipstick}{ApplyLipstick} and \textcolor{BlowDryHair}{BlowDryHair}.  TCLR results in more coherent clusters compared to the instance contrastive lossand is able to discriminate between similar classes with fine-grained differences such as \textcolor{ApplyEyeMakeup}{ApplyEyeMakeup}, \textcolor{ApplyLipstick}{ApplyLipstick} and \textcolor{BlowDryHair}{BlowDryHair}.}
    \label{fig:tsnecompare}
    \vspace{0.8in}
\end{figure*}

\section{Visualization of Learned Representations}
\label{sec:tsne}

\vspace{-0.1in}

\begin{figure*}[h!]
\vspace{-2mm}
\centering
\begin{subfigure}[t]{0.49\textwidth}
    \includegraphics[clip, trim=2cm 1.2cm 0.5cm 1cm, width=0.49\textwidth]{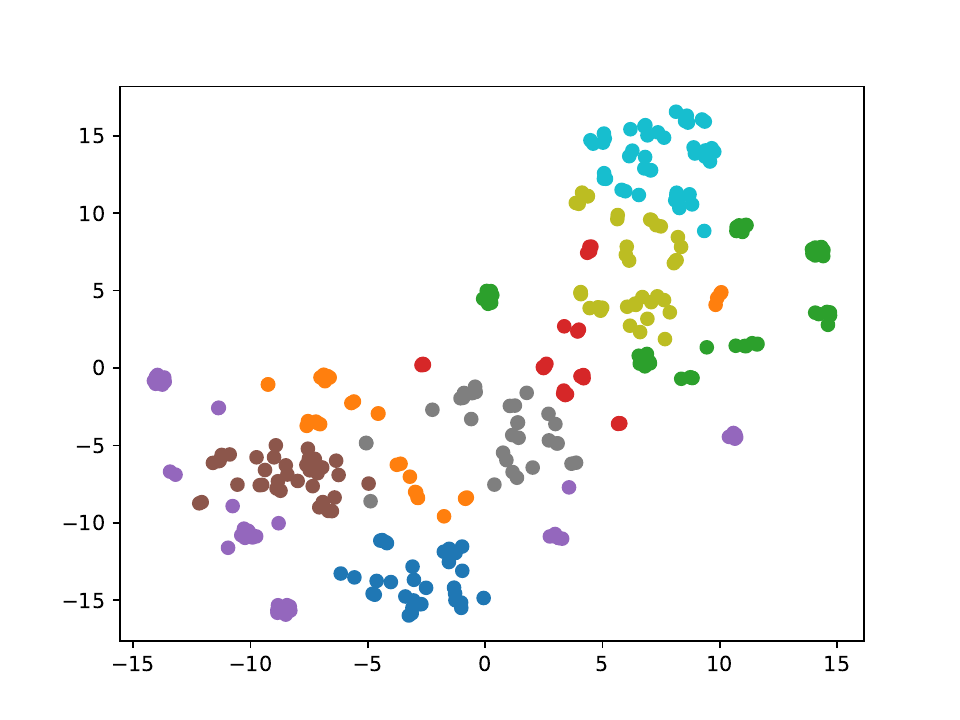}
    \includegraphics[clip, trim=2cm 1.2cm 0.5cm 1cm,width=0.49\textwidth]{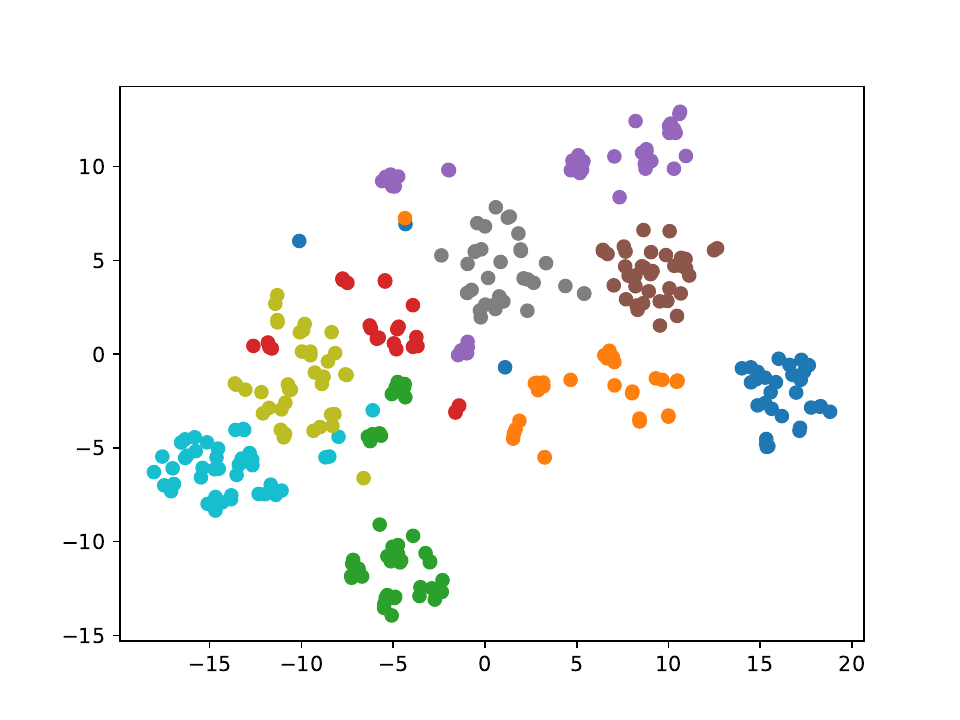}
    \caption{\textcolor{c0}{MilitaryParade}, \textcolor{c1}{TrampolineJumping}, \textcolor{c2}{PlayingSitar}, \textcolor{c3}{PlayingViolin}, \textcolor{c4}{FieldHockeyPenalty}, \textcolor{c5}{HighJump}, \textcolor{c6}{StillRings}, \textcolor{c7}{BlowDryHair}, \textcolor{c8}{ApplyEyeMakeup}}
\end{subfigure}
\hfill%
\begin{subfigure}[t]{0.49\textwidth}
    \includegraphics[clip, trim=2cm 1.2cm 0.5cm 1cm, width=0.49\textwidth]{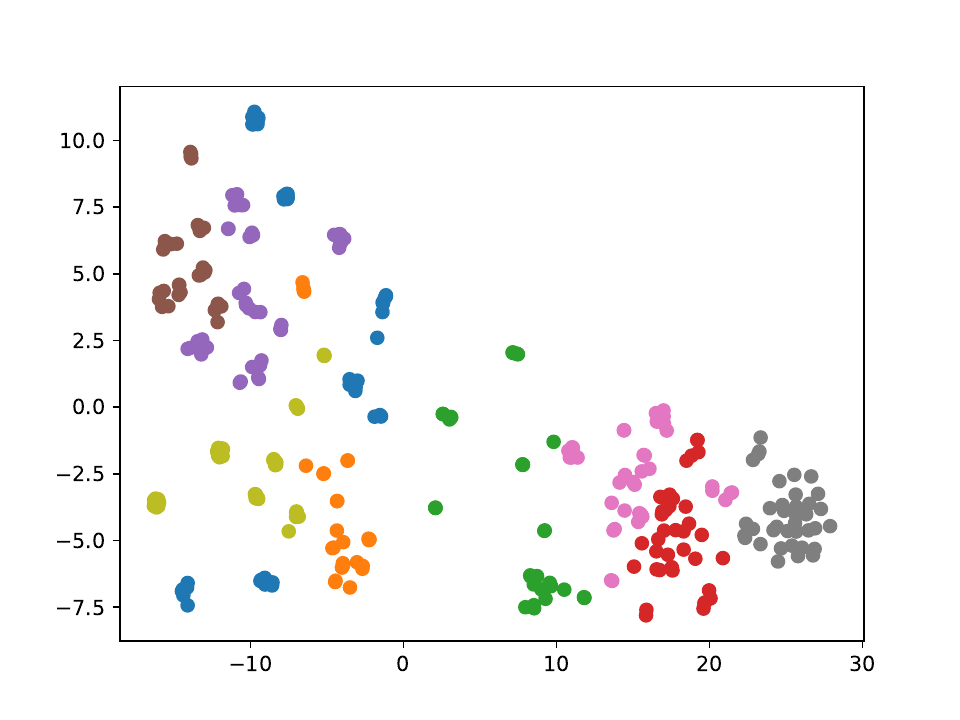}
    \includegraphics[clip, trim=2cm 1.2cm 0.5cm 1cm,width=0.49\textwidth]{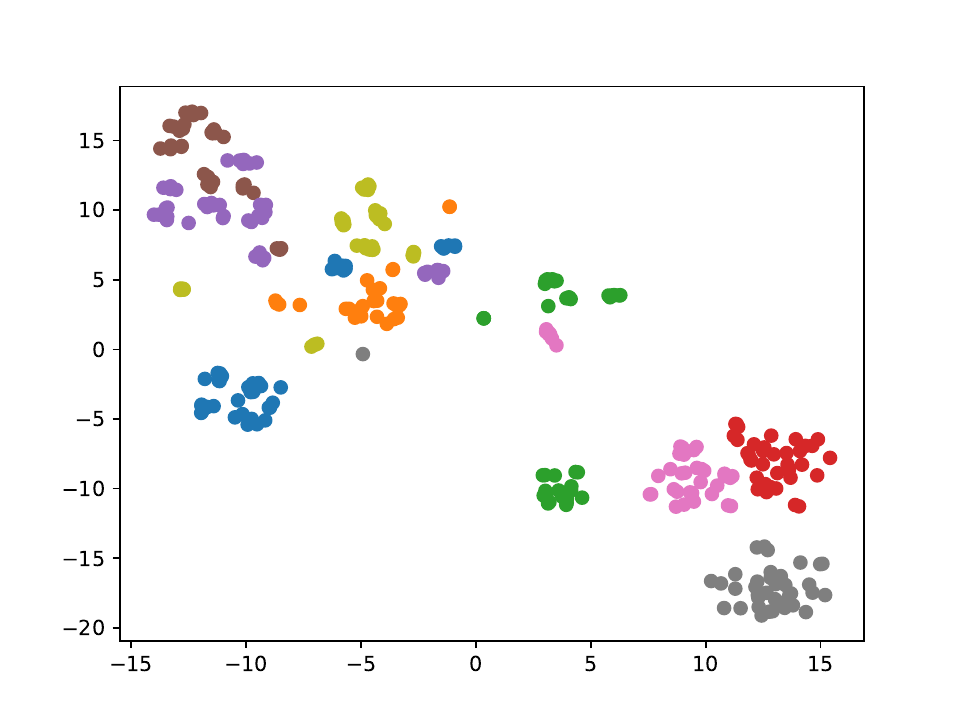}
    \caption{\textcolor{c0}{PlayingSitar}, \textcolor{c1}{PlayingPiano}, \textcolor{c2}{GolfSwing}, \textcolor{c3}{HighJump}, \textcolor{c4}{ShavingBeard}, \textcolor{c5}{ApplyLipstick}, \textcolor{c6}{PoleVault}, \textcolor{c7}{BasketballDunk}, \textcolor{c8}{Typing}, \textcolor{c9}{MixingBatter}}
\end{subfigure}
\begin{subfigure}[t]{0.49\textwidth}
    \includegraphics[clip, trim=2cm 1.2cm 0.5cm 1cm, width=0.49\textwidth]{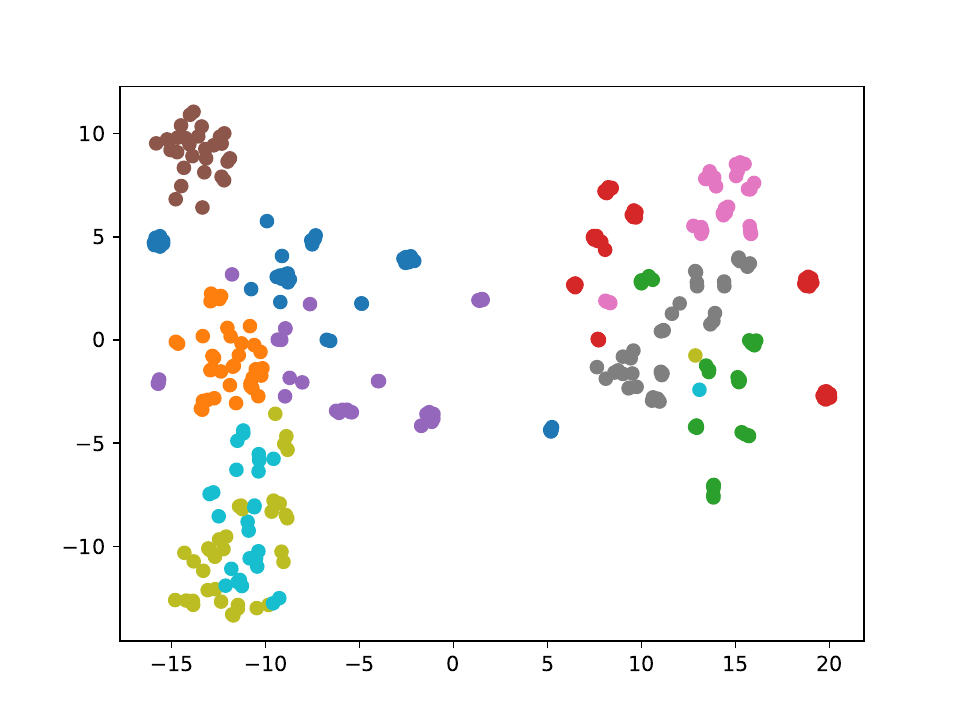}
    \includegraphics[clip, trim=2cm 1.2cm 0.5cm 1cm,width=0.49\textwidth]{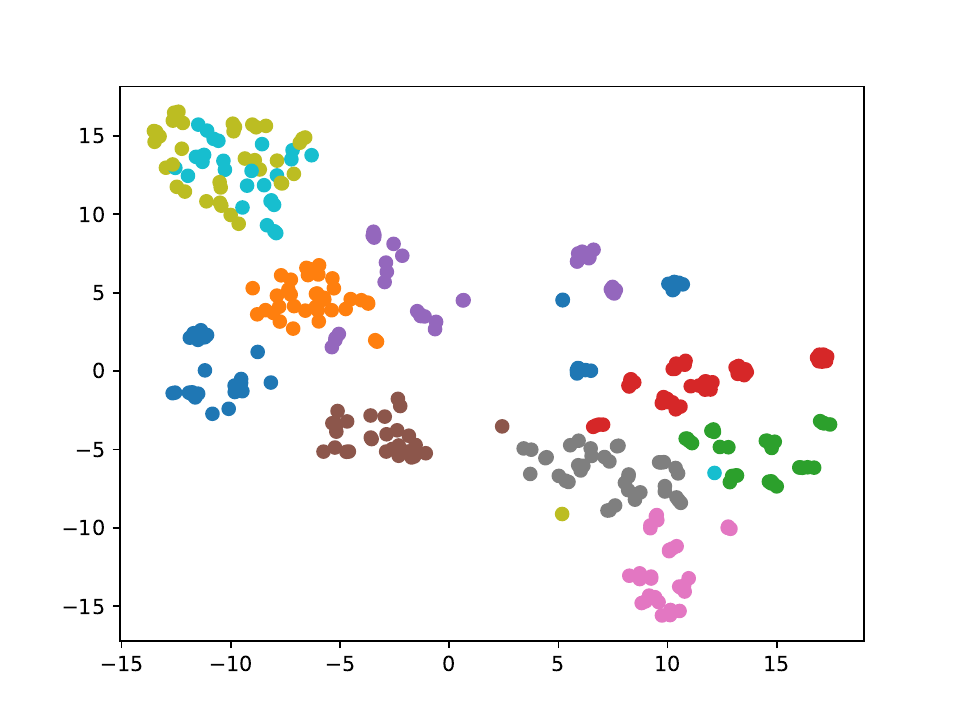}
    \caption{\textcolor{c0}{FieldHockeyPenalty}, \textcolor{c1}{HighJump}, \textcolor{c2}{Knitting}, \textcolor{c3}{Typing}, \textcolor{c4}{Basketball}, \textcolor{c5}{UnevenBars}, \textcolor{c6}{ApplyLipstick}, \textcolor{c7}{BlowDryHair}, \textcolor{c8}{FrontCrawl}, \textcolor{c9}{BreastStroke}}
\end{subfigure}
\hfill%
\begin{subfigure}[t]{0.49\textwidth}
    \includegraphics[clip, trim=2cm 1.2cm 0.5cm 1cm, width=0.49\textwidth]{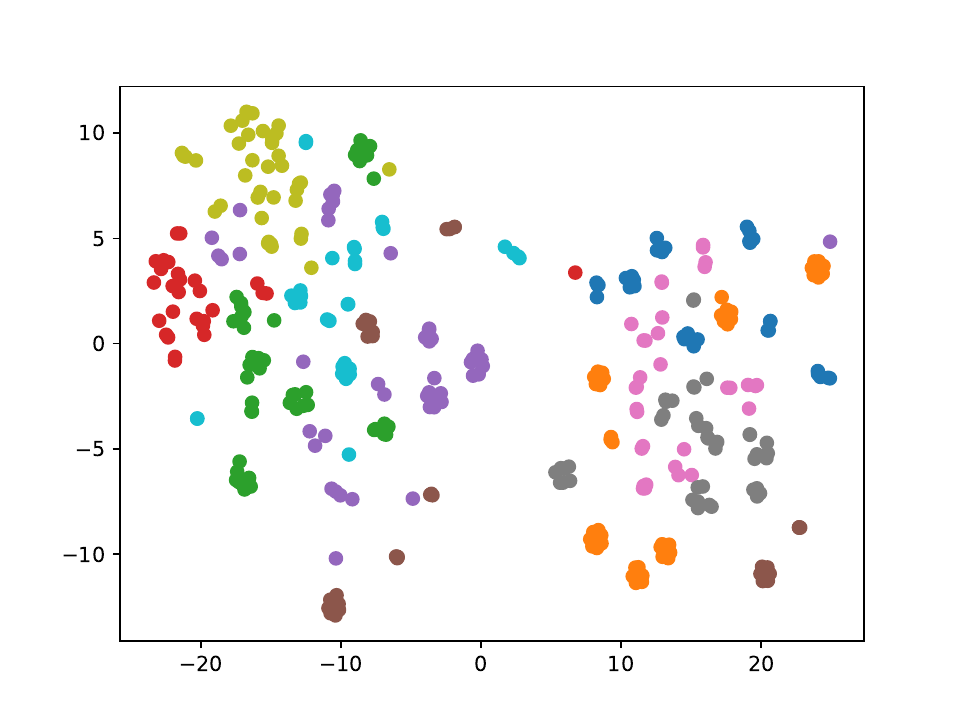}
    \includegraphics[clip, trim=2cm 1.2cm 0.5cm 1cm,width=0.49\textwidth]{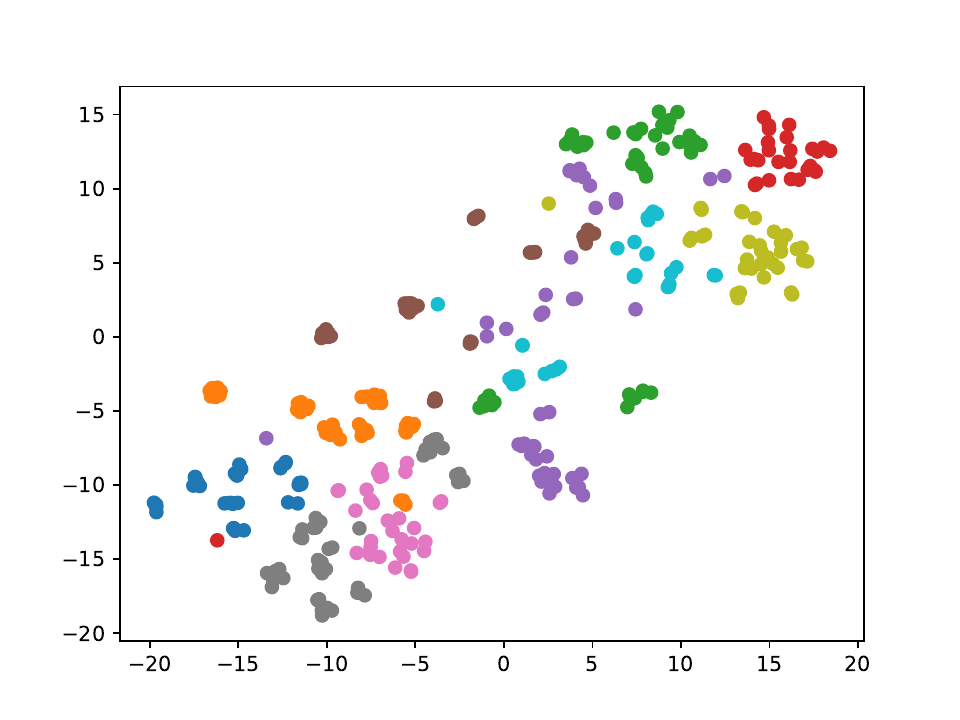}
    \caption{\textcolor{c0}{Knitting}, \textcolor{c1}{Typing}, \textcolor{c2}{Diving}, \textcolor{c3}{BreastStroke}, \textcolor{c4}{HammerThrow}, \textcolor{c5}{Nunchucks}, \textcolor{c6}{Haircut}, \textcolor{c7}{ShavingBeard}, \textcolor{c8}{Biking}, \textcolor{c9}{TrampolineJumping}}
\end{subfigure}
\begin{subfigure}[t]{0.49\textwidth}
    \includegraphics[clip, trim=2cm 1.2cm 0.5cm 1cm, width=0.49\textwidth]{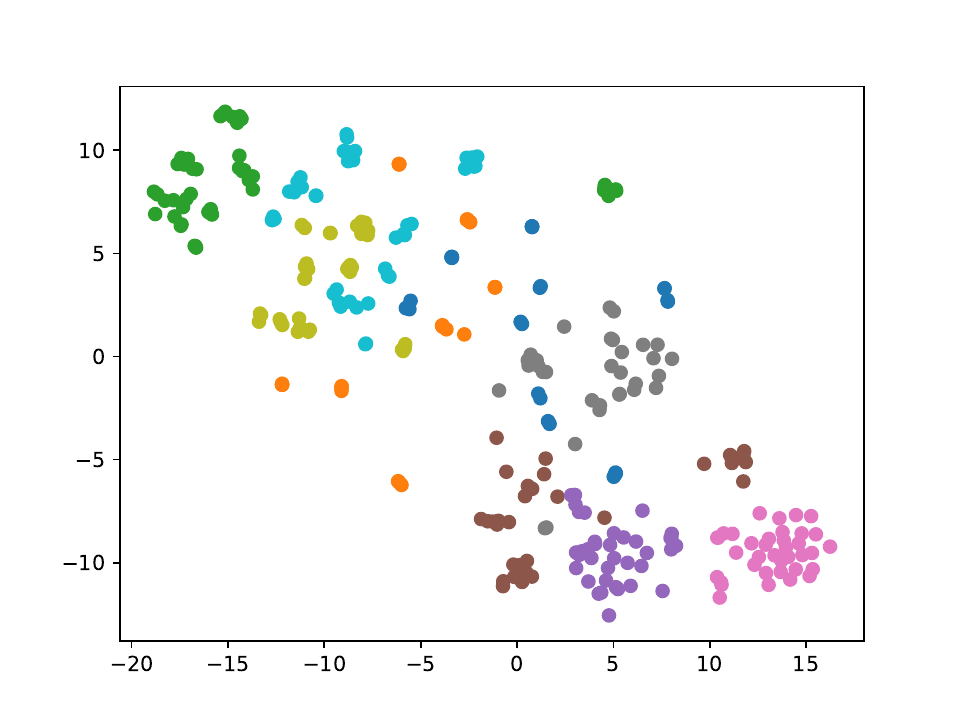}
    \includegraphics[clip, trim=2cm 1.2cm 0.5cm 1cm,width=0.49\textwidth]{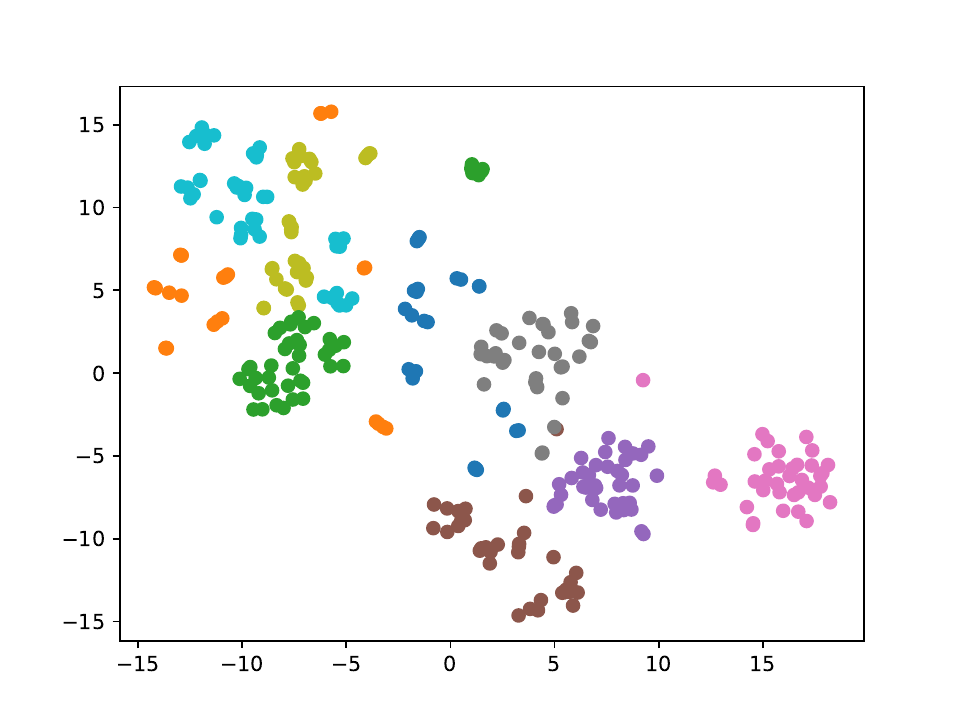}
    \caption{\textcolor{c0}{PullUps}, \textcolor{c1}{PushUps}, \textcolor{c2}{WritingOnBoard}, \textcolor{c3}{WalkingWithADog}, \textcolor{c4}{HighJump}, \textcolor{c5}{VolleyballSpiking}, \textcolor{c6}{BasketballDunk}, \textcolor{c7}{StillRings}, \textcolor{c8}{BrushingTeeth}, \textcolor{c9}{ShavingBeard}}
\end{subfigure}
\hfill%
\begin{subfigure}[t]{0.49\textwidth}
    \includegraphics[clip, trim=2cm 1.2cm 0.5cm 1cm, width=0.49\textwidth]{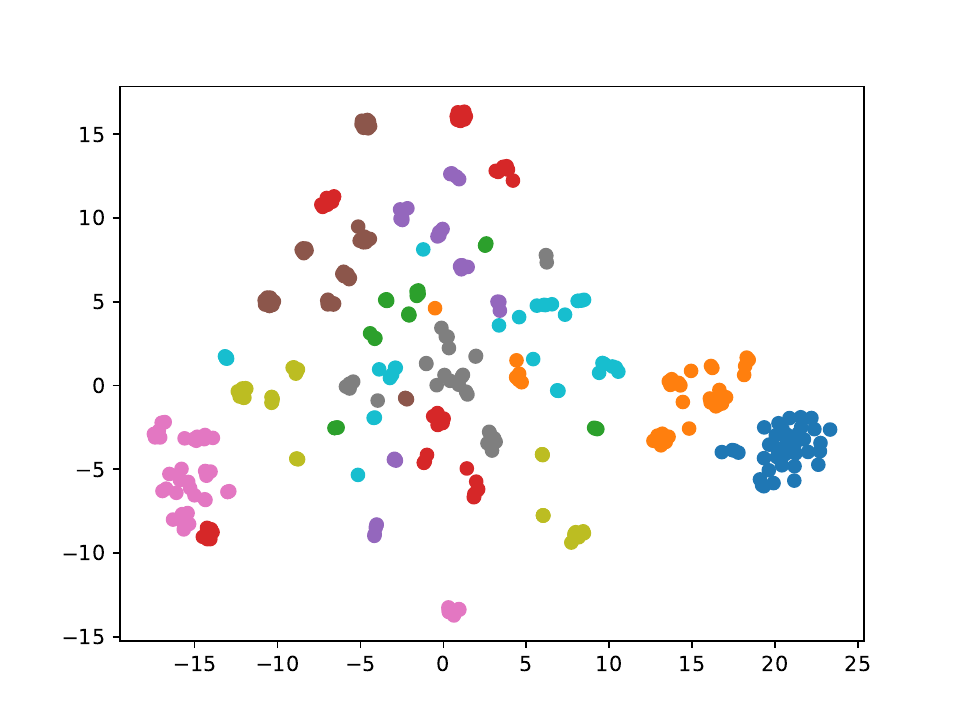}
    \includegraphics[clip, trim=2cm 1.2cm 0.5cm 1cm,width=0.49\textwidth]{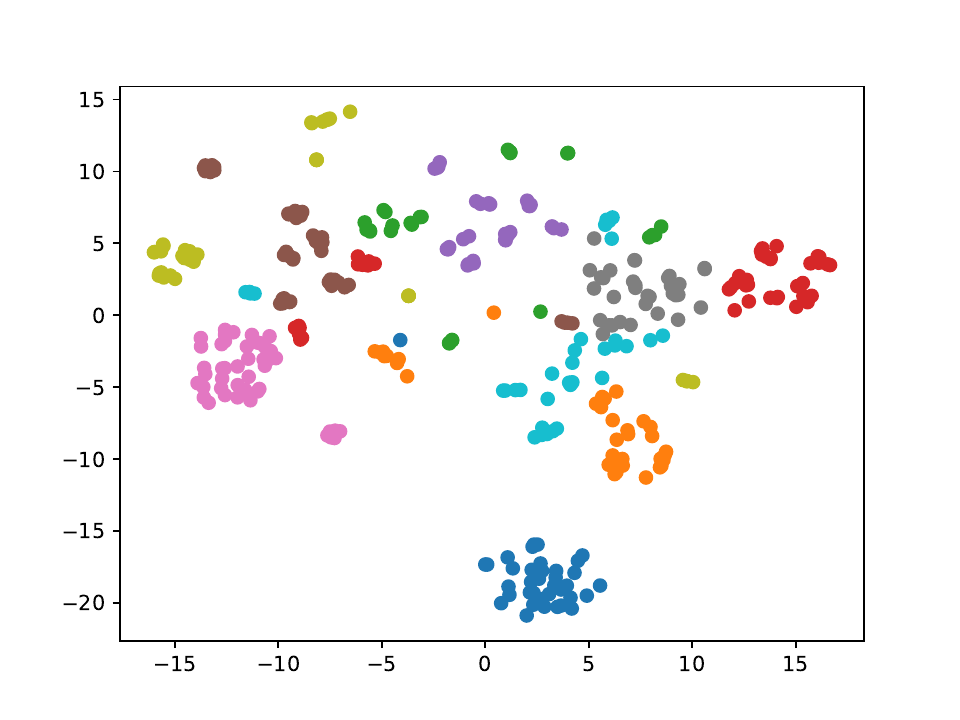}
    \caption{\textcolor{c0}{BasketballDunk}, \textcolor{c1}{ParallelBars}, \textcolor{c2}{PlayingViolin}, \textcolor{c3}{PlayingSitar}, \textcolor{c4}{Knitting}, \textcolor{c5}{Typing}, \textcolor{c6}{WritingOnBoard}, \textcolor{c7}{BlowingCandles}, \textcolor{c8}{JumpingJack}, \textcolor{c9}{HandstandWalking}}
\end{subfigure}
\begin{subfigure}[t]{0.49\textwidth}
    \includegraphics[clip, trim=2cm 1.2cm 0.5cm 1cm, width=0.49\textwidth]{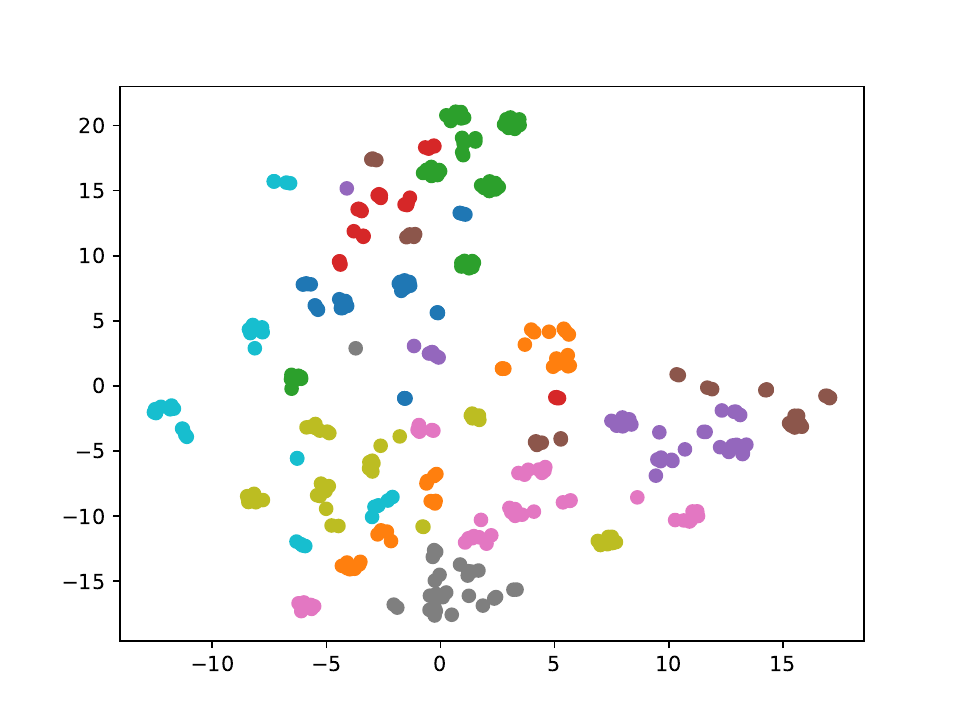}
    \includegraphics[clip, trim=2cm 1.2cm 0.5cm 1cm,width=0.49\textwidth]{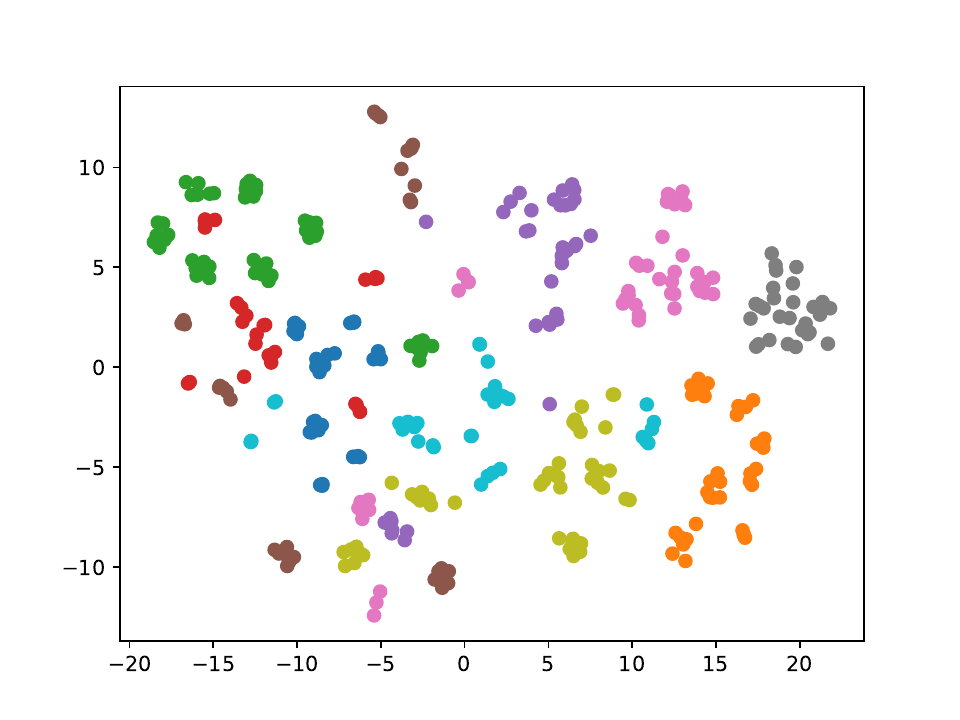}
    \caption{\textcolor{c0}{WallPushups}, \textcolor{c1}{Lunges}, \textcolor{c2}{PlayingFlute}, \textcolor{c3}{PlayingViolin}, \textcolor{c4}{ParallelBars}, \textcolor{c5}{PommelHorse}, \textcolor{c6}{Diving}, \textcolor{c7}{BreastStroke}, \textcolor{c8}{Shotput}, \textcolor{c9}{CricketBowling}}
\end{subfigure}
\hfill%
\begin{subfigure}[t]{0.49\textwidth}
    \includegraphics[clip, trim=2cm 1.2cm 0.5cm 1cm, width=0.49\textwidth]{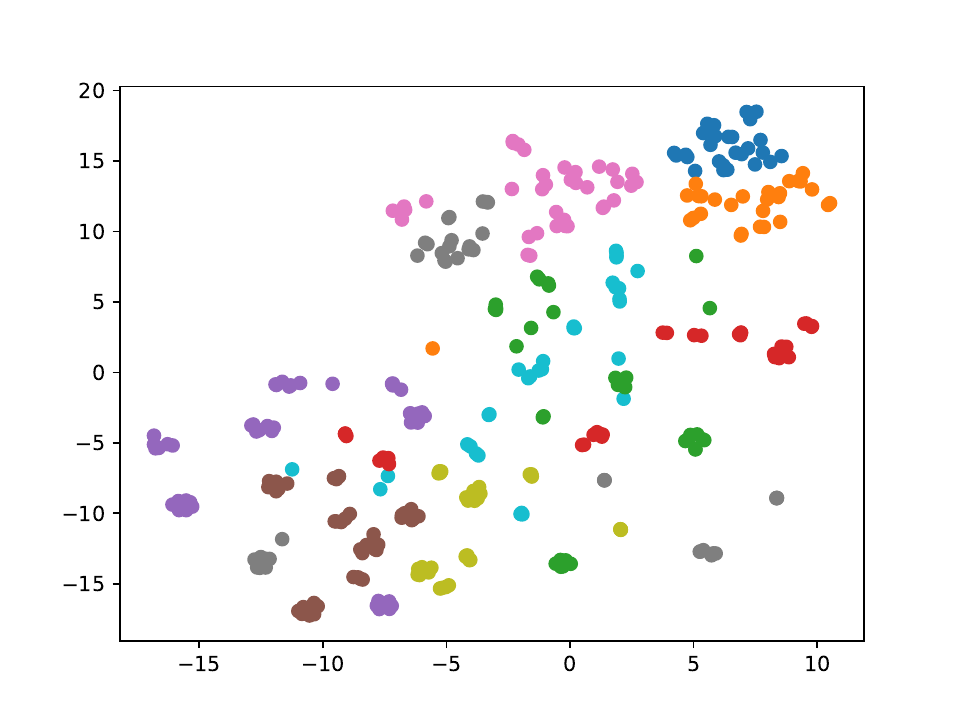}
    \includegraphics[clip, trim=2cm 1.2cm 0.5cm 1cm,width=0.49\textwidth]{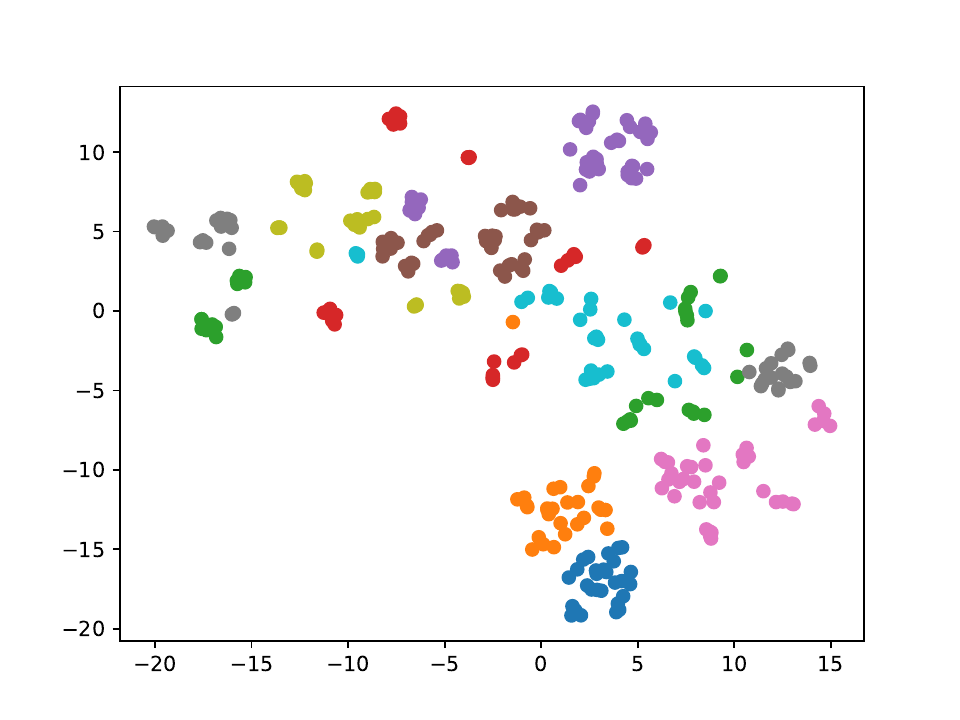}
    
    \caption{\textcolor{c0}{Skijet}, \textcolor{c1}{BreastStroke}, \textcolor{c2}{Basketball}, \textcolor{c3}{PommelHorse}, \textcolor{c4}{PlayingSitar}, \textcolor{c5}{PlayingDaf}, \textcolor{c6}{FrisbeeCatch}, \textcolor{c7}{GolfSwing}, \textcolor{c8}{WallPushups}, \textcolor{c9}{HandstandWalking}}
\end{subfigure}
\begin{subfigure}[t]{0.49\textwidth}
    \includegraphics[clip, trim=2cm 1.2cm 0.5cm 1cm, width=0.49\textwidth]{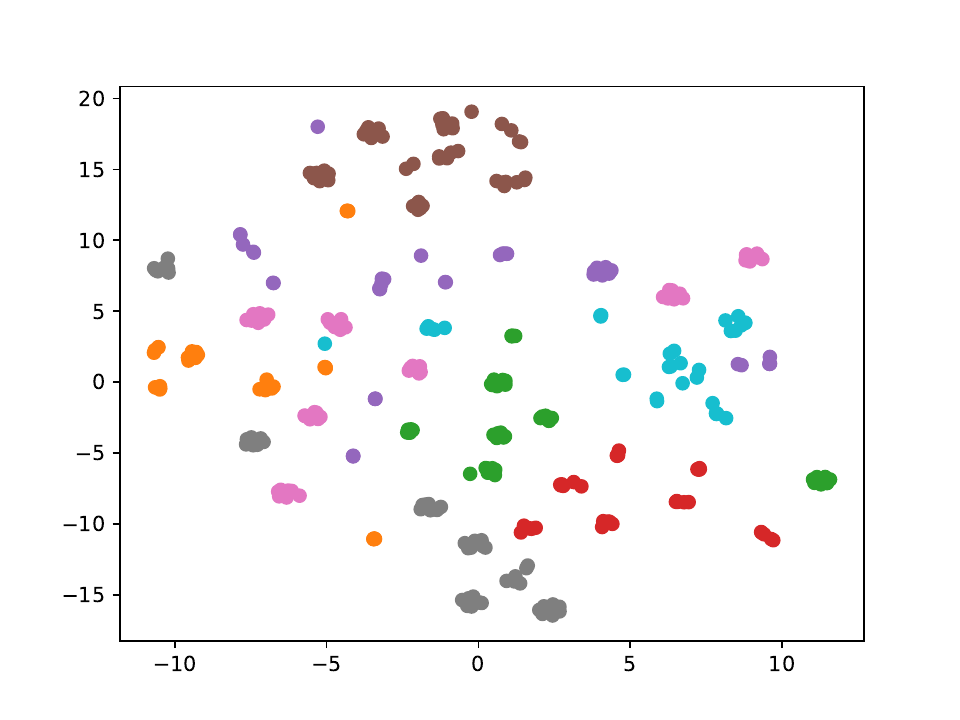}
    \includegraphics[clip, trim=2cm 1.2cm 0.5cm 1cm,width=0.49\textwidth]{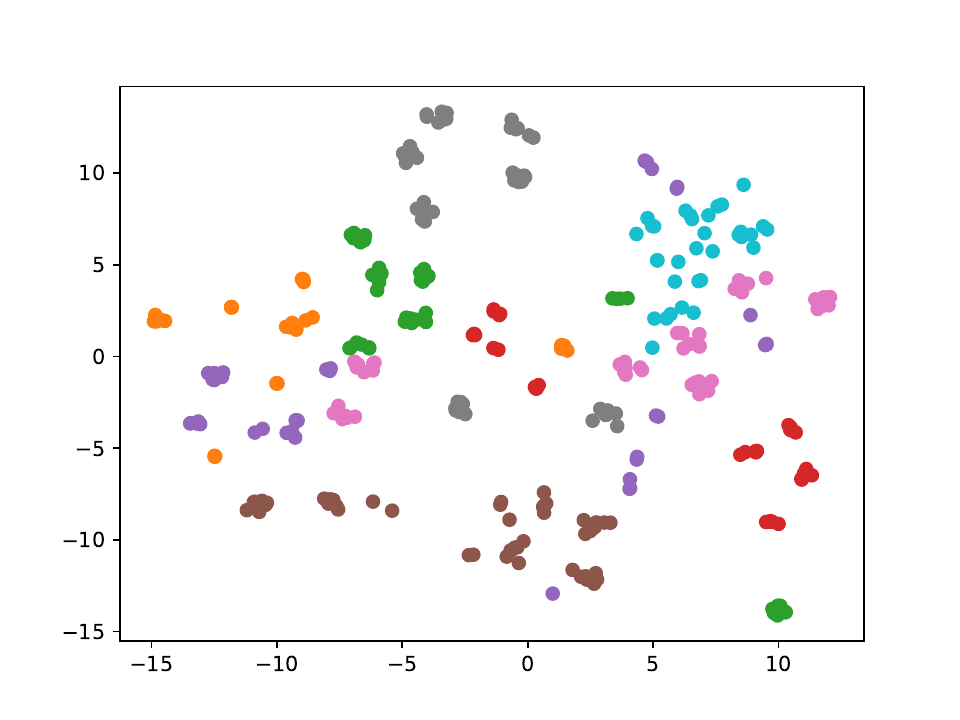}
    \caption{\textcolor{c0}{HandstandPushups}, \textcolor{c1}{WallPushups}, \textcolor{c2}{Typing}, \textcolor{c3}{Knitting}, \textcolor{c4}{Archery}, \textcolor{c5}{Shotput}, \textcolor{c6}{PlayingDhol}, \textcolor{c7}{PlayingFlute}, \textcolor{c8}{WalkingWithADog}, \textcolor{c9}{BlowingCandles}}
\end{subfigure}
\hfill%
\begin{subfigure}[t]{0.49\textwidth}
    \includegraphics[clip, trim=2cm 1.2cm 0.5cm 1cm, width=0.49\textwidth]{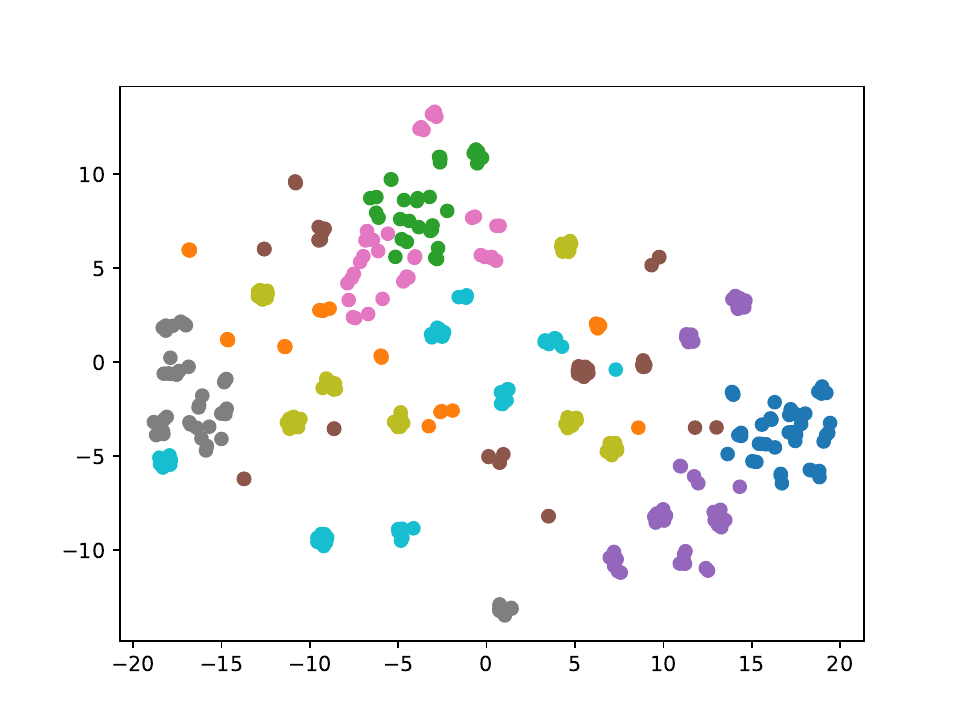}
    \includegraphics[clip, trim=2cm 1.2cm 0.5cm 1cm,width=0.49\textwidth]{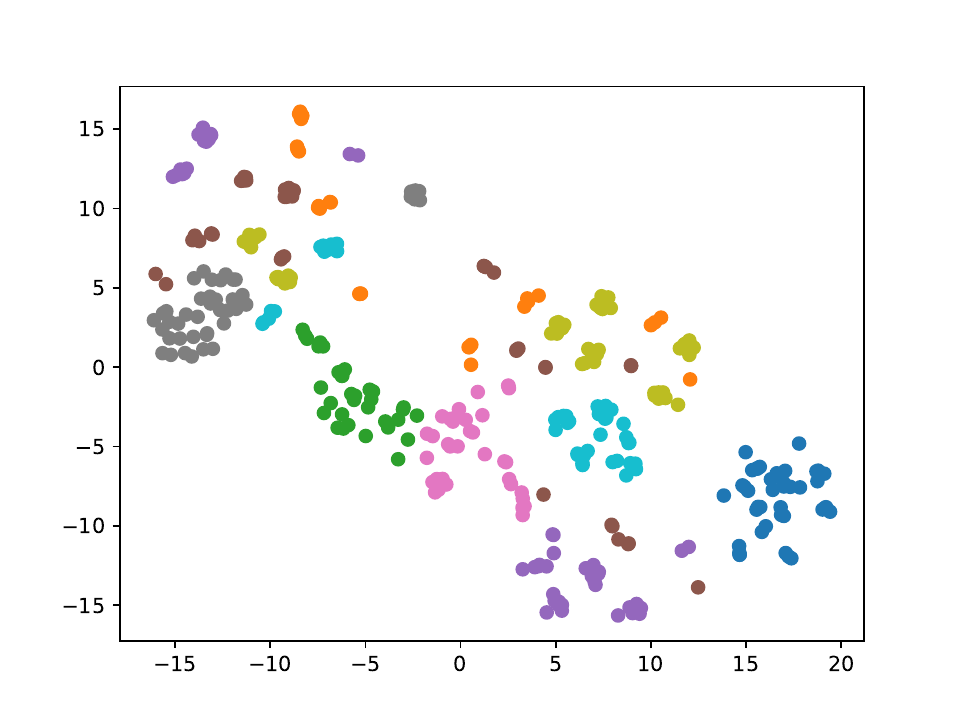}
    \caption{\textcolor{c0}{Swing}, \textcolor{c1}{BodyWeightSquats}, \textcolor{c2}{CuttingInKitchen}, \textcolor{c3}{MixingBatter}, \textcolor{c4}{Shotput}, \textcolor{c5}{Archery}, \textcolor{c6}{BabyCrawling}, \textcolor{c7}{WritingOnBoard}, \textcolor{c8}{PlayingDhol}, \textcolor{c9}{PlayingSitar}}
\end{subfigure}
\vspace{-3mm}
\caption{t-SNE visualization of representations learned using \textbf{Instance Contrastive loss (left)} and  \textbf{TCLR (right)} of  randomly chosen action classes from UCF101 Test set. In case of TCLR, we observe that the class boundaries are more compact and discriminative.}
\label{fig:tsne}
\vspace{-3mm}
\end{figure*}

We use t-SNE \cite{tsne} to visualize and study the representation learned by the model during self-supervised pre-training (without any supervised finetuning).  We compare our method to the standard instance contrastive pretraining and randomly initialized features for a selected set of classes in Figure~\ref{fig:tsnecompare} for better visibility. In Figure~\ref{fig:tsne} we compare TCLR to standard instance contrastive training for different sets of classes are selected at random to get a broad look at the overall representation space. The training set (split-1) of UCF101 without labels is used for pre-training and the test set is used for visualization. In order to ensure reproducible t-SNE results, we use PCA initialization and set perplexity to 50. As can be seen by comparing the representations, we can see that TCLR results in well separated clusters compared to the standard instance contrastive loss.

\section{Model Attention}
\label{sec:attention}

\begin{figure*}[h!]
    \centering
\begin{overpic}[width=0.49\textwidth]{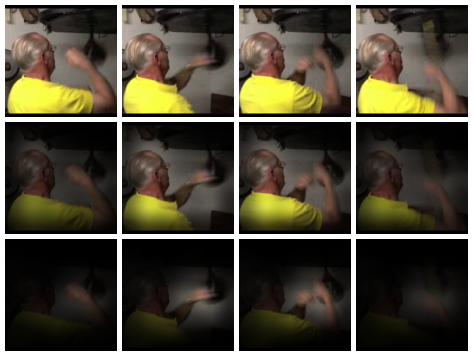}
 \put (1,71.5) {\colorbox{white}{\textcolor{blue}{\scriptsize BoxingSpeedBag}}}
 \put (1,46.75) {\colorbox{white}{\textcolor{c2}{\scriptsize BoxingSpeedBag}}}
 \put (1,22) {\colorbox{white}{\textcolor{c2}{\scriptsize BoxingSpeedBag}}}
\end{overpic}
\hfill %
\begin{overpic}[width=0.49\textwidth]{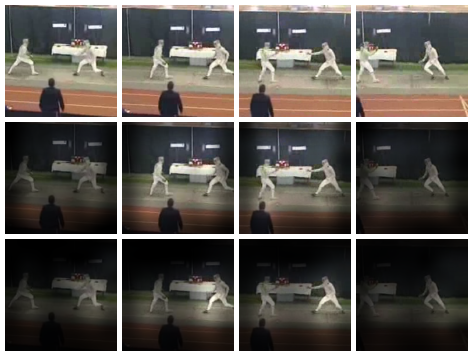}
 \put (1,71.5) {\colorbox{white}{\textcolor{blue}{\scriptsize Fencing}}}
 \put (1,46.75) {\colorbox{white}{\textcolor{c2}{\scriptsize Fencing}}}
 \put (1,22) {\colorbox{white}{\textcolor{c2}{\scriptsize Fencing}}}
\end{overpic}
\hfill %
\begin{overpic}[width=0.49\textwidth]{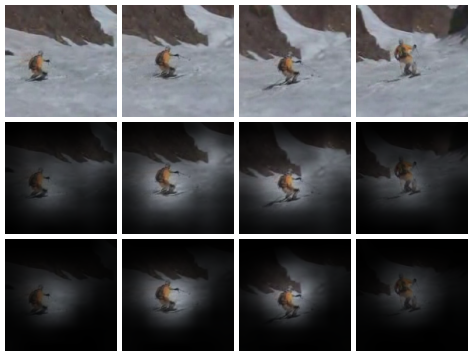}
 \put (1,71.5) {\colorbox{white}{\textcolor{blue}{\scriptsize Skiing}}}
 \put (1,46.75) {\colorbox{white}{\textcolor{c3}{\scriptsize Kayaking}}}
 \put (1,22) {\colorbox{white}{\textcolor{c2}{\scriptsize Skiing}}}
\end{overpic}
\hfill %
\begin{overpic}[width=0.49\textwidth]{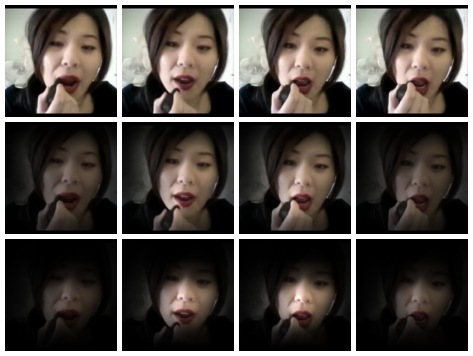}
 \put (1,71.5) {\colorbox{white}{\textcolor{blue}{\scriptsize ApplyLipstick}}}
 \put (1,46.75) {\colorbox{white}{\textcolor{c2}{\scriptsize ApplyLipstick}}}
 \put (1,22) {\colorbox{white}{\textcolor{c2}{\scriptsize ApplyLipstick}}}
\end{overpic}
\hfill %
\begin{overpic}[width=0.49\textwidth]{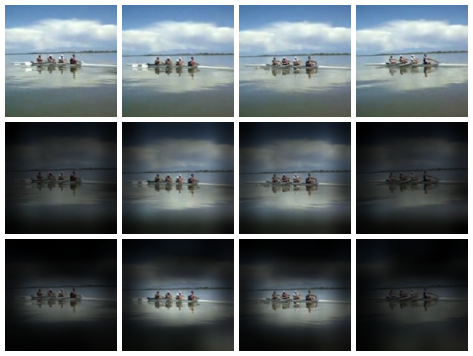}
 \put (1,71.5) {\colorbox{white}{\textcolor{blue}{\scriptsize Rowing}}}
 \put (1,46.75) {\colorbox{white}{\textcolor{c3}{\scriptsize Skijet}}}
 \put (1,22) {\colorbox{white}{\textcolor{c2}{\scriptsize Rowing}}}
\end{overpic}
\hfill %
\begin{overpic}[width=0.49\textwidth]{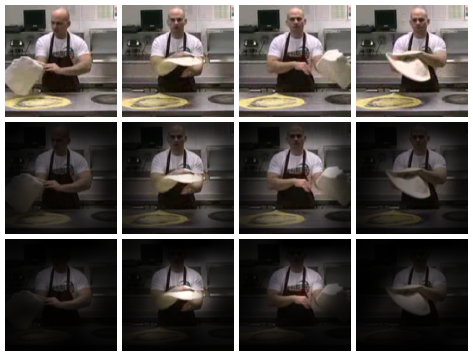}
 \put (1,71.5) {\colorbox{white}{\textcolor{blue}{\scriptsize PizzaTossing}}}
 \put (1,46.75) {\colorbox{white}{\textcolor{c2}{\scriptsize PizzaTossing}}}
 \put (1,22) {\colorbox{white}{\textcolor{c2}{\scriptsize PizzaTossing}}}
\end{overpic}
\caption{\textbf{Model attention:}  Ground-Truth Label on Video frames \textbf{(Top Row)}, Attention for baseline fully supervised model \textbf{(Middle Row)} and Attention for TCLR pre-trained model\textbf{(Bottom Row)}. We notice that for the TCLR pre-trained model the attention is more focused on action-centric regions.}
    \label{fig:qual1}
\end{figure*}

\begin{figure*}[h!]
    \centering
\begin{overpic}[width=0.49\textwidth]{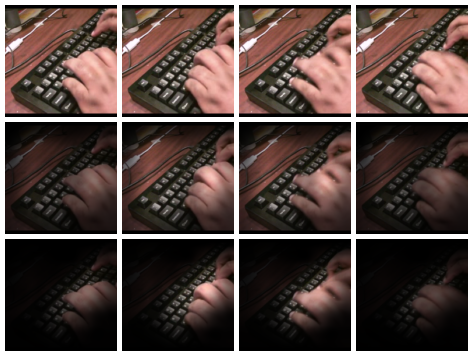}
 \put (1,71.5) {\colorbox{white}{\textcolor{blue}{\scriptsize Typing}}}
 \put (1,46.75) {\colorbox{white}{\textcolor{c2}{\scriptsize Typing}}}
 \put (1,22) {\colorbox{white}{\textcolor{c2}{\scriptsize Typing}}}
\end{overpic}
\hfill %
\begin{overpic}[width=0.49\textwidth]{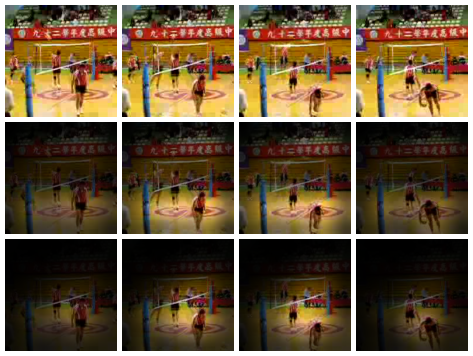}
 \put (1,71.5) {\colorbox{white}{\textcolor{blue}{\scriptsize VolleyballSpiking}}}
 \put (1,46.75) {\colorbox{white}{\textcolor{c3}{\scriptsize BasketballDunk}}}
 \put (1,22) {\colorbox{white}{\textcolor{c2}{\scriptsize VolleyballSpiking}}}
\end{overpic}
\hfill %
\begin{overpic}[width=0.49\textwidth]{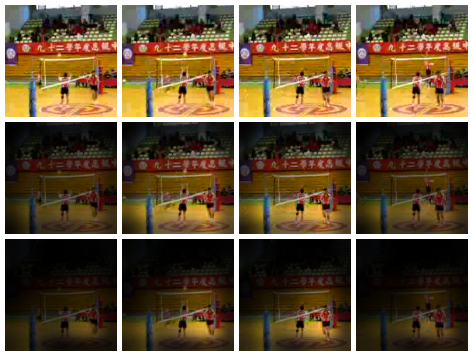}
 \put (1,71.5) {\colorbox{white}{\textcolor{blue}{\scriptsize VolleyballSpiking}}}
 \put (1,46.75) {\colorbox{white}{\textcolor{c3}{\scriptsize Basketball}}}
 \put (1,22) {\colorbox{white}{\textcolor{c2}{\scriptsize VolleyballSpiking}}}
\end{overpic}
\hfill %
\begin{overpic}[width=0.49\textwidth]{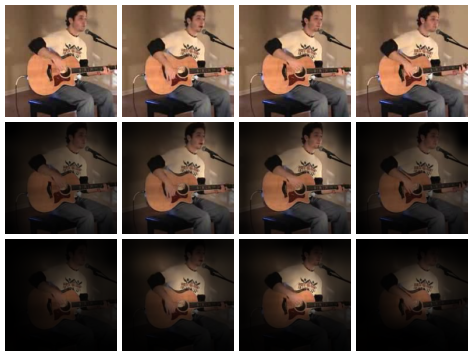}
 \put (1,71.5) {\colorbox{white}{\textcolor{blue}{\scriptsize PlayingGuitar}}}
 \put (1,46.75) {\colorbox{white}{\textcolor{c2}{\scriptsize PlayingGuitar}}}
 \put (1,22) {\colorbox{white}{\textcolor{c2}{\scriptsize PlayingGuitar}}}
\end{overpic}
\hfill %
\begin{overpic}[width=0.49\textwidth]{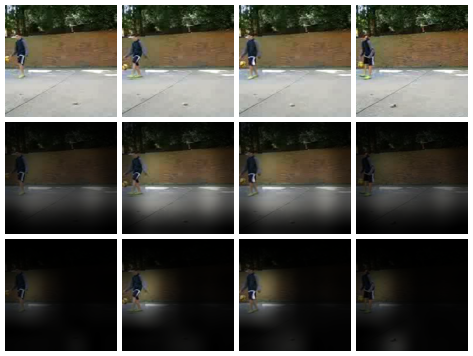}
 \put (1,71.5) {\colorbox{white}{\textcolor{blue}{\scriptsize SoccerJuggling}}}
 \put (1,46.75) {\colorbox{white}{\textcolor{c3}{\scriptsize Shotput}}}
 \put (1,22) {\colorbox{white}{\textcolor{c2}{\scriptsize SoccerJuggling}}}
\end{overpic}
\hfill %
\begin{overpic}[width=0.49\textwidth]{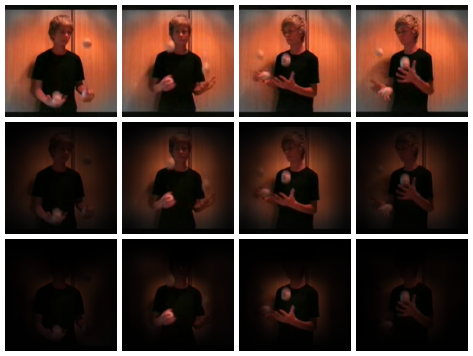}
 \put (1,71.5) {\colorbox{white}{\textcolor{blue}{\scriptsize JugglingBalls}}}
 \put (1,46.75) {\colorbox{white}{\textcolor{c2}{\scriptsize JugglingBalls}}}
 \put (1,22) {\colorbox{white}{\textcolor{c2}{\scriptsize JugglingBalls}}}
\end{overpic}
\caption{\textbf{Model attention (continued):}  Ground-Truth Label on Video frames \textbf{(Top Row)}, Attention for baseline fully supervised model \textbf{(Middle Row)} and Attention for TCLR pre-trained model\textbf{(Bottom Row)}. We notice that for the TCLR pre-trained model the attention is more focused on action-centric regions. %
}
    \label{fig:qual2}
\end{figure*}

\begin{figure*}[h!]
    \centering
\begin{overpic}[width=0.49\textwidth]{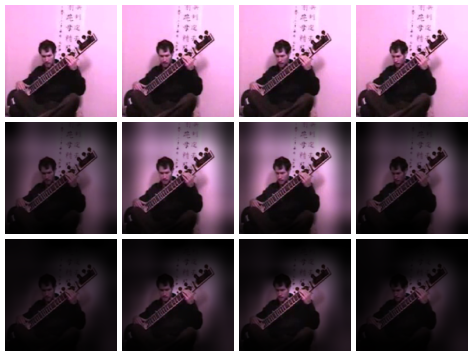}
 \put (1,71.5) {\colorbox{white}{\textcolor{blue}{\scriptsize PlayingSitar}}}
 \put (1,46.75) {\colorbox{white}{\textcolor{c3}{\scriptsize CuttingInKitchen}}}
 \put (1,22) {\colorbox{white}{\textcolor{c2}{\scriptsize PlayingSitar}}}
\end{overpic}
\hfill %
\begin{overpic}[width=0.49\textwidth]{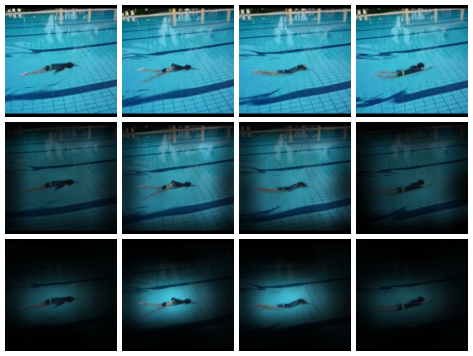}
 \put (1,71.5) {\colorbox{white}{\textcolor{blue}{\scriptsize FrontCrawl}}}
 \put (1,46.75) {\colorbox{white}{\textcolor{c2}{\scriptsize FrontCrawl}}}
 \put (1,22) {\colorbox{white}{\textcolor{c2}{\scriptsize FrontCrawl}}}
\end{overpic}
\hfill %
\begin{overpic}[width=0.49\textwidth]{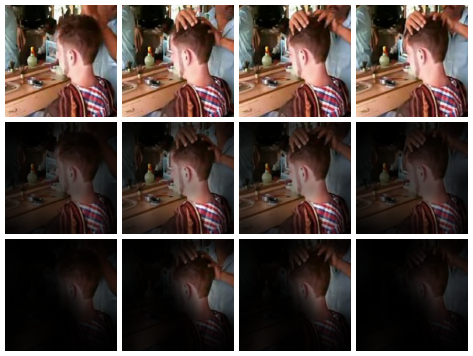}
 \put (1,71.5) {\colorbox{white}{\textcolor{blue}{\scriptsize HeadMassage}}}
 \put (1,46.75) {\colorbox{white}{\textcolor{c3}{\scriptsize PlayingTabla}}}
 \put (1,22) {\colorbox{white}{\textcolor{c2}{\scriptsize HeadMassage}}}
\end{overpic}
\hfill %
\begin{overpic}[width=0.49\textwidth]{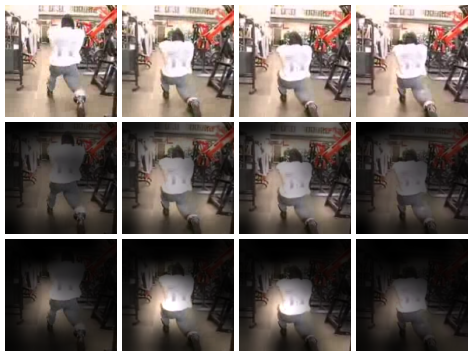}
 \put (1,71.5) {\colorbox{white}{\textcolor{blue}{\scriptsize Lunges}}}
 \put (1,46.75) {\colorbox{white}{\textcolor{c3}{\scriptsize PizzaTossing}}}
 \put (1,22) {\colorbox{white}{\textcolor{c2}{\scriptsize Lunges}}}
\end{overpic}
\hfill %
\begin{overpic}[width=0.49\textwidth]{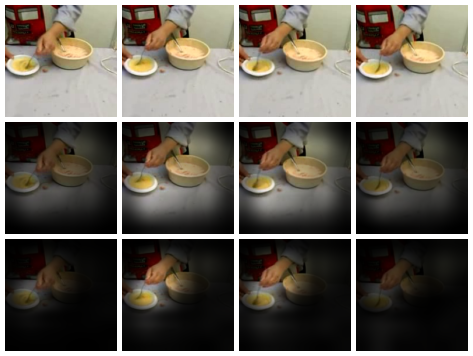}
 \put (1,71.5) {\colorbox{white}{\textcolor{blue}{\scriptsize Mixing}}}
 \put (1,46.75) {\colorbox{white}{\textcolor{c3}{\scriptsize CuttingInKitchen}}}
 \put (1,22) {\colorbox{white}{\textcolor{c2}{\scriptsize Mixing}}}
\end{overpic}
\hfill %
\begin{overpic}[width=0.49\textwidth]{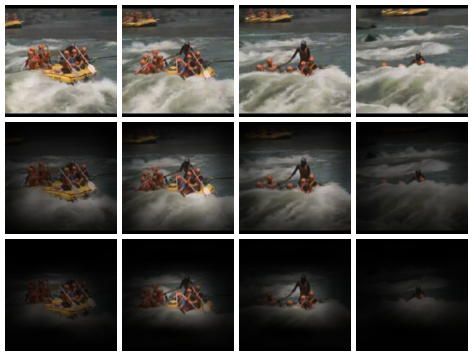}
 \put (1,71.5) {\colorbox{white}{\textcolor{blue}{\scriptsize Rafting}}}
 \put (1,46.75) {\colorbox{white}{\textcolor{c2}{\scriptsize Rafting}}}
 \put (1,22) {\colorbox{white}{\textcolor{c2}{\scriptsize Rafting}}}
\end{overpic}
\caption{\textbf{Model attention (continued):}  Ground-Truth Label on Video frames \textbf{(Top Row)}, Attention for baseline fully supervised model \textbf{(Middle Row)} and Attention for TCLR pre-trained model\textbf{(Bottom Row)}.We notice that for the TCLR pre-trained model the attention is more focused on action-centric regions. }
    \label{fig:qual3}
\end{figure*}

\begin{figure*}[h!]
    \centering
\begin{overpic}[width=0.49\textwidth]{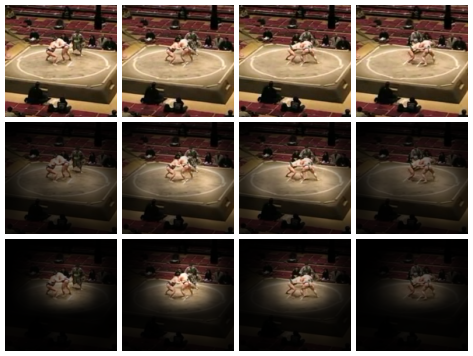}
 \put (1,71.5) {\colorbox{white}{\textcolor{blue}{\scriptsize SumoWrestling}}}
 \put (1,46.75) {\colorbox{white}{\textcolor{c2}{\scriptsize SumoWrestling}}}
 \put (1,22) {\colorbox{white}{\textcolor{c2}{\scriptsize SumoWrestling}}}
\end{overpic}
\hfill %
\begin{overpic}[width=0.49\textwidth]{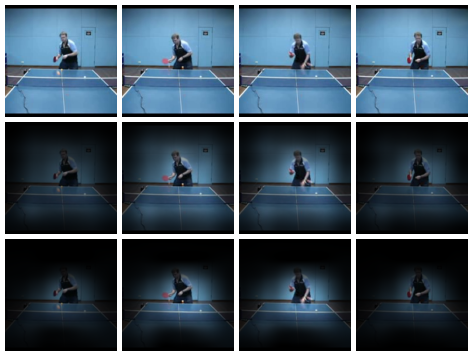}
 \put (1,71.5) {\colorbox{white}{\textcolor{blue}{\scriptsize TableTennisShot}}}
 \put (1,46.75) {\colorbox{white}{\textcolor{c2}{\scriptsize TableTennisShot}}}
 \put (1,22) {\colorbox{white}{\textcolor{c2}{\scriptsize TableTennisShot}}}
\end{overpic}
\hfill %
\begin{overpic}[width=0.49\textwidth]{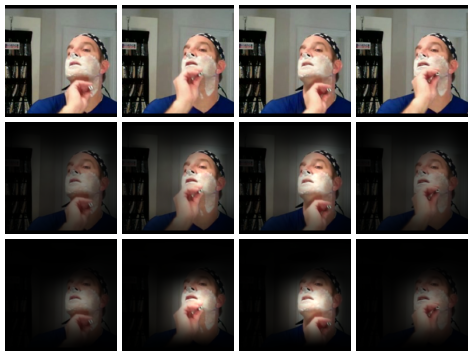}
 \put (1,71.5) {\colorbox{white}{\textcolor{blue}{\scriptsize ShavingBeard}}}
 \put (1,46.75) {\colorbox{white}{\textcolor{c2}{\scriptsize ShavingBeard}}}
 \put (1,22) {\colorbox{white}{\textcolor{c2}{\scriptsize ShavingBeard}}}
\end{overpic}
\hfill %
\begin{overpic}[width=0.49\textwidth]{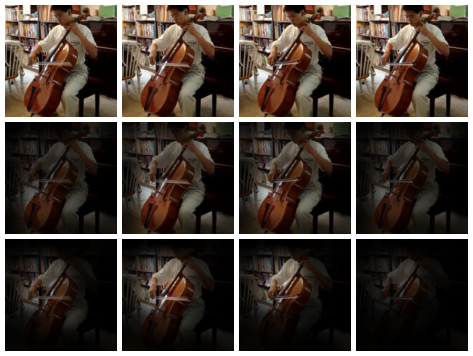}
 \put (1,71.5) {\colorbox{white}{\textcolor{blue}{\scriptsize PlayingCello}}}
 \put (1,46.75) {\colorbox{white}{\textcolor{c2}{\scriptsize PlayingCello}}}
 \put (1,22) {\colorbox{white}{\textcolor{c2}{\scriptsize PlayingCello}}}
\end{overpic}
\hfill %
\begin{overpic}[width=0.49\textwidth]{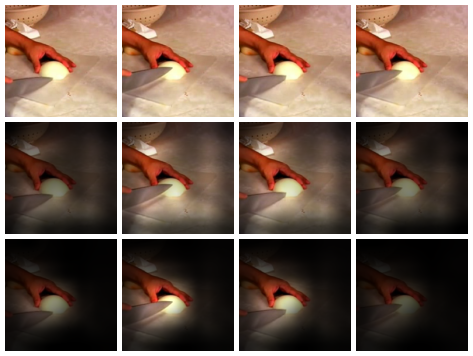}
 \put (1,71.5) {\colorbox{white}{\textcolor{blue}{\scriptsize CuttingInKitchen}}}
 \put (1,46.75) {\colorbox{white}{\textcolor{c3}{\scriptsize BabyCrawling}}}
 \put (1,22) {\colorbox{white}{\textcolor{c2}{\scriptsize CuttingInKitchen}}}
\end{overpic}
\hfill %
\begin{overpic}[width=0.49\textwidth]{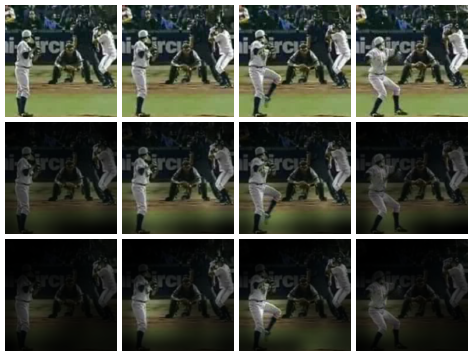}
 \put (1,71.5) {\colorbox{white}{\textcolor{blue}{\scriptsize BaseballPitch}}}
 \put (1,46.75) {\colorbox{white}{\textcolor{c3}{\scriptsize GolfSwing}}}
 \put (1,22) {\colorbox{white}{\textcolor{c2}{\scriptsize BaseballPitch}}}
\end{overpic}
\caption{\textbf{Model attention (continued):}  Ground-Truth Label on Video frames \textbf{(Top Row)}, Attention for baseline fully supervised model \textbf{(Middle Row)} and Attention for TCLR pre-trained model\textbf{(Bottom Row)}. We notice that for the TCLR pre-trained model the attention is more focused on action-centric regions. }
    \label{fig:qual4}
\end{figure*}
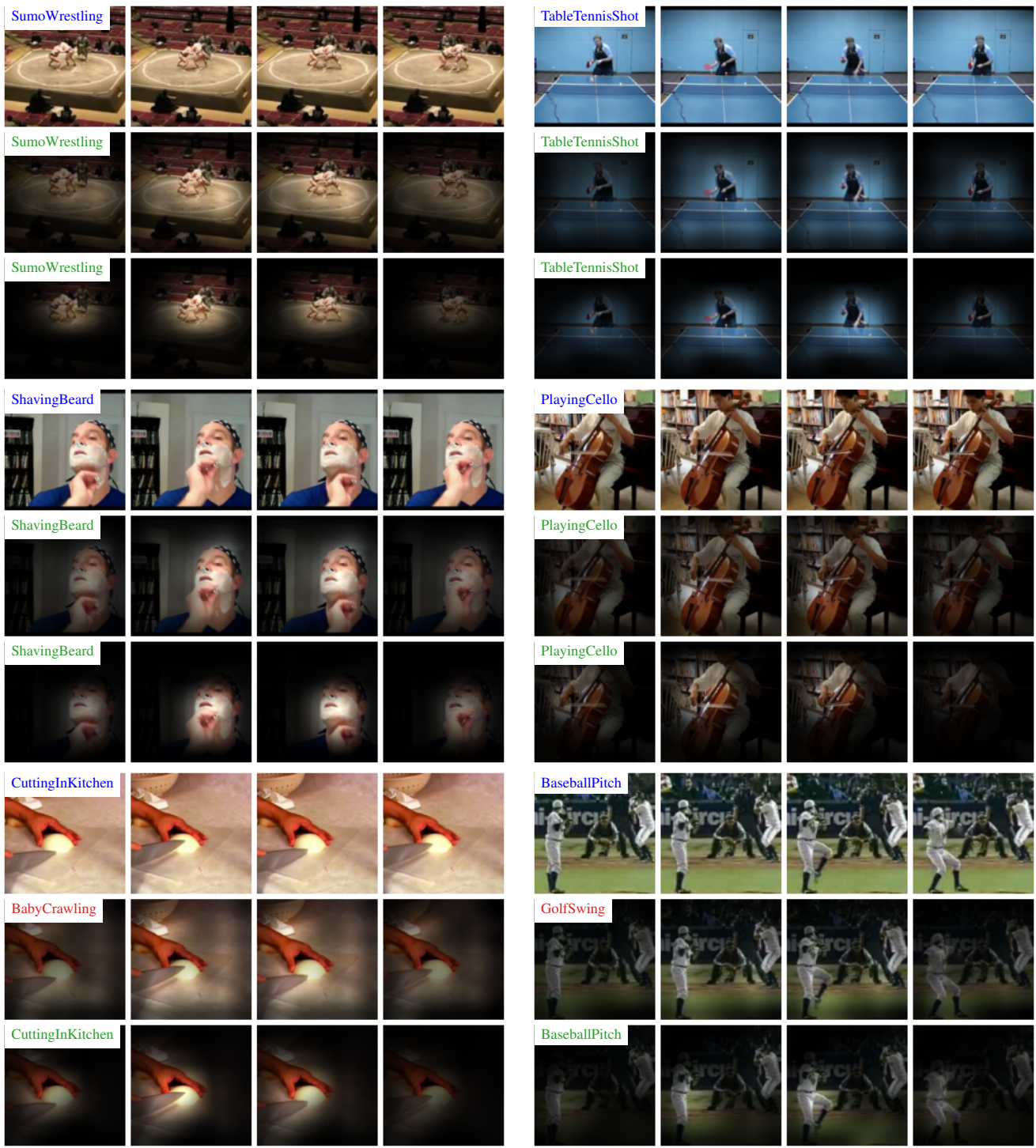

In order to further examine the effect of TCLR pre-training on model performance, we use the method of Zagoruyko and  Komodakis~\cite{attentiontransfer} to generate model attention for our method (from the fourth convolutional block of the R3D model) and compare it with the baseline supervised model. The qualitative samples are presented in Figures~\ref{fig:qual1}, \ref{fig:qual2}, \ref{fig:qual3}, and \ref{fig:qual4}. We observe that the TCLR pre-trained model has significantly better focus on relevant portions of the video.

\clearpage
\clearpage

\section{Detailed comparison with recent prior work}
\label{sec:detailed_comparison}

We contrast our method with 6 recent works in the area: CVRL, TaCo, CoCLR, IIC,  Video DeepInfoMax and SeCo.

\subsection{CVRL~\cite{cvrl}}

\begin{figure}[h]
\begin{center}
 \includegraphics[width=1.0\linewidth]{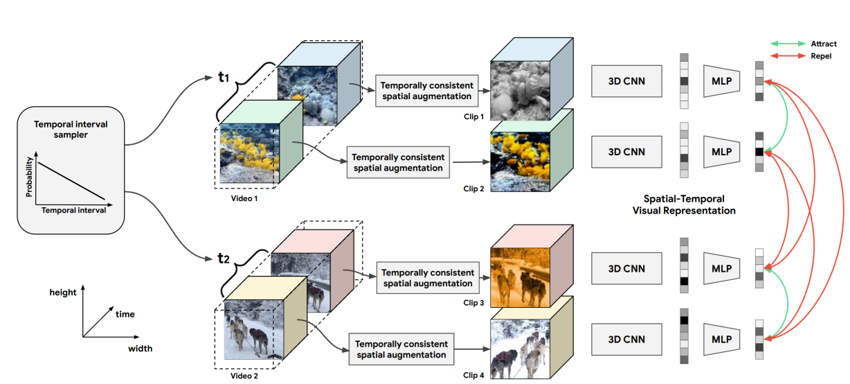}
\end{center}
\vspace{-5mm}
  \caption{CVRL Framework from Qian et al.~\cite{cvrl}}
\label{fig:cvrl}
\end{figure}

CVRL introduces the Temporal Interval Sampler, the key idea behind it being to not take positives from uniformly random timestamps from the same instance, rather take positives which close to each other temporally. CVRL improves upon standard instance contrastive learning by trying to avoid learning excessive temporal invariance while TCLR explicitly learns within-instance temporal distinctiveness by taking negatives from the same instance, and hence is an orthogonal approach to the problem of excessive temporal invariance. 

\subsection{TaCo~\cite{taco}}

\begin{figure}[h]
\begin{center}
 \includegraphics[width=1.0\linewidth]{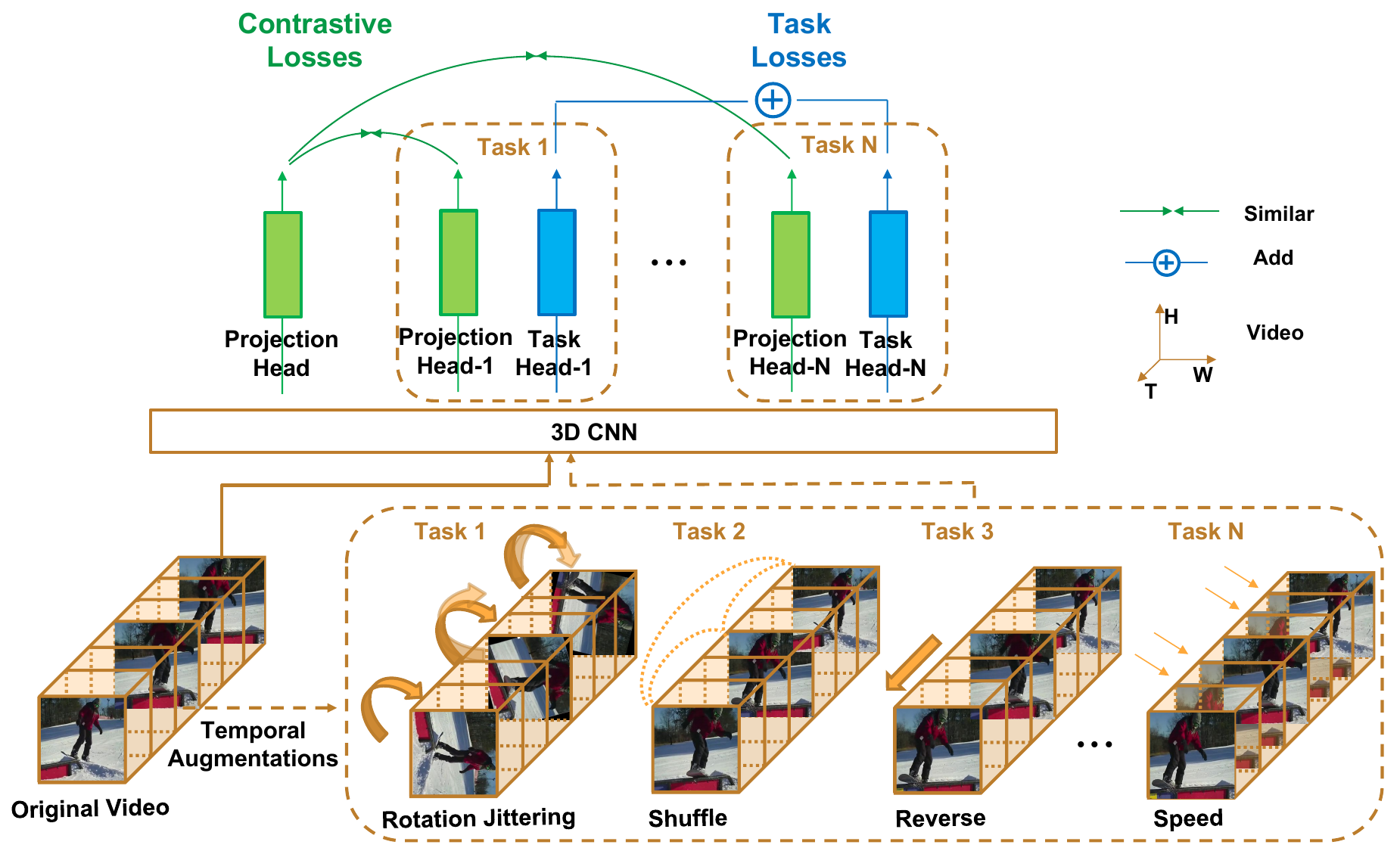}
\end{center}
\vspace{-5mm}
   \caption{TaCo Framework from Bai et al.~\cite{taco}}
\label{fig:taco}
\end{figure}

TaCo combines instance contrastive learning with multiple temporal pretext tasks (rotation, shuffled, reverse, or speed) in a multi-task setting in order to learn temporally varying features. The different tasks have their own projection and task heads, but share a common backbone. The TaCo approach to learning temporally distinct features is a differnt approach from TCLR, where we utilize temporal contrastive losses instead of any pretext tasks.

\subsection{CoCLR~\cite{cotraining}}

\begin{figure}[h]
\begin{center}
 \includegraphics[width=1.0\linewidth]{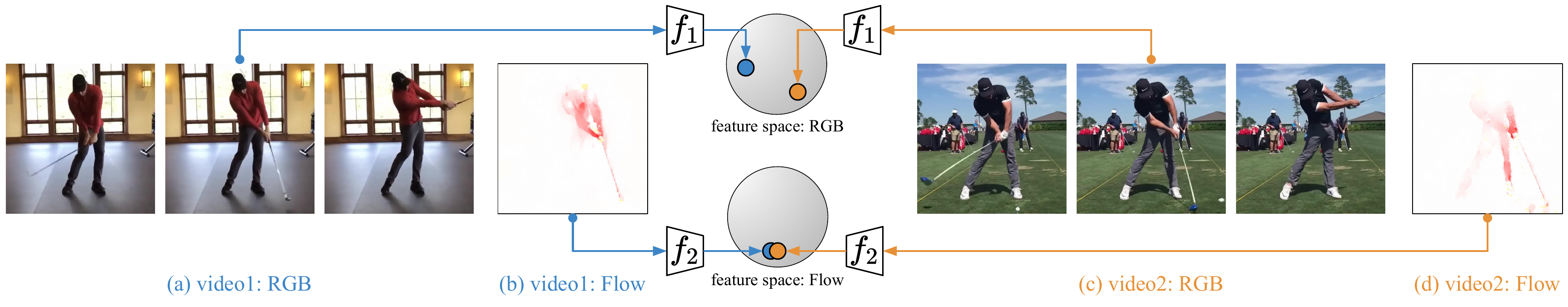}
\end{center}
\vspace{-5mm}
   \caption{CoCLR Framework from Han et al.~\cite{cotraining}}
\label{fig:coclr}
\end{figure}
CoCLR introduces a multi-stage pretraining process that improves upon instance contrastive learning by mining positive pairs across instances in a cross-modal fashion, i.e. positive pairs mined from RGB modality are used for optical flow, and vice versa. 

\subsection{IIC~\cite{iic}}

\begin{figure}[h]
\begin{center}
 \includegraphics[width=1.0\linewidth]{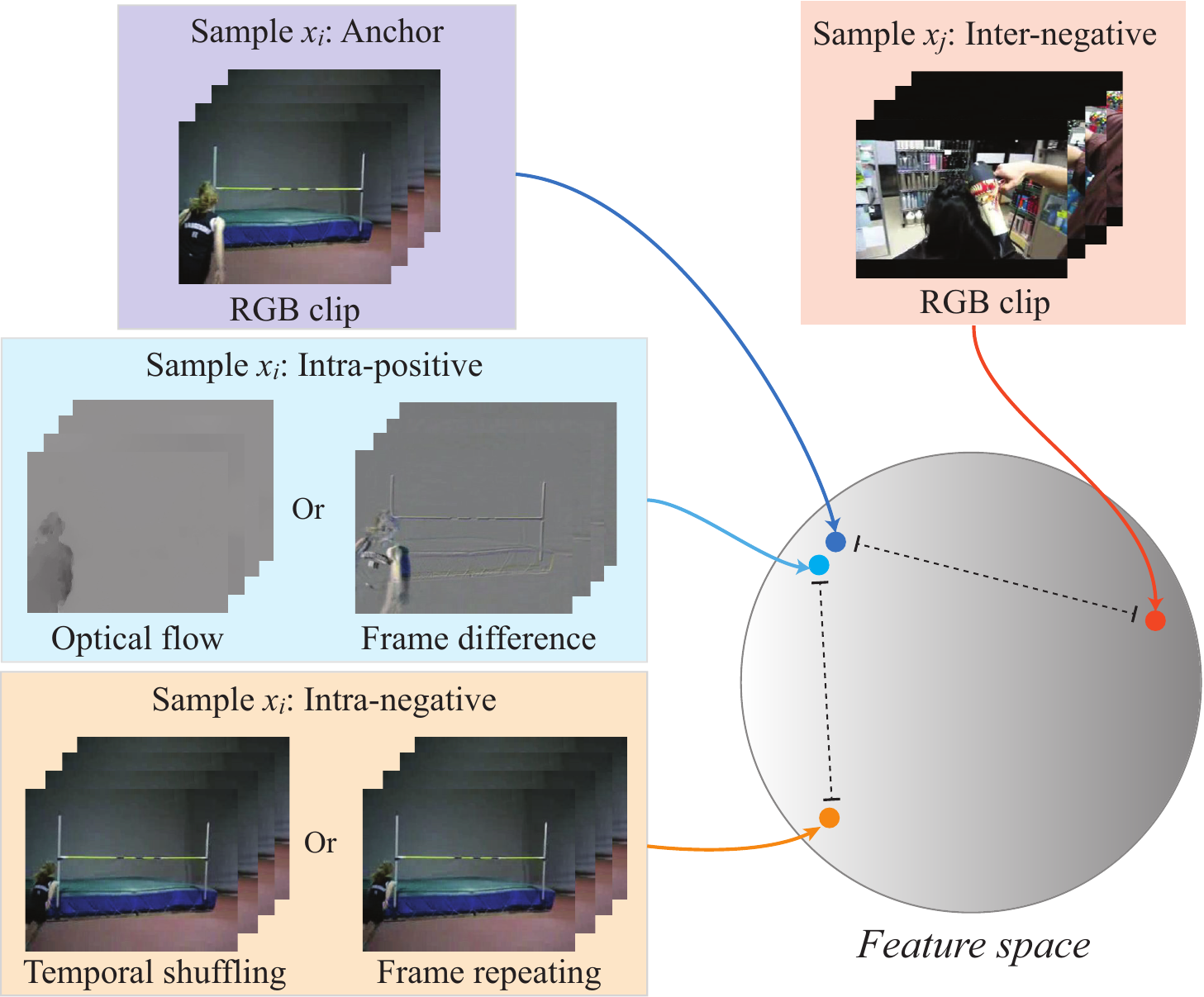}
\end{center}
\vspace{-5mm}
   \caption{IIC Framework from Tao et al.~\cite{iic}}
\label{fig:iic}
\end{figure}

IIC framework builds on the contrastive loss by utilizing positive pairs from different modalities (such as optical flow) and creating a new class of negatives by shuffling and repeating the anchor. Unlike TCLR's $\mathcal{L}_{LL}$ loss, IIC does not use temporally distinct clips as negative pairs.

\subsection{VideoDeepInfoMax~\cite{vdim}}

\begin{figure}[h]
\begin{center}
 \includegraphics[width=1.0\linewidth]{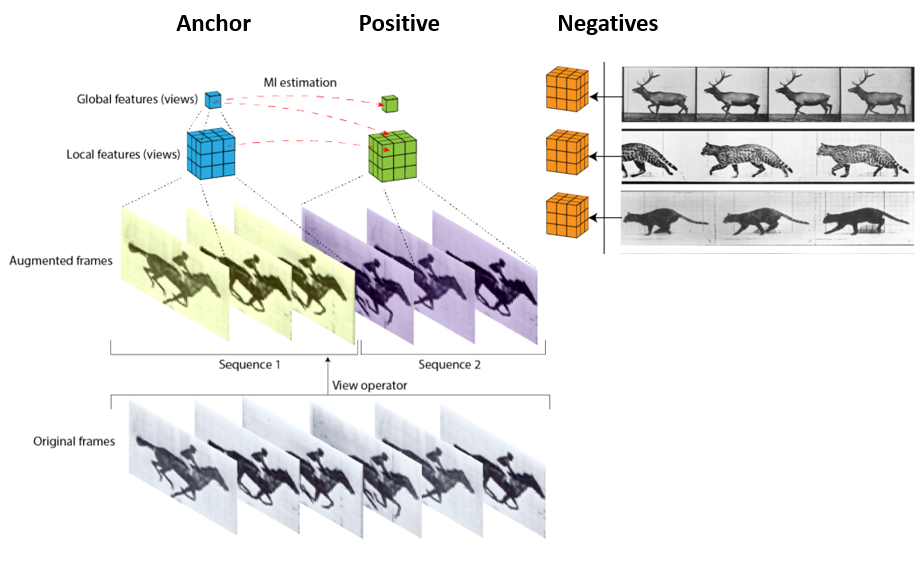}
\end{center}
\vspace{-5mm}
   \caption{Video DIM Framework from Devon Hjelm and Bachman~\cite{cotraining}}
\label{fig:vdim}
\end{figure}

VideoDIM learns through \textit{instance discrimination} using both local and global views of the features. \textit{Positive} pairs are formed from local and global features of the \textit{same} instance, whereas \textit{negative} pairs are formed from local and global features of \textit{different} instances.The local features are obtained from lower level convolutional layers, whereas the global feature comes from the final layer. Like other forms of instance discrimination based learning VideoDIM does not achieve temporal distinctiveness, and rather enforces invariance. Whereas, $\mathcal{L}_{GL}$ loss in TCLR promotes distinctiveness in the local features along the temporal dimension.

\subsection{Comparison with SeCo~\cite{seco}}

SeCo combines multiple pretext tasks and its results cannot be compared with VideoSSL methods like TCLR which {\em start from scratch}; SeCo uses a 2D CNN initialized with \textit{ImageNet MoCov2 pre-trained} weights. Moreover, most of the gain of SeCo over ImageNet trained initialization comes from instance contrastive loss and adding intra-frame loss in SeCo only results in gain of $\sim$1-2\% across downstream tasks (Table 1 of~\cite{seco}). Whereas with TCLR, gains across downstream tasks upon adding the Local-Local loss are significant ($\sim$6-11\%, see Table 4 of the main paper). SeCo doesn't include any temporal component (like 3D-Conv or RNN) in the backbone during the SSL phase; it uses 2D CNN and as a result their intra-frame contrastive loss only learns to discriminate \textit{visual appearance} instead of \textit{spatio-temporal} features.

\end{document}